\documentclass{article} 
\usepackage{iclr2024_conference,times}

\usepackage{amsmath,amsfonts,bm}

\def\eqref#1{equation~\ref{#1}}

\def\1{\bm{1}}

\DeclareMathAlphabet{\mathsfit}{\encodingdefault}{\sfdefault}{m}{sl}
\SetMathAlphabet{\mathsfit}{bold}{\encodingdefault}{\sfdefault}{bx}{n}

\usepackage{hyperref}
\usepackage{url}
\usepackage{graphicx}
\usepackage{subcaption}
\usepackage{float}
\usepackage{caption}
\usepackage{color}
\usepackage{comment}
\usepackage{array}

\newcommand{\Figure}[1]{Fig.~\ref{#1}}

\newcommand{\Section}[1]{\S\ref{#1}}

\newcommand{\bs}[1]{\boldsymbol{#1}}

\title{Impacts of Color and Texture Distortions on Earth Observation Data in Deep Learning}

\author{Martin Willbo$^1$, Aleksis Pirinen$^1$, John Martinsson$^{1,2}$, Edvin Listo Zec$^{1,3}$, \\ \textbf{Olof Mogren$^{1,4}$, Mikael Nilsson$^{2}$}\\
$^1$RISE Research Institutes of Sweden\\
$^2$Centre for Mathematical Sciences, Lund University, Sweden \\
$^3$KTH Royal Institute of Technology \\
$^4$Swedish Centre for Impacts of Climate Extremes (climes)\\
\tt\small \{martin.willbo@ri.se, aleksis.pirinen@ri.se, john.martinsson@ri.se \\
\tt\small edvin.listo.zec@ri.se, olof.mogren@ri.se, mikael.nilsson@math.lth.se\}\vspace{-4mm}
}

\iclrfinalcopy 
\begin{document}

\maketitle

\begin{abstract}
Land cover classification and change detection are two important applications of remote sensing and Earth observation (EO) that have benefited greatly from the advances in deep learning. Convolutional and transformer-based U-net models are the state-of-the-art architectures for these tasks, and their performances have been boosted by an increased availability of large-scale annotated EO datasets. However, the influence of different visual characteristics of the input EO data on a model’s predictions is not well understood. In this work we systematically examine model sensitivities with respect to several color- and texture-based distortions on the input EO data during inference, given models that have been trained without such distortions. We conduct experiments with multiple state-of-the-art segmentation networks for land cover classification and show that they are in general more sensitive to texture than to color distortions. Beyond revealing intriguing characteristics of widely used land cover classification models, our results can also be used to guide the development of more robust models within the EO domain.
\end{abstract}

\section{Introduction}
Land cover classification is a key application for remote sensing and Earth observation (EO) data, as it provides essential information for various domains, such as urban planning, environmental monitoring, disaster management, and agriculture. Deep neural networks, such as CNNs and transformers, have demonstrated impressive capabilities and results for processing satellite imagery \citep{florian2017rethinking,wang2022unetformer,zhao2023land}. However, these models, and the methods used to regularize them (e.g.~common image augmentation techniques), are mostly developed and tested on standard imagery (e.g.~from ImageNet), which may differ from EO images in several aspects. For instance, previous studies have shown that CNNs trained on ImageNet rely more on texture than on color or shape~\citep{conv_texture_bias}. Such dependencies are less explored in the EO domain, where texture and color may change due to factors such as seasonality, weather, and sensor noise. It is thus essential to understand how different types of visual features and data distortions affect the performance and robustness of deep learning models for EO tasks, and to develop new models and methods that are more suitable for EO data \citep{rolf2024mission}.

\begin{figure}[t]
    \centering
    \includegraphics[width=0.7\textwidth]{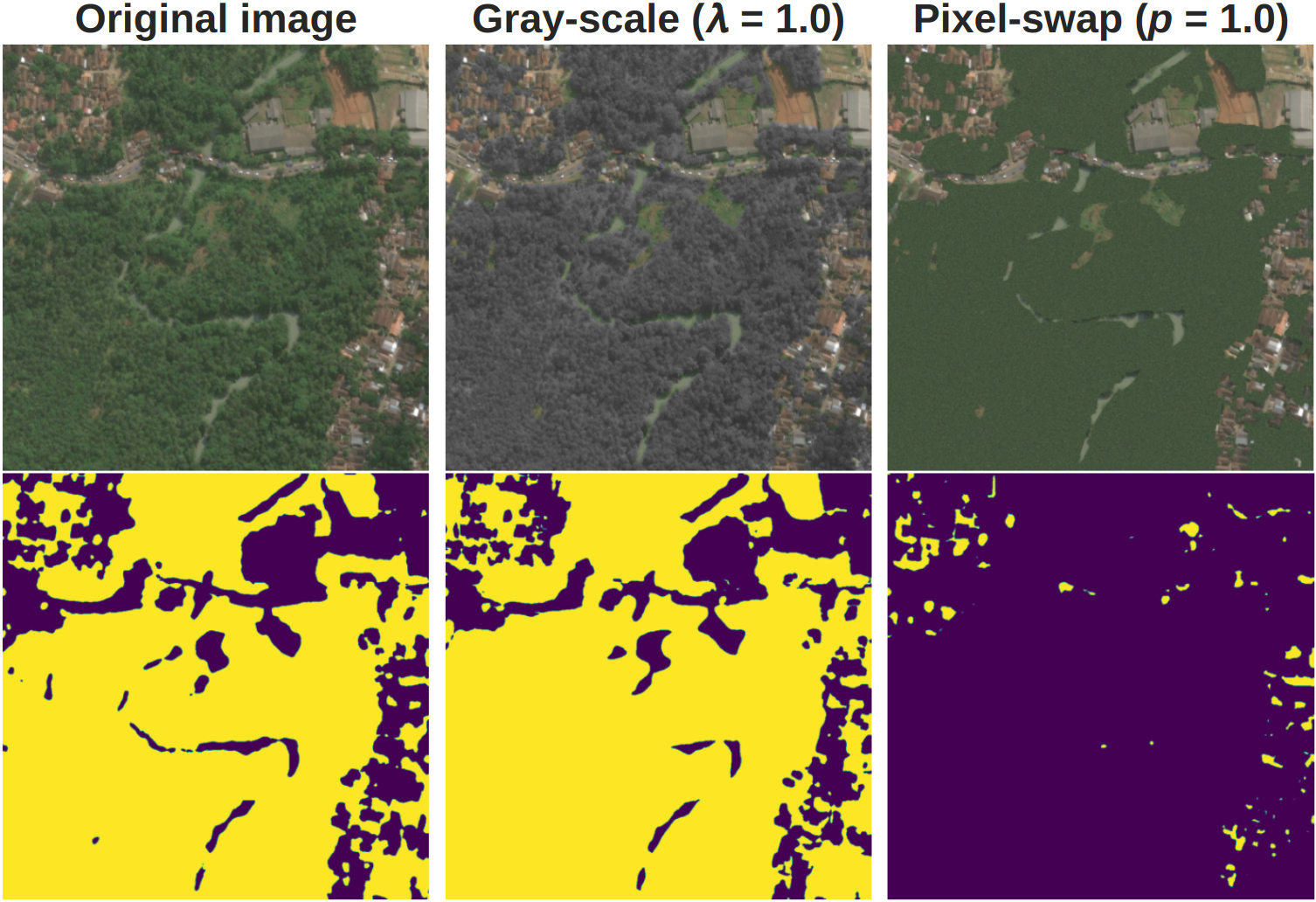}
    \vspace{-2pt}
    \caption{Example image from the training dataset (OpenEarthMap). The class considered here is \emph{tree}. Yellow and dark blue respectively show pixels predicted as \textit{tree} and not \textit{tree}. Top row: Original image, image with \textbf{gray-scale} transformation (color distortion) applied, and image with \textbf{pixel-swap} transformation (texture distortion) applied, respectively. Note that in the middle, the trees are gray even if they appear to be in color at a glance. Bottom row: Model predictions for the corresponding images in the first row. The transformations are defined in \Section{sec:method}; more transformations are explored in the appendix. Predictions made using U-Net-Efficientnet-B4.}
    \label{fig:prediction_demo}
\end{figure}

In this work we aim to provide a better understanding of how different types of test time distortions affect the performance of popular models trained on EO data for land cover classification, and to motivate the development of new data augmentation techniques that are more appropriate for EO models. We study the inductive biases and invariances of popular deep learning models for land cover classification, by applying various test time image distortions that the models have not seen during training. We propose a set of image distortion functions that are \emph{independently applied per image and semantic class} in land cover data: \textbf{(i)} converting the pixels of a class into gray-scale\footnote{Other transformations are also explored in the appendix.} (color distortion), and \textbf{(ii)} randomly swapping pixel values within a class in an image (texture distortion); see \Figure{fig:prediction_demo}. We then evaluate the performance of the models on OpenEarthMap \cite{xia2023openearthmap}, a large-scale and fully labeled benchmark dataset of high-resolution aerial images, under different distortion settings. Our results reveal the strengths and limitations of deep learning models for land cover classification, and offer guidance for future research and improvement.

\section{Related Work}
Image distortions, such as blur, noise, contrast variation, and JPEG compression, can substantially degrade the accuracy of deep neural networks (DNNs) when applied to the input data. This phenomenon has been demonstrated by \citet{dodge2016understanding}, who evaluated the impact of different quality distortions on CNNs. \citet{zhou2017classification} further investigated the performance of CNNs under blur and noise distortions, and proposed to improve the model robustness by 
re-training with noisy data. Our work is related to these works, as we also investigate the robustness of DNNs on distorted data. However, our focus is to investigate model sensitivities specifically on EO data, and we leave the development of more robust solutions to future work.

It was shown by \citet{conv_texture_bias} that common augmentation techniques, such as Random Resized Crop, can introduce bias towards texture rather than shape in the domain of standard image classification. They postulated that aggressive crops may remove distinguishing shape information and push the network to learn to discriminate by texture. We show that DNNs trained on EO data are inherently biased to discriminate by texture even without applying such augmentation techniques. Contemporary work by \citep{gong2024exploring} explores segmentation model sensitivity with respect to color variances that are introduced at test time in the standard image domain. They also propose training strategies oriented towards rendering models invariant to color perturbations. However, as our results suggest, segmentation models trained on EO data are already largely color invariant without applying such training strategies. This in turn points to important differences between biases in models trained on data from these two domains.
\begin{figure}[t]
    \centering
    \includegraphics[width=0.95\textwidth]{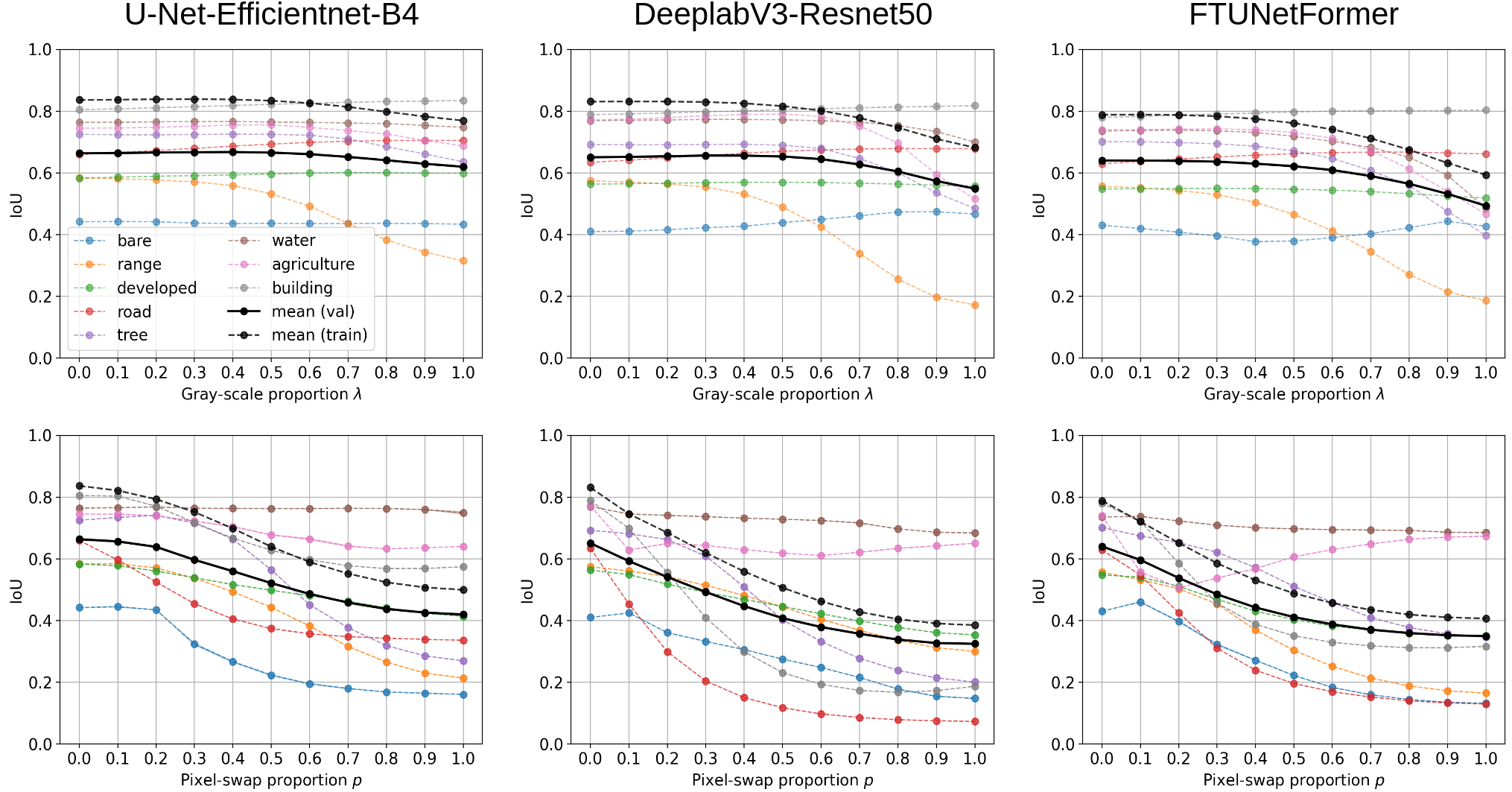}
    \caption{Impact of the \textbf{gray-scale} (top) and \textbf{pixel-swap} (bottom) transformations at test time on the validation set for the three segmentation models outlined in \Section{sec:method}. From left to right: U-Net-Efficientnet-B4, DeeplabV3-Resnet50, and FTUNetFormer. The solid black curve is the mean of the colored curves (validation data), and the dashed black curve is the corresponding mean on training data (included for comparison). Models are generally more sensitive to texture than color distortions.
     The pixel-swap plot curves are the mean over three realisations of the pixel-swap transform.}
    \label{fig:main_result_plot}
\end{figure}

\section{Experimental Setup and Description of Image Distortions} \label{sec:method}
We train and evaluate three models -- both CNN- and transformer-based ones -- on OpenEarthMap,
a large-scale benchmark for land cover classification. It contains 5,000 aerial and satellite images (RGB imagery) with 8 class labels at a resolution of 0.25-0.5m. The images span 97 regions from 44 countries across 6 continents, and are split into training, validation, and test sets (we use the official splits). We compare three segmentation models: U-Net-EfficientNet-B4 \citep{Iakubovskii:2019}, DeeplabV3-ResNet50 \citep{florian2017rethinking}, and FTUNetFormer \cite{wang2022unetformer}. During training, we apply only horizontal and vertical flips (an independent 50\% probability for each) as data augmentations. \textbf{Thus note that we do not apply any of the distortions described below\footnote{Please also see to the appendix for additional test time image distortions and associated results.} during training} (relevant follow-up experiments would however include investigating the effect of applying them also during model training). We use random crops of size $512 \times 512$ with a batch size of 10 during training, and full-size images during evaluation (more details are in the appendix). We ignore the background class; it constitutes $\sim0.6$\% of all pixels and is also ignored in the official OpenEarthMap benchmark.


\textbf{Gray-scale transformation (color distortion).}
Let $\mathcal{I}_{c} = \{(i_k, j_k)\}_{k=1}^{K}$ denote the set of pixel positions corresponding to class $c$ in a given RGB aerial or satellite image $\bs{I}$. Let $\bs{I}(i,j) \in [0,255]^3$ denote the pixel (three channels) at coordinate $(i,j) \in \mathcal{I}_{c}$, and let $\lambda \in [0,1]$. Then, the gray-scale transformation for image $\bs{I}$ and class $c$ is defined as $\bs{I}(i, j) = (1-\lambda) \bs{I}(i, j) + \lambda \bs{G}(i, j)$, where $\bs{G}$ is a corresponding gray-scale image derived by the \textit{rgb2gray} transformation from \citet{skimage-rgb2gray}. Other pixels are left unchanged. The gray-scale image is duplicated over the three color channels, i.e.$~\bs{G}(i, j) \in [0,255]^3$ has the same value in each element. See \Figure{fig:prediction_demo} for an example with $\lambda = 1$.  

\textbf{Pixel-swap transformation (texture distortion).}
For a pixel-swap proportion $p$ we randomly sample $pK$ pixel positions without replacement from $\mathcal{I}_c$ and randomly permute the pixel values at these positions. Other pixels are left unchanged here as well. See \Figure{fig:prediction_demo} for an example with $p=1$.

\section{Experimental Results}
\begin{figure*}[t]
     \centering
     \begin{subfigure}{0.19\textwidth}
         \centering
         \includegraphics[width=\textwidth]{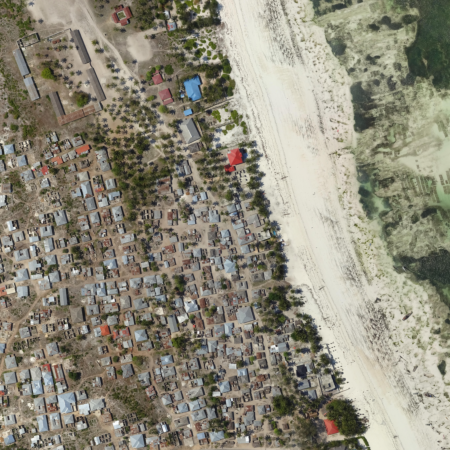}
         \label{fig:zanzibar_pixelswap_range_b4_0}
     \end{subfigure}
     \hspace{-0.15cm} 
     \begin{subfigure}{0.19\textwidth}
         \centering
         \includegraphics[width=\textwidth]{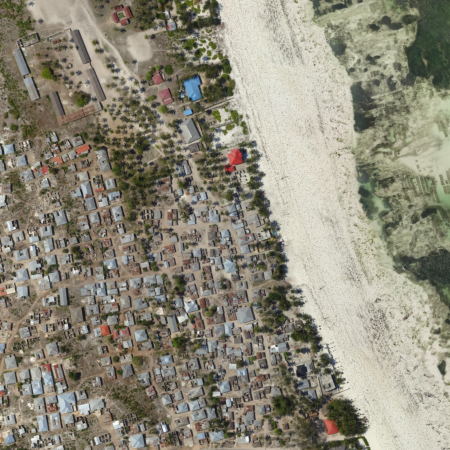}
         \label{fig:zanzibar_pixelswap_range_b4_0.33}
     \end{subfigure}
     \hspace{-0.15cm} 
     \begin{subfigure}{0.19\textwidth}
         \centering
         \includegraphics[width=\textwidth]{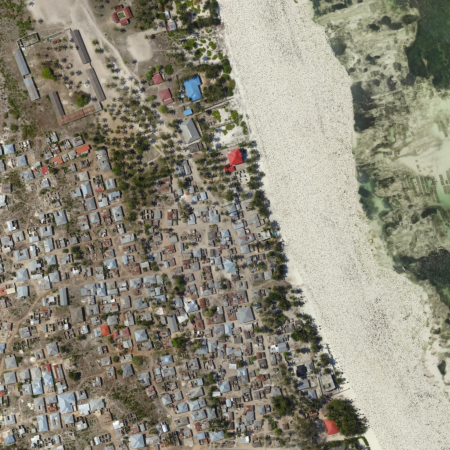}
         \label{fig:zanzibar_pixelswap_range_b4_0.66}
     \end{subfigure}
     \hspace{-0.15cm} 
     \begin{subfigure}{0.19\textwidth}
         \centering
         \includegraphics[width=\textwidth]{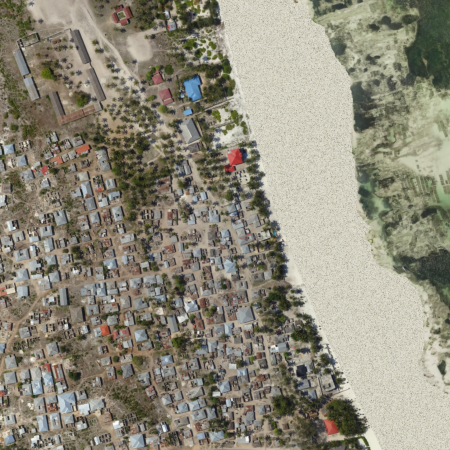}
         \label{fig:zanzibar_pixelswap_range_b4_1}
     \end{subfigure}
     \\ \vspace{-0.4cm} 
     \begin{subfigure}{0.19\textwidth}
         \centering
         \includegraphics[width=\textwidth]{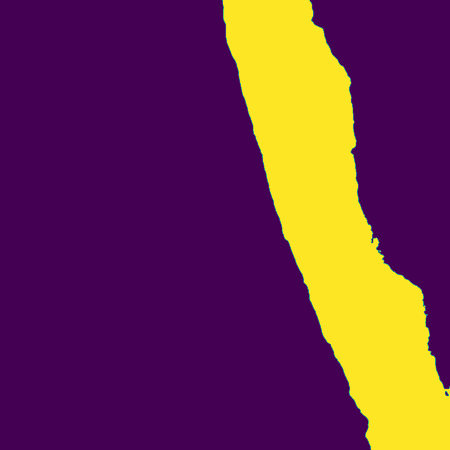}
         \label{fig:zanzibar_pixelswap_range_b4_0_pred}
     \end{subfigure}
     \hspace{-0.15cm} 
     \begin{subfigure}{0.19\textwidth}
         \centering
         \includegraphics[width=\textwidth]{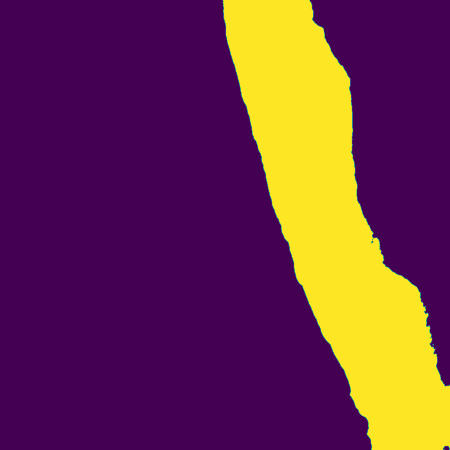}
         \label{fig:zanzibar_pixelswap_range_b4_0.33_pred}
     \end{subfigure}
     \hspace{-0.15cm} 
     \begin{subfigure}{0.19\textwidth}
         \centering
         \includegraphics[width=\textwidth]{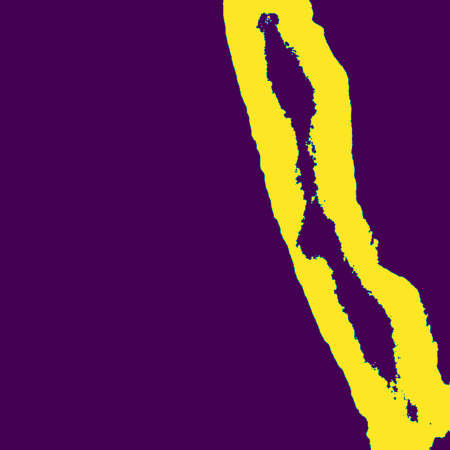}
         \label{fig:zanzibar_pixelswap_range_b4_0.66_pred}
     \end{subfigure}
     \hspace{-0.15cm} 
     \begin{subfigure}{0.19\textwidth}
         \centering
         \includegraphics[width=\textwidth]{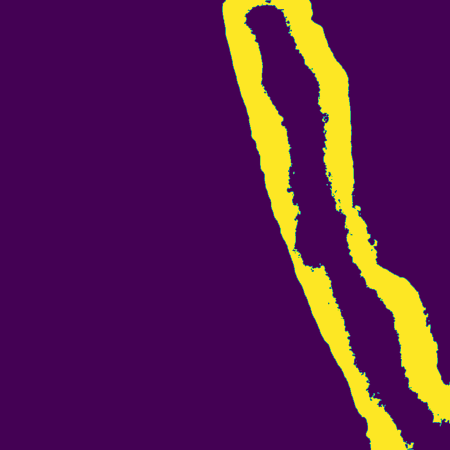}
         \label{fig:zanzibar_pixelswap_range_b4_1_pred}
     \end{subfigure}
     \\ \vspace{-0.4cm} 
     \begin{subfigure}{0.19\textwidth}
         \centering
         \includegraphics[width=\textwidth]{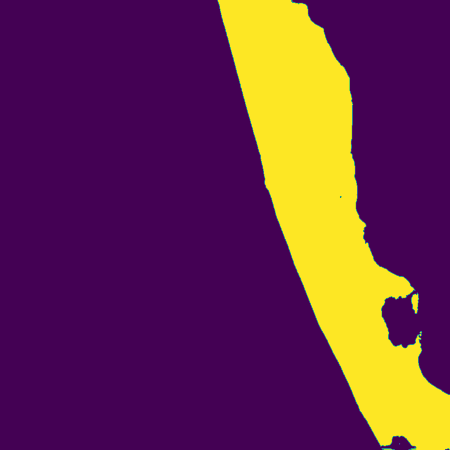}
         \label{fig:zanzibar_pixelswap_range_b4_0_predvv}
     \end{subfigure}
     \hspace{-0.15cm} 
     \begin{subfigure}{0.19\textwidth}
         \centering
         \includegraphics[width=\textwidth]{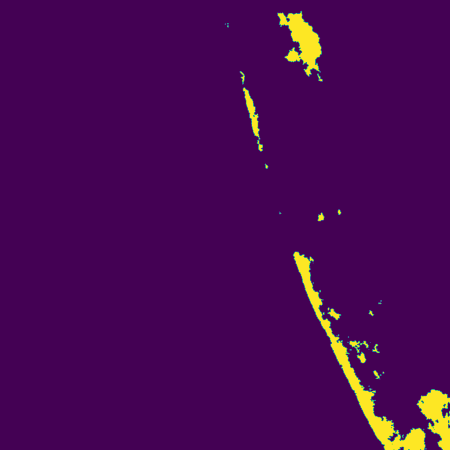}
         \label{fig:zanzibar_pixelswap_range_b4_0.33_predvv}
     \end{subfigure}
     \hspace{-0.15cm} 
     \begin{subfigure}{0.19\textwidth}
         \centering
         \includegraphics[width=\textwidth]{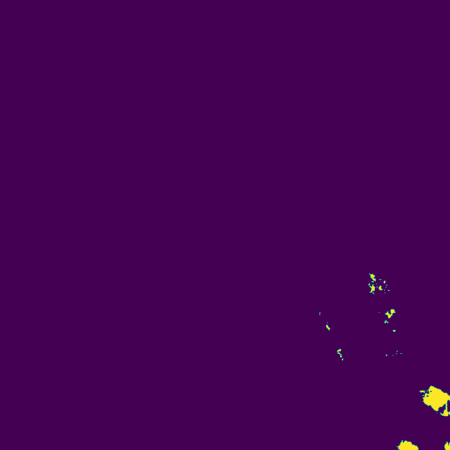}
         \label{fig:zanzibar_pixelswap_range_b4_0.66_predvv}
     \end{subfigure}
     \hspace{-0.15cm} 
     \begin{subfigure}{0.19\textwidth}
         \centering
         \includegraphics[width=\textwidth]{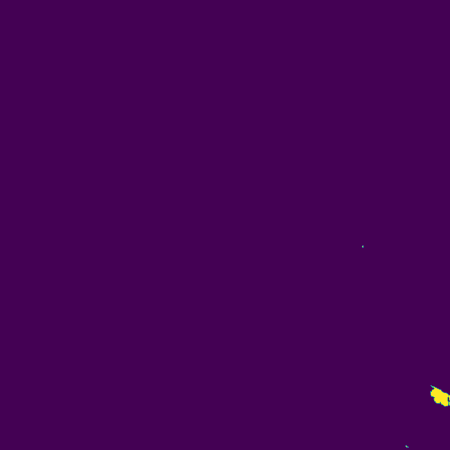}
         \label{fig:zanzibar_pixelswap_range_b4_1_predvv}
     \end{subfigure}

     \vspace{-12pt}
     \caption{\textbf{Top:} Zanzibar region, \textbf{pixel-swap} transformation on the \textit{bare} class with proportion $p$ swapped, where $p \in \{0, 0.33, 0.66, 1\}$ (left to right). \textbf{Middle:} Corresponding model predictions, where yellow and dark blue respectively show pixels predicted as \emph{bare} and \emph{not bare}. The border region remains correctly classified regardless of transformation intensity, so the surrounding context is critical. \textbf{Bottom:} Same as middle, but predictions obtained from the same images and distortions as in the first row, but where all pixels except \emph{bare} ones have been masked out in the images by replacing them with the per-channel mean of the training set (more such results are in the appendix). The importance of context is clear. Predictions made using the U-Net-Efficientnet-B4 model.}
     \label{fig:borderbias}
     \vspace{-5pt}
\end{figure*}

\Figure{fig:main_result_plot} demonstrates the impact of the gray-scale and pixel-swap distortions (at varying intensities, see $x$-axes) on test time predictions on unseen validation data. Recall that the models were trained without any such distortions. We see that for all models, but particularly for U-Net-Efficientnet-B4 (left), there is quite a strong invariance to color distortions (top row), i.e.~even though the images are severely altered and moved far in image space, performance does not suffer much in general. For some classes however (e.g.~\emph{range}), performance deteriorates as the images become fully gray-scale transformed. We also see that although the models have similar mIoUs (black curves) under no gray-scale transformation, and similar degradation trends, the transformer-based model (right in \Figure{fig:main_result_plot}) seems to be the most sensitive to color distortions.

As for texture distortion (pixel-swaps, see bottom row) we see more rapid performance drops in general. 
For smaller distortions, e.g.~pixel-swap proportions of 0.1-0.3, we see performance degradations for several classes and on average for all three models. At these lower distortion proportions it is difficult for the human eye to notice any changes, as can be seen in the examples in the appendix. However, note that even though the models are shown to be sensitive to texture distortions, none of the IoUs approach zero as the intensity of the transformation grows. In \Figure{fig:borderbias} we show how the model prediction for the \textit{range} class for one of the images changes as the intensity of the pixel-swap transformation increases. Note how the interior of the region associated with the class becomes increasingly misclassified, while the border remains correctly classified. This coupled with the third row of \Figure{fig:borderbias} clearly indicates that the network leverages surrounding context for its predictions. \emph{We refer the reader to the appendix for more quantitative and qualitative results.} 

Finally, we note that \emph{range} and \emph{tree} are among the classes that are most affected by degradations -- be them color- or texture-based -- even though they are among the most common classes in the training set (measured in total number of pixels). Thus, somewhat counter-intuitively, the robustness to distortions for a given class is not related to the relative amount of training data for said class.

\textbf{More on the importance of surrounding context.} We here further examine how model predictions for a given class are affected by context from neighboring pixels of other classes (beyond the qualitative example in \Figure{fig:borderbias}, more of which are found in the appendix). As a basis for this investigation we use the pixel-swap transformation, but here we also replace all pixel values \emph{not} related to the class under investigation with the per-channel mean of the training data set, in order to remove all surrounding context. See \Figure{fig:borderbias_zeron} and \Figure{appfig:pixelswap_zeron_bigdemo} in the appendix for two qualitative examples. We see that removing context affects model predictions significantly, and in particular that the accuracy deteriorates much faster as the pixel-swap proportion $p$ increases, compared to the setting where the context is kept intact. In \Figure{fig:borderbias_zeron} we see that the border predictions also fail when context is absent, while in \Figure{appfig:pixelswap_zeron_bigdemo} we see a significant drop in prediction accuracy even at $p=0$.

\begin{figure*}[t]
     \centering
    \begin{subfigure}{0.32\textwidth}
         \centering
         \includegraphics[width=\textwidth]{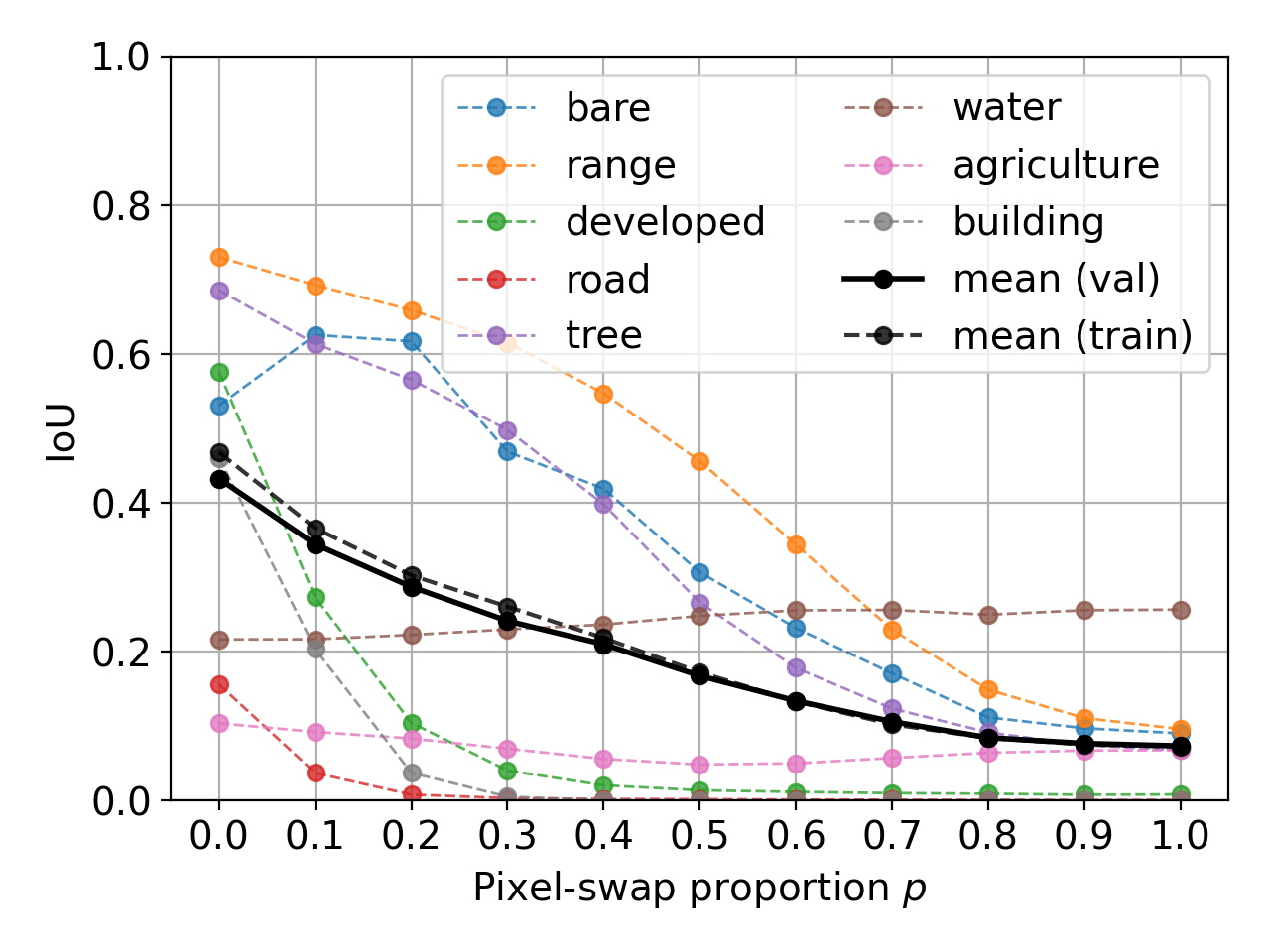}
         \label{fig:pixelswap_zeron_val_b4}
     \end{subfigure}
     \begin{subfigure}{0.32\textwidth}
         \centering
         \includegraphics[width=\textwidth]{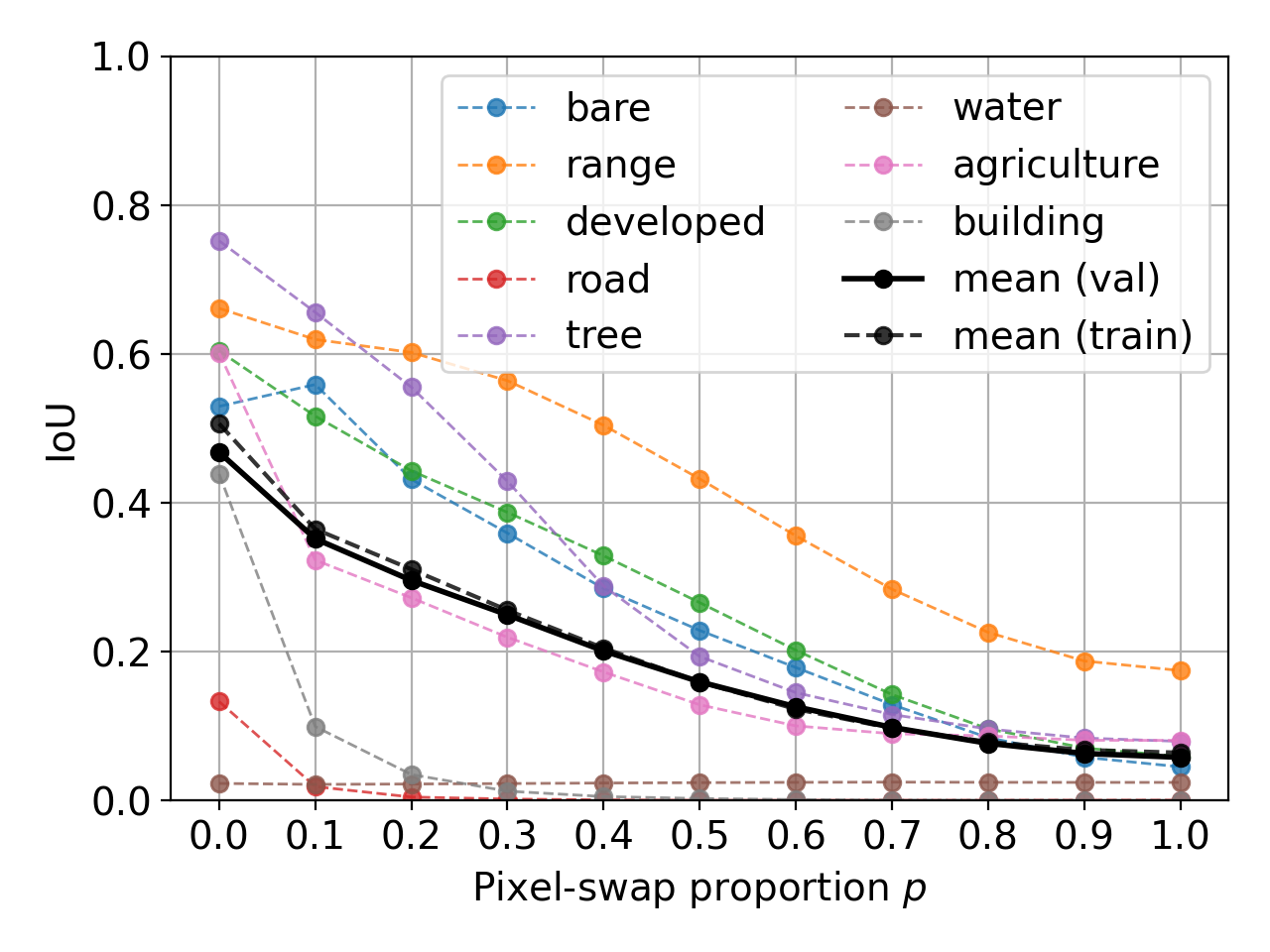}
         \label{fig:pixelswap_zeron_val_dl3}
     \end{subfigure}
     \begin{subfigure}{0.32\textwidth}
         \centering
         \includegraphics[width=\textwidth]{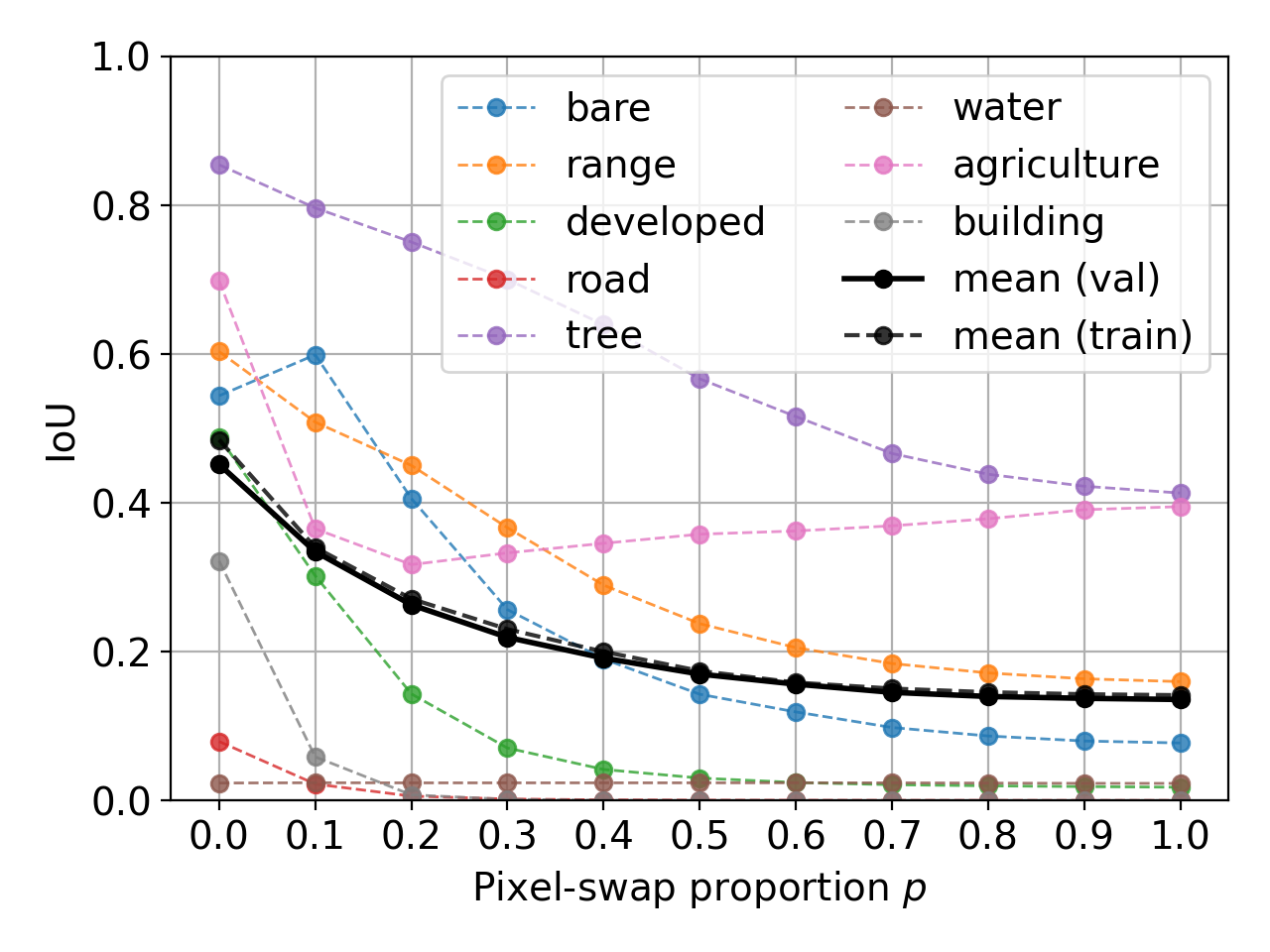}
         \label{fig:pixelswap_zeron_val_ftunet}
     \end{subfigure}
     \vspace{-13pt}
     \caption{Impact of the \textbf{pixel-swap} transformation where all pixels except for the class under investigation are replaced by the per-channel mean of the training set. Results are shown for the three segmentation models outlined in \Section{sec:method}. From left to right: U-Net-Efficientnet-B4, DeeplabV3-Resnet50, and FTUNetFormer. The solid black curve is the mean of the colored curves (validation data), and the dashed black curve is the corresponding mean on training data (included for comparison). Models perform significantly worse in general, compared to the case where context is kept intact (see also \Figure{fig:mainresult_borderbias_zeron_gap}).}
     \label{fig:mainresult_borderbias_zeron}
\end{figure*}

\begin{figure*}[t]
     \centering
    \begin{subfigure}{0.32\textwidth}
         \centering
         \includegraphics[width=\textwidth]{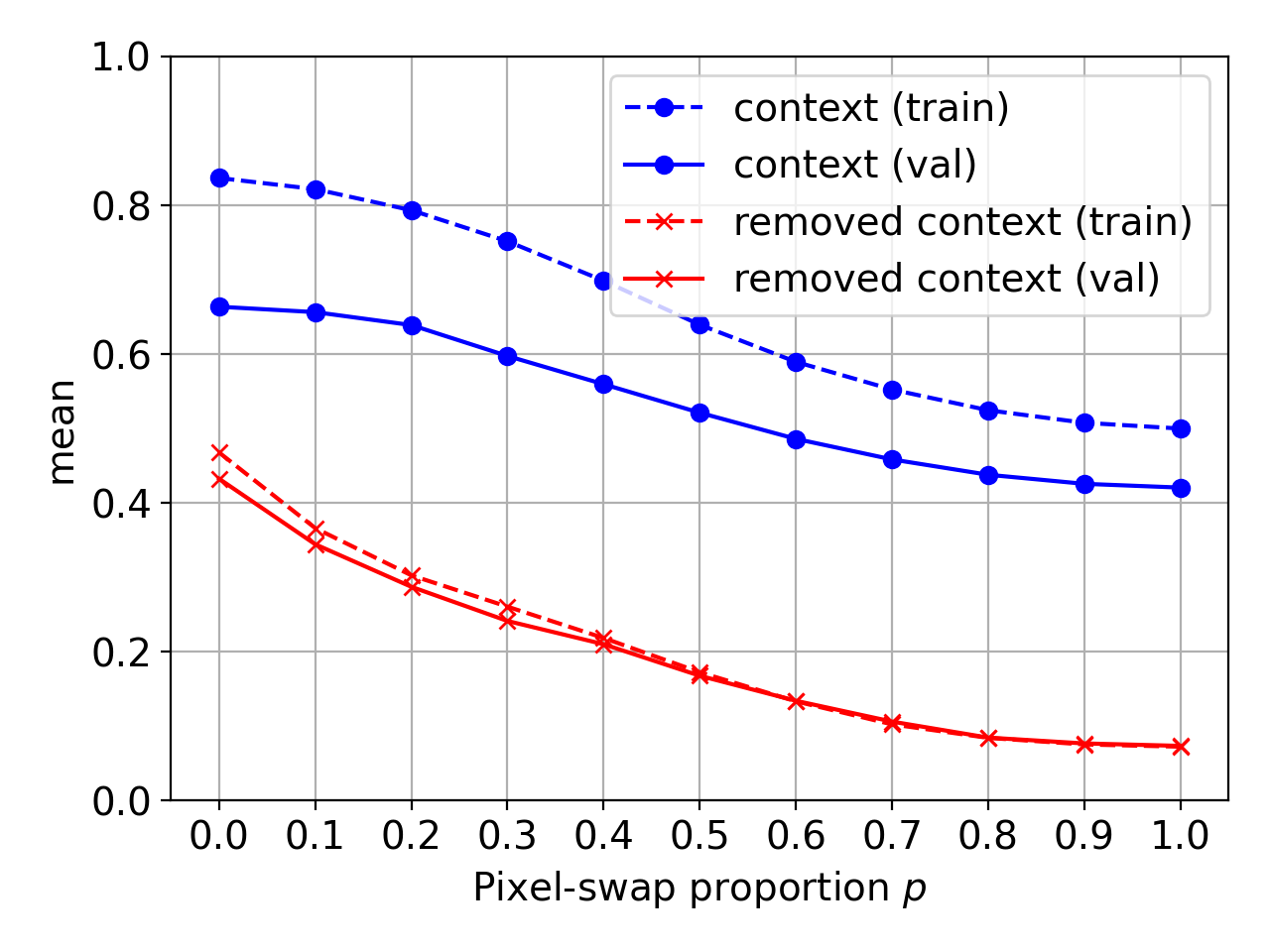}
         \label{fig:gap_pixelswap_zeron_val_b4}
     \end{subfigure}
     \begin{subfigure}{0.32\textwidth}
         \centering
         \includegraphics[width=\textwidth]{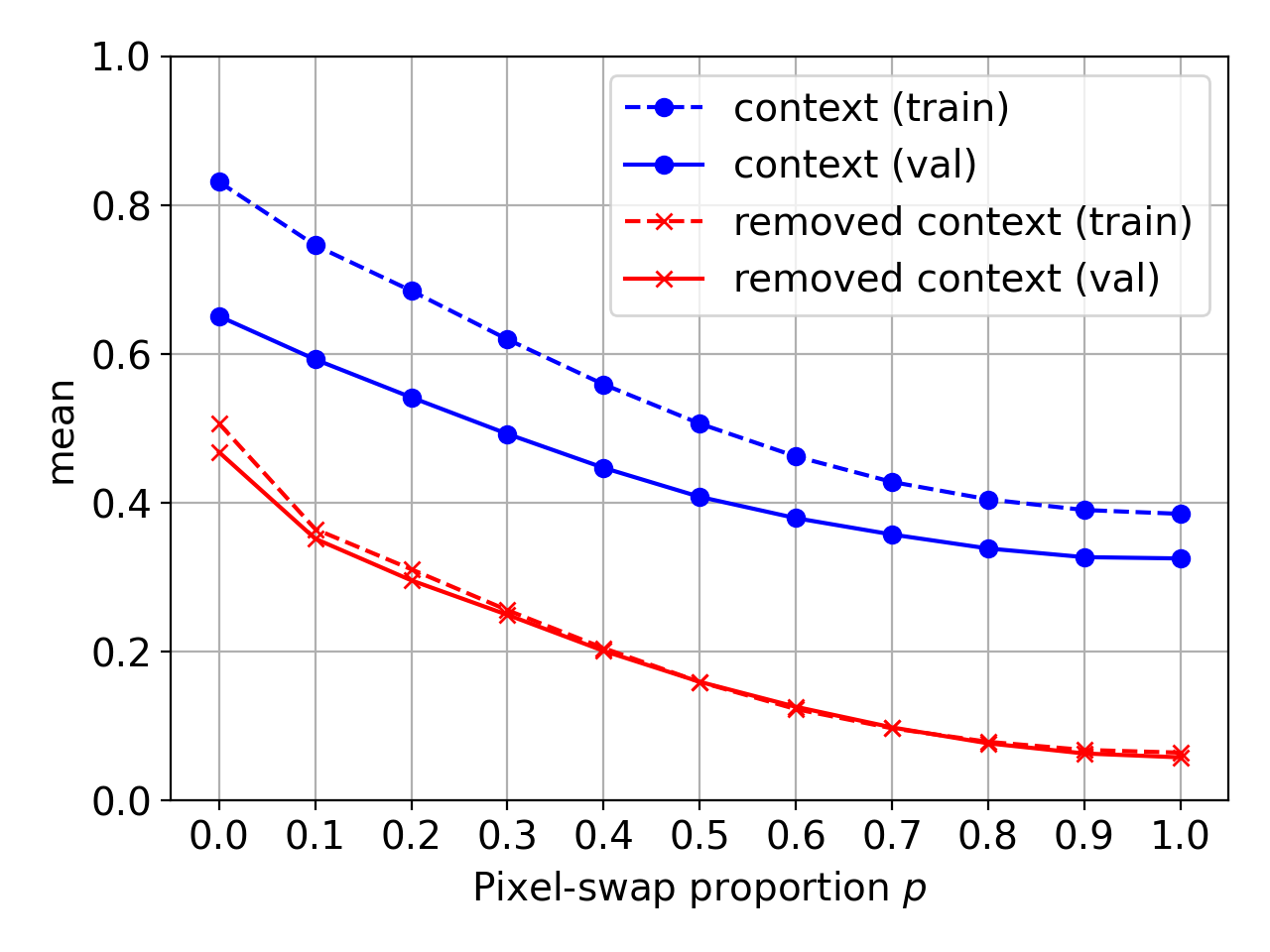}
         \label{fig:gap_pixelswap_zeron_val_dl3}
     \end{subfigure}
     \begin{subfigure}{0.32\textwidth}
         \centering
         \includegraphics[width=\textwidth]{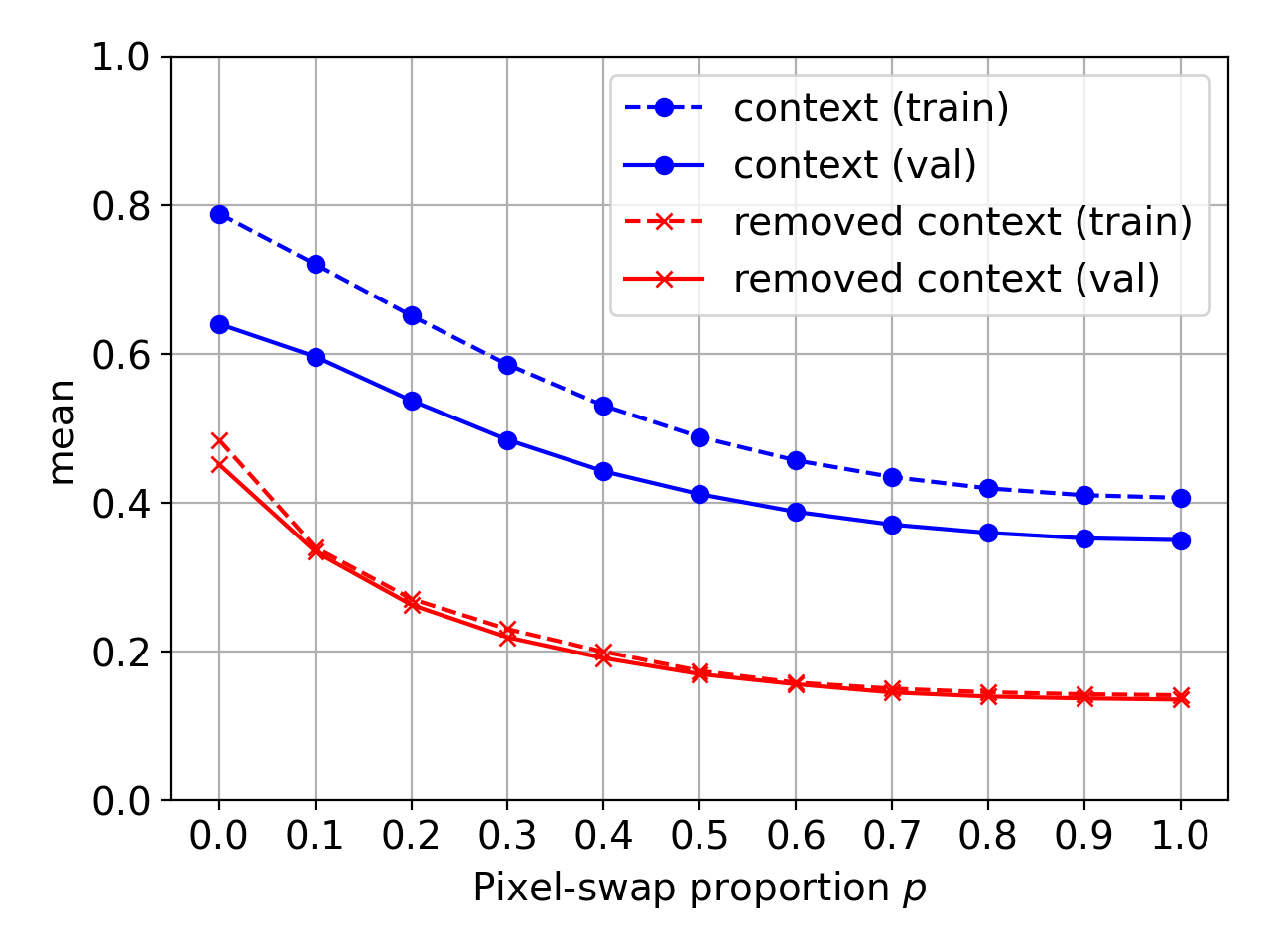}
         \label{fig:gap_pixelswap_zeron_val_ftunet}
     \end{subfigure}
     \vspace{-13pt}
     \caption{Comparison of keeping context intact (blue) and removing context (red), with various proportions of the \textbf{pixel-swap} transformation. The blue curves are identical to the means (in black) of \Figure{fig:main_result_plot}. The red curves are identical to the means (in black) of \Figure{fig:mainresult_borderbias_zeron}. From left to right: U-Net-Efficientnet-B4, DeeplabV3-Resnet50, and FTUNetFormer. There is a significant performance drop at all proportions $p$, even when no pixel-swap is applied ($p = 0$), and the difference between the training and validation set is vastly smaller when surrounding context is removed.}
     \label{fig:mainresult_borderbias_zeron_gap}
\end{figure*}

In \Figure{fig:mainresult_borderbias_zeron}, the effect of context removal is quantitatively examined on the whole validation set (including also average results on the training set, for comparison). Comparing with \Figure{fig:main_result_plot}, we see that removing context yields a significant performance drop in general. There is also a smaller gap in performance between the training and validation sets when context is removed, even for $p=0$ (i.e.~with no pixel-swap applied). This suggests that removing surrounding information inhibits the model from making correct predictions for the class of interest, whether or not this is data on which the model has been trained. To make these comparisons easier, see also  \Figure{fig:mainresult_borderbias_zeron_gap}.


\section{Conclusions}
In this paper we have investigated the impacts of several test time color and texture distortions on EO imagery. Our experiments -- conducted using popular CNN- and transformer-based models -- suggest that deep networks, which have not been exposed to these distortions in training, are relatively robust to EO image color distortions
but are sensitive to texture distortions. Further, our experiments indicate that models use the surrounding context when making predictions and are sensitive to changes in this context. 
These empirical findings, while intriguing on their own, also point to many future areas of potential research and improvements regarding land cover classification and EO tasks more broadly.
For example, since our results suggest that there is a large variation in how different classes are affected by the various distortions, it may be possible to leverage these insights to develop effective class-dependent data augmentation techniques in the EO domain.

\clearpage
\bibliography{iclr2024_conference}

\begin{thebibliography}{12}
\providecommand{\natexlab}[1]{#1}
\providecommand{\url}[1]{\texttt{#1}}
\expandafter\ifx\csname urlstyle\endcsname\relax
  \providecommand{\doi}[1]{doi: #1}\else
  \providecommand{\doi}{doi: \begingroup \urlstyle{rm}\Url}\fi

\bibitem[Dodge \& Karam(2016)Dodge and Karam]{dodge2016understanding}
Samuel Dodge and Lina Karam.
\newblock Understanding how image quality affects deep neural networks.
\newblock In \emph{2016 eighth international conference on quality of multimedia experience (QoMEX)}, pp.\  1--6. IEEE, 2016.

\bibitem[Florian \& Adam(2017)Florian and Adam]{florian2017rethinking}
L-CCGP Florian and Schroff~Hartwig Adam.
\newblock Rethinking atrous convolution for semantic image segmentation.
\newblock In \emph{Conference on computer vision and pattern recognition (CVPR). IEEE/CVF}, volume~6, 2017.

\bibitem[Gong et~al.(2024)Gong, Li, Chen, and Jiang]{gong2024exploring}
Yunpeng Gong, Jiaquan Li, Lifei Chen, and Min Jiang.
\newblock Exploring color invariance through image-level ensemble learning.
\newblock \emph{arXiv preprint arXiv:2401.10512}, 2024.

\bibitem[Hermann et~al.(2020)Hermann, Chen, and Kornblith]{conv_texture_bias}
Katherine Hermann, Ting Chen, and Simon Kornblith.
\newblock The origins and prevalence of texture bias in convolutional neural networks.
\newblock In H.~Larochelle, M.~Ranzato, R.~Hadsell, M.F. Balcan, and H.~Lin (eds.), \emph{Advances in Neural Information Processing Systems}, volume~33, pp.\  19000--19015. Curran Associates, Inc., 2020.
\newblock URL \url{https://proceedings.neurips.cc/paper_files/paper/2020/file/db5f9f42a7157abe65bb145000b5871a-Paper.pdf}.

\bibitem[Iakubovskii(2019)]{Iakubovskii:2019}
Pavel Iakubovskii.
\newblock Segmentation models pytorch.
\newblock \url{https://github.com/qubvel/segmentation_models.pytorch}, 2019.

\bibitem[Kingma \& Ba(2014)Kingma and Ba]{kingma2014adam}
Diederik~P Kingma and Jimmy Ba.
\newblock Adam: A method for stochastic optimization.
\newblock \emph{arXiv preprint arXiv:1412.6980}, 2014.

\bibitem[Rolf et~al.(2024)Rolf, Klemmer, Robinson, and Kerner]{rolf2024mission}
Esther Rolf, Konstantin Klemmer, Caleb Robinson, and Hannah Kerner.
\newblock Mission critical--satellite data is a distinct modality in machine learning.
\newblock \emph{arXiv preprint arXiv:2402.01444}, 2024.

\bibitem[Scikit-image()]{skimage-rgb2gray}
Scikit-image.
\newblock rgb2gray.
\newblock \url{https://scikit-image.org/docs/stable/auto_examples/color_exposure/plot_rgb_to_gray.html}.

\bibitem[Wang et~al.(2022)Wang, Li, Zhang, Fang, Duan, Meng, and Atkinson]{wang2022unetformer}
Libo Wang, Rui Li, Ce~Zhang, Shenghui Fang, Chenxi Duan, Xiaoliang Meng, and Peter~M Atkinson.
\newblock Unetformer: A unet-like transformer for efficient semantic segmentation of remote sensing urban scene imagery.
\newblock \emph{ISPRS Journal of Photogrammetry and Remote Sensing}, 190:\penalty0 196--214, 2022.

\bibitem[Xia et~al.(2023)Xia, Yokoya, Adriano, and Broni-Bediako]{xia2023openearthmap}
Junshi Xia, Naoto Yokoya, Bruno Adriano, and Clifford Broni-Bediako.
\newblock Openearthmap: A benchmark dataset for global high-resolution land cover mapping.
\newblock In \emph{Proceedings of the IEEE/CVF Winter Conference on Applications of Computer Vision}, pp.\  6254--6264, 2023.

\bibitem[Zhao et~al.(2023)Zhao, Tu, Ye, Tang, Hu, and Xie]{zhao2023land}
Shengyu Zhao, Kaiwen Tu, Shutong Ye, Hao Tang, Yaocong Hu, and Chao Xie.
\newblock Land use and land cover classification meets deep learning: A review.
\newblock \emph{Sensors}, 23\penalty0 (21):\penalty0 8966, 2023.

\bibitem[Zhou et~al.(2017)Zhou, Song, and Cheung]{zhou2017classification}
Yiren Zhou, Sibo Song, and Ngai-Man Cheung.
\newblock On classification of distorted images with deep convolutional neural networks.
\newblock In \emph{2017 IEEE International Conference on Acoustics, Speech and Signal Processing (ICASSP)}, pp.\  1213--1217. IEEE, 2017.

\end{thebibliography}
\bibliographystyle{iclr2024_conference}

\newpage
\appendix
\begin{center}
    \LARGE{Appendix: Impacts of Color and Texture Distortions on
Earth Observation Data in Deep Learning}
\end{center}

In this appendix we provide several additional results (both quantitative and qualitative) for the image distortions introduced in the main paper, as well as for additional distortions defined in this appendix. We also provide more details regarding our experimental setup.

\textbf{Additional experimental setup details.} For convenience, before expanding with additional details, we first attach here some of the implementation details that were already explained in the main paper. During training, we apply only horizontal and vertical flips (an independent 50\% probability for each) as data augmentations. \emph{Thus note that we do not apply any of the color- or texture-based distortions during training.} All three models (see \S 3 in the main paper) are trained and evaluated on images resized to $512 \times 512$ or $1024 \times 1024$, however during training we sample random crops of size $512 \times 512$ from these images, for 2,000 epochs with a batch size of 10, and the best ones are selected based on the validation mIoU. We use a standard cross-entropy loss and the Adam optimizer \citep{kingma2014adam} with learning rate $0.0002$, $\beta_1 = 0.5$, and $\beta_2 = 0.999$. We ignore the background
class; it constitutes $\sim0.6$\% of all pixels and is also ignored in the official OpenEarthMap benchmark.

\begin{figure*}[b]
     \centering
    \begin{subfigure}{0.32\textwidth}
         \centering
         \includegraphics[width=\textwidth]{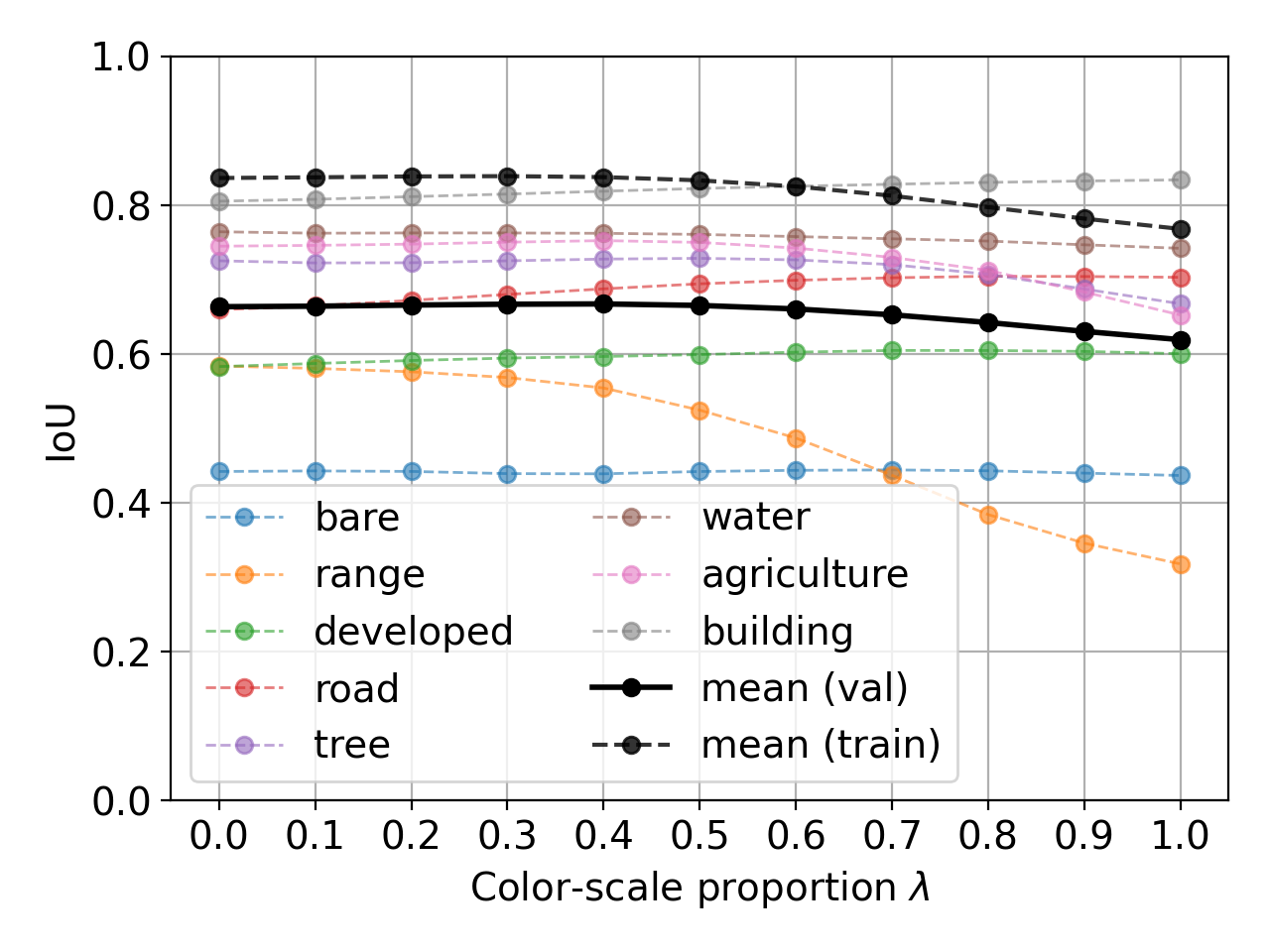}
         \label{fig:colorduplication_R_val_b4}
     \end{subfigure}
     \begin{subfigure}{0.32\textwidth}
         \centering
         \includegraphics[width=\textwidth]{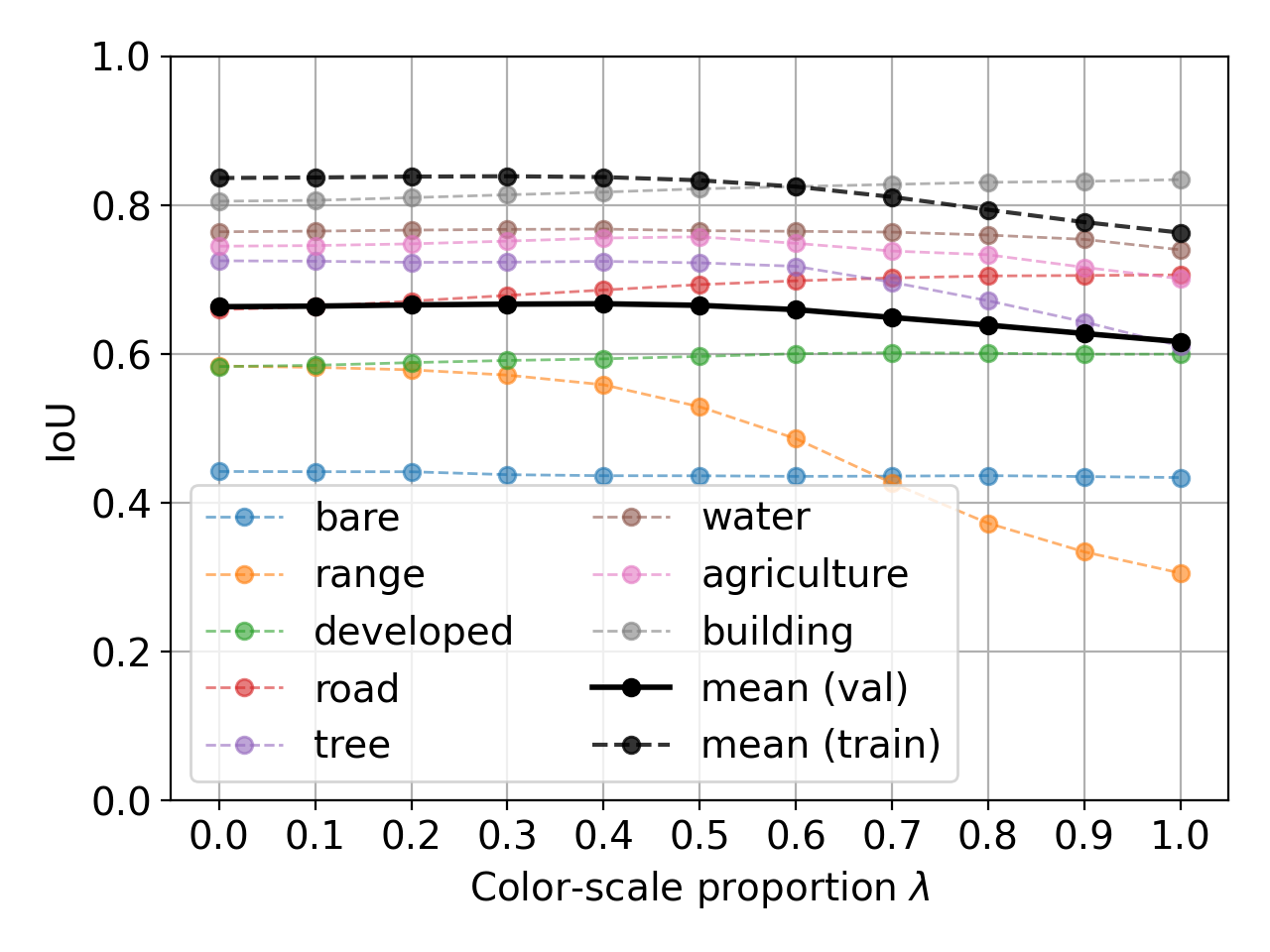}
         \label{fig:colorduplication_G_val_b4}
     \end{subfigure}
     \begin{subfigure}{0.32\textwidth}
         \centering
         \includegraphics[width=\textwidth]{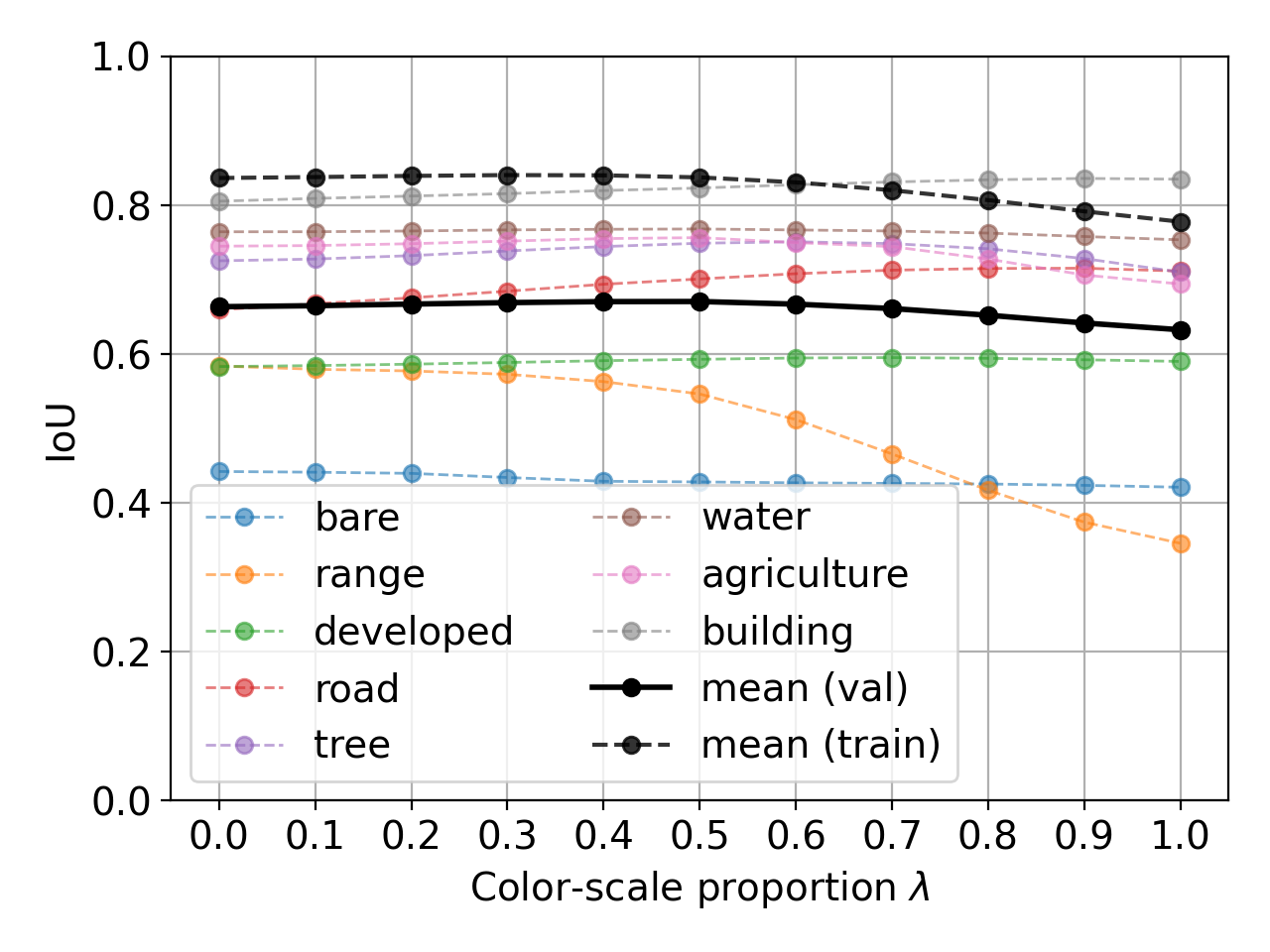}
         \label{fig:colorduplication_B_val_b4}
     \end{subfigure}
     \vspace{-13pt}
     \caption{Impact of the \textbf{color-duplication} transformations for the U-Net-Efficientnet-B4 model. From left to right: Red-duplication, green-duplication, and blue-duplication. The solid black curve is the mean of the colored curves (validation data), and the dashed black curve is the corresponding mean on training data (included for comparison). The performance degradation trends are similar for all three transformations, and are also comparable to the gray-scale color distortion (see main paper). All taken together, these results suggest that deep networks are quite robust to color distortions on EO imagery.}
     \label{fig:colorpicker}
\end{figure*}

\begin{figure*}[t]
     \centering
     \begin{subfigure}{0.32\textwidth}
         \centering
         \includegraphics[width=\textwidth]{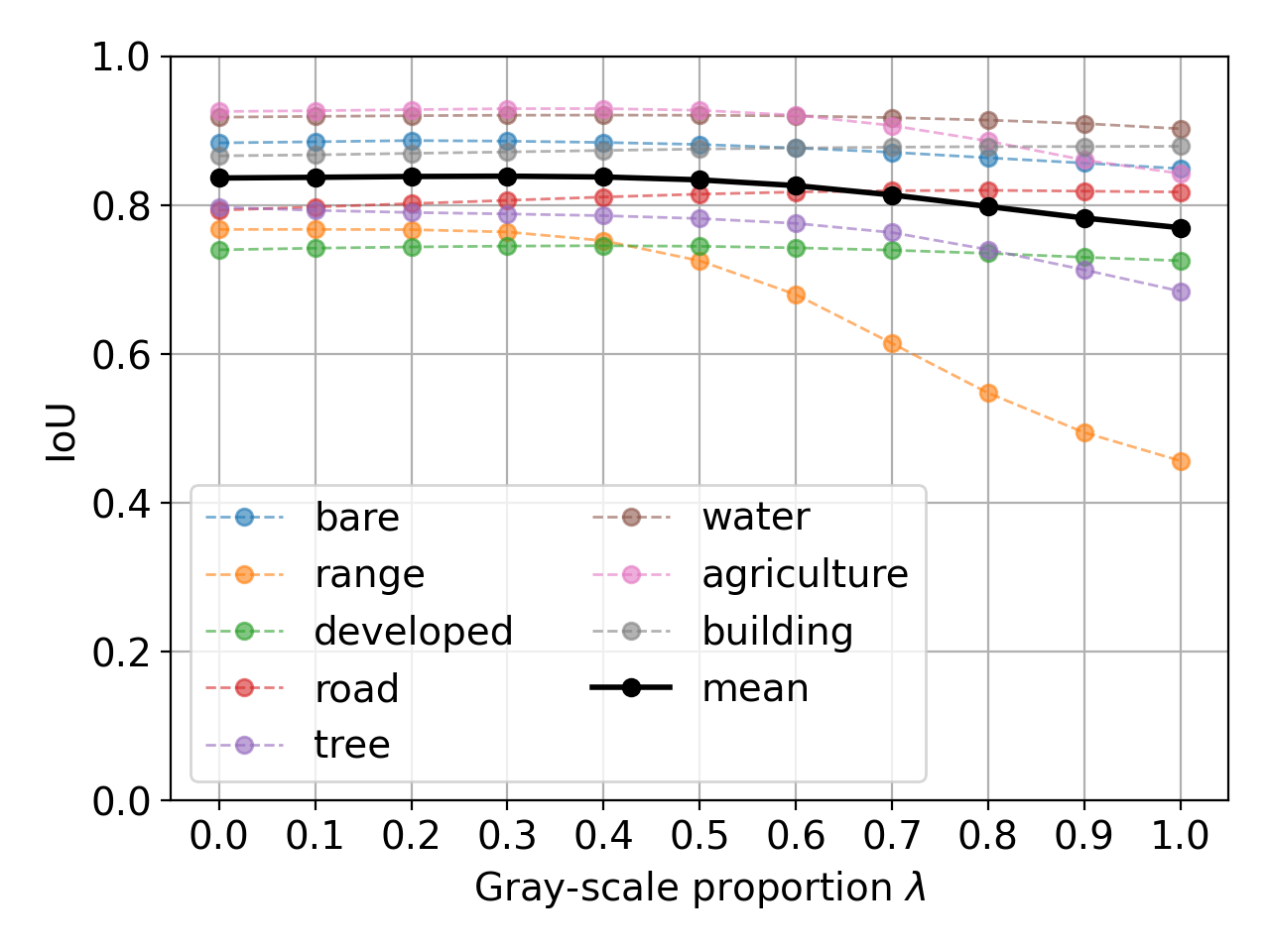}
         \label{fig:grayconvex_train_b4}
     \end{subfigure}
     \hfill
     \begin{subfigure}{0.32\textwidth}
         \centering
         \includegraphics[width=\textwidth]{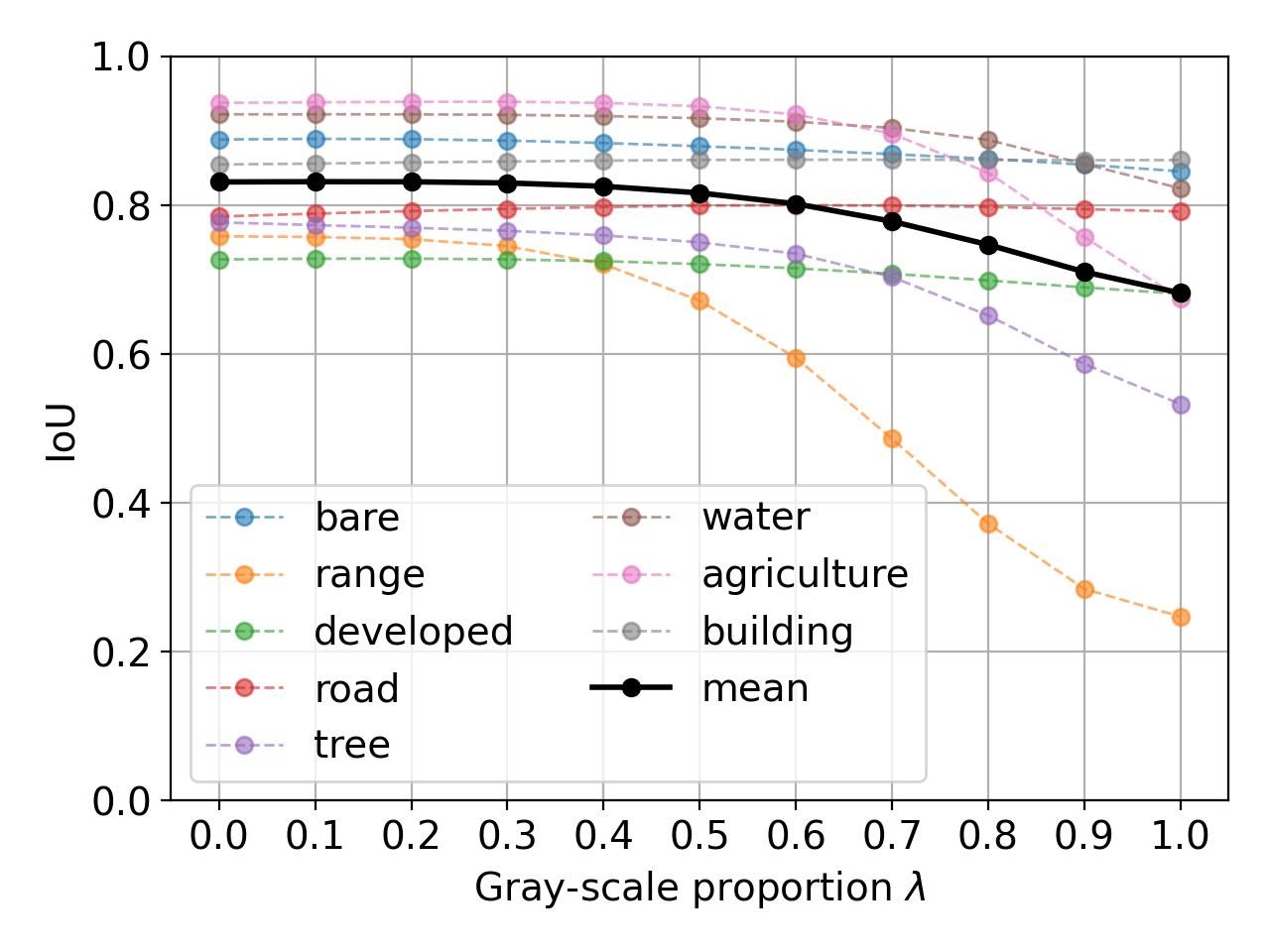}
         \label{fig:grayconvex_train_dl3}
     \end{subfigure}
     \hfill
     \begin{subfigure}{0.32\textwidth}
         \centering
         \includegraphics[width=\textwidth]{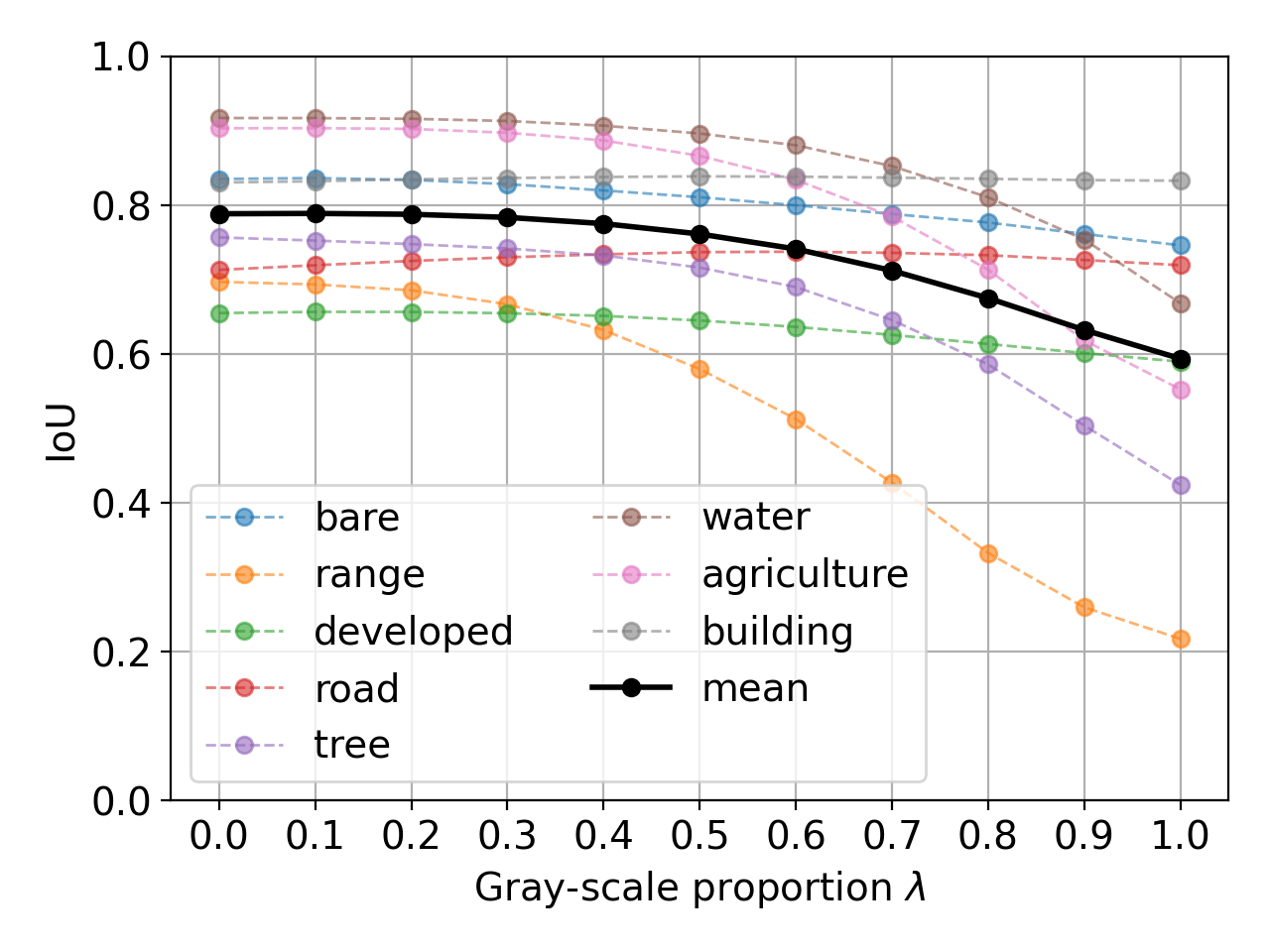}
         \label{fig:grayconvex_train_ftunet}
     \end{subfigure}
     ~
     \begin{subfigure}{0.32\textwidth}
         \centering
         \includegraphics[width=\textwidth]{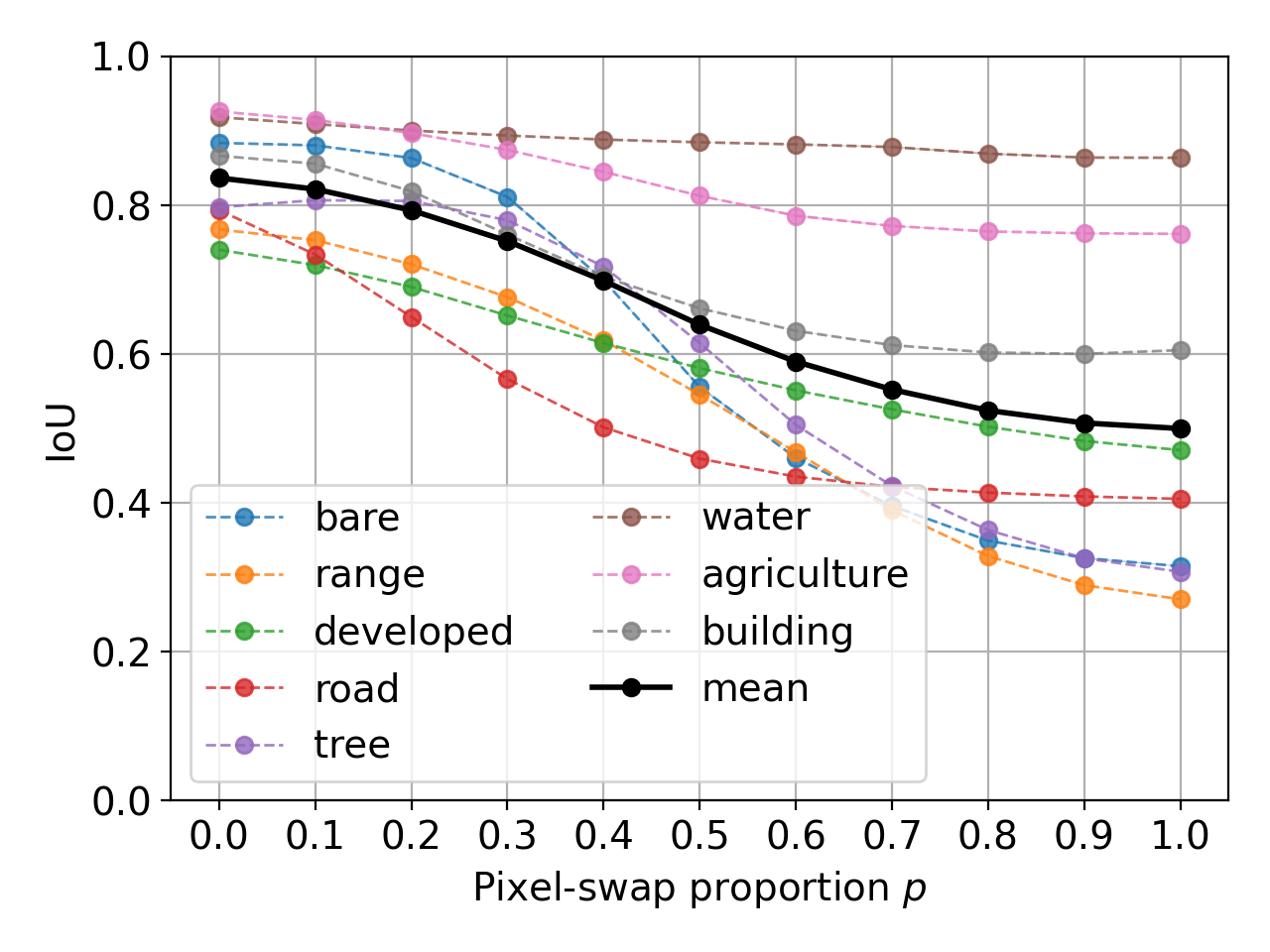}
         \label{fig:pixelswap_train_b4}
     \end{subfigure}
     \hfill
     \begin{subfigure}{0.32\textwidth}
         \centering
         \includegraphics[width=\textwidth]{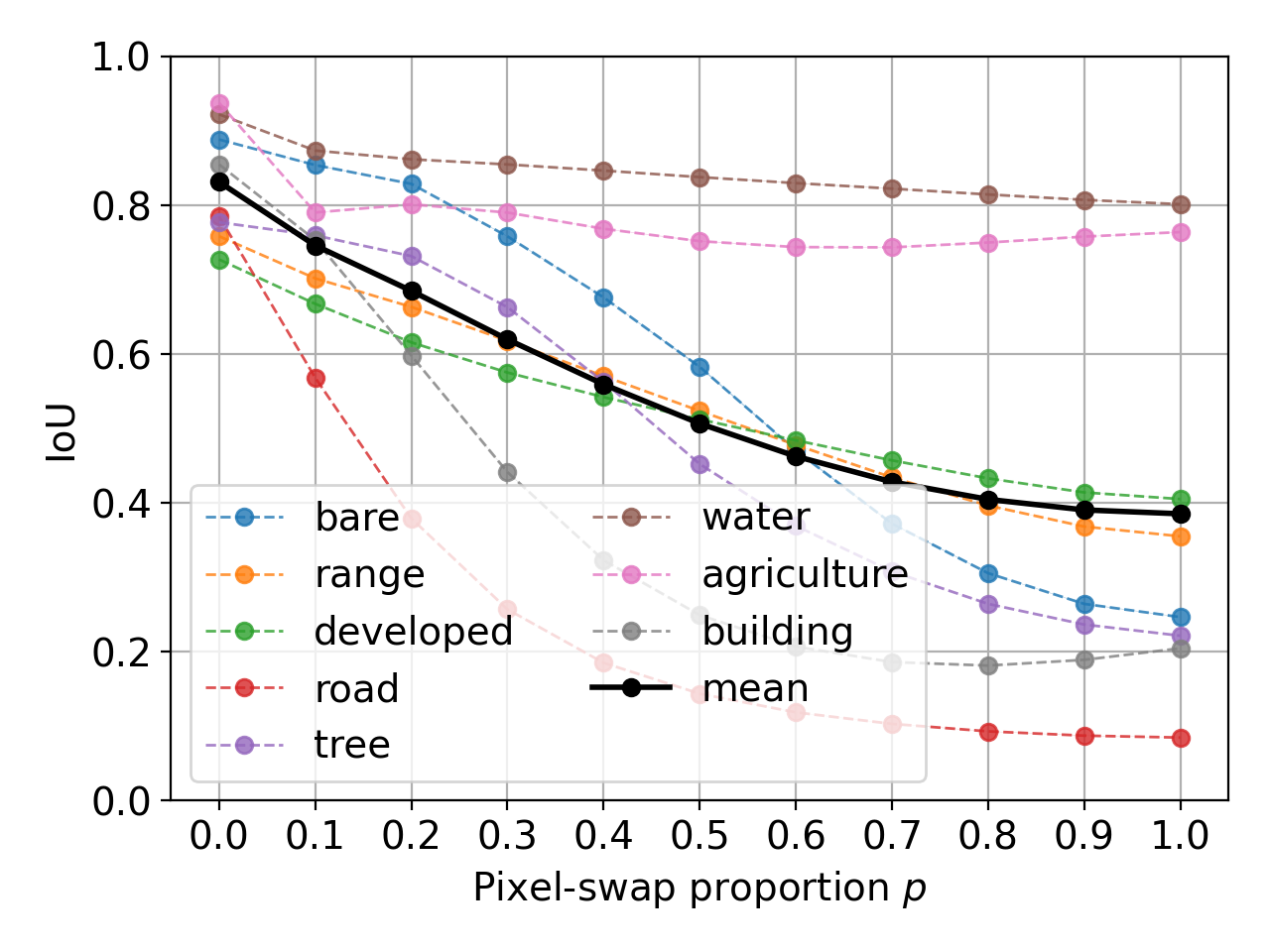}
         \label{fig:pixelswap_train_dl3}
     \end{subfigure}
     \hfill
     \begin{subfigure}{0.32\textwidth}
         \centering
         \includegraphics[width=\textwidth]{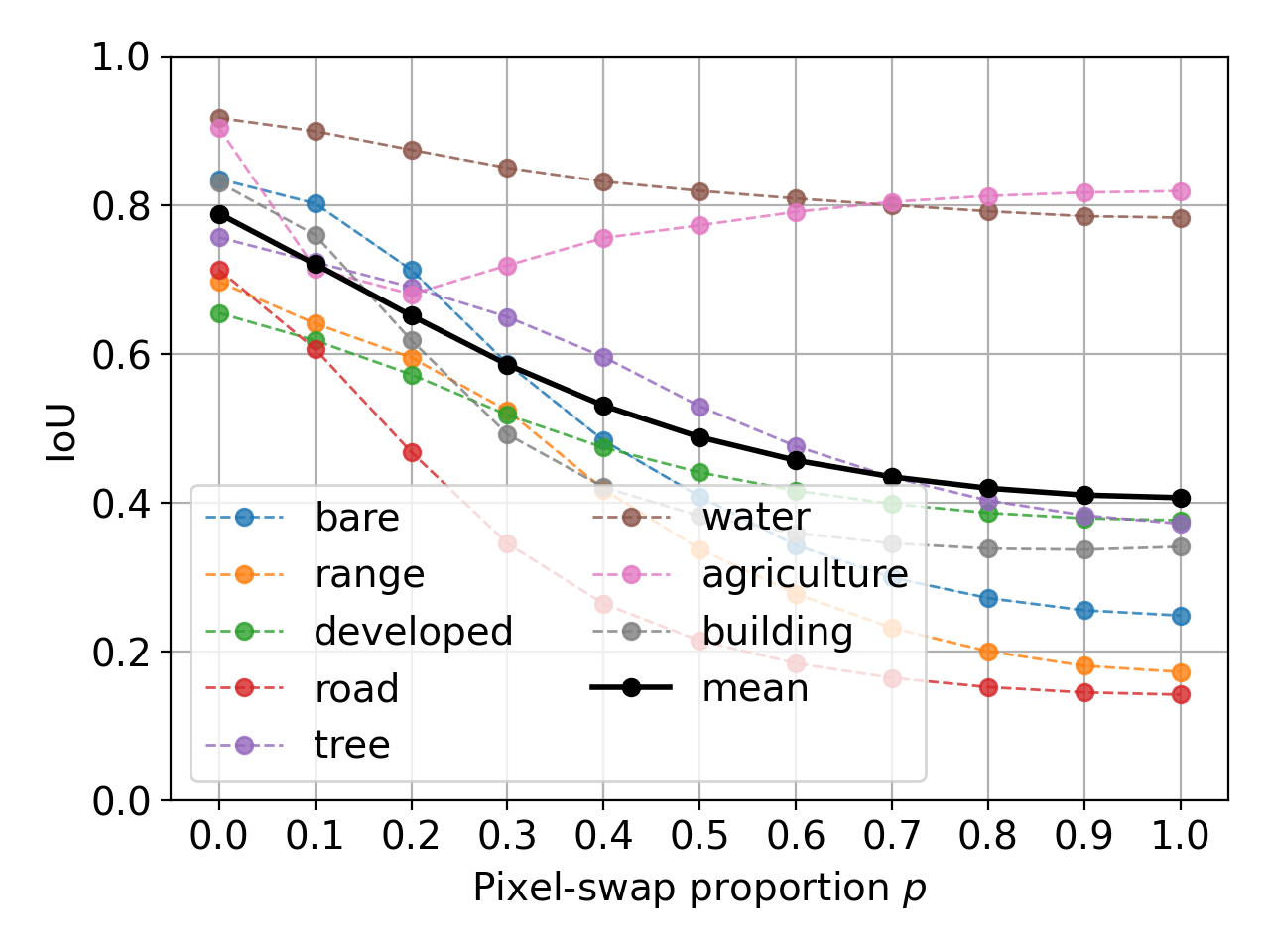}
         \label{fig:pixelswap_train_ftunet}
     \end{subfigure}
      \vspace{-10pt}
     \caption{Impact of the \textbf{gray-scale} (top row) and \textbf{pixel-swap} (bottom row) transformations at test time on the training set for the three segmentation models outlined in \Section{sec:method} (main paper). From left to right: U-Net-Efficientnet-B4, DeeplabV3-Resnet50, and FTUNetFormer. The solid black curve is the mean of the colored curves. As is the case on validation data (cf.~\Figure{fig:main_result_plot} in the main paper), models are generally more sensitive to texture than color distortions on training data as well.}
     \label{fig:main_results_plot_train}
\end{figure*}

\begin{figure*}[h]
     \centering
     \begin{subfigure}{0.24\textwidth}
         \centering
         \includegraphics[width=\textwidth]{iclr2024-ml4rs/fig/pixelswap_zanzibar_0_ps_0_cropped.png}
         \label{fig:zanzibar_pixelswap_bare_b4_0_app}
     \end{subfigure}
     \begin{subfigure}{0.24\textwidth}
         \centering
         \includegraphics[width=\textwidth]{iclr2024-ml4rs/fig/pixelswap_zanzibar_0_ps_0.33_cropped.png}
         \label{fig:zanzibar_pixelswap_bare_b4_0.33_app}
     \end{subfigure}
     \begin{subfigure}{0.24\textwidth}
         \centering
         \includegraphics[width=\textwidth]{iclr2024-ml4rs/fig/pixelswap_zanzibar_0_ps_0.66_cropped.png}
         \label{fig:zanzibar_pixelswap_bare_b4_0.66_app}
     \end{subfigure}
     \begin{subfigure}{0.24\textwidth}
         \centering
         \includegraphics[width=\textwidth]{iclr2024-ml4rs/fig/pixelswap_zanzibar_0_ps_1_cropped.png}
         \label{fig:zanzibar_pixelswap_bare_b4_1_app}
     \end{subfigure}
     \\ \vspace{-0.3cm} 
     \begin{subfigure}{0.24\textwidth}
         \centering
         \includegraphics[width=\textwidth]{iclr2024-ml4rs/fig/pixelswap_zanzibar_0_ps_0_pred_cropped.png}
         \label{fig:zanzibar_pixelswap_bare_b4_0_pred_app}
     \end{subfigure}
     \begin{subfigure}{0.24\textwidth}
         \centering
         \includegraphics[width=\textwidth]{iclr2024-ml4rs/fig/pixelswap_zanzibar_0_ps_0.33_pred_cropped.png}
         \label{fig:zanzibar_pixelswap_bare_b4_0.33_pred_app}
     \end{subfigure}
     \begin{subfigure}{0.24\textwidth}
         \centering
         \includegraphics[width=\textwidth]{iclr2024-ml4rs/fig/pixelswap_zanzibar_0_ps_0.66_pred_cropped.png}
         \label{fig:zanzibar_pixelswap_bare_b4_0.66_pred_app}
     \end{subfigure}
     \begin{subfigure}{0.24\textwidth}
         \centering
         \includegraphics[width=\textwidth]{iclr2024-ml4rs/fig/pixelswap_zanzibar_0_ps_1_pred_cropped.png}
         \label{fig:zanzibar_pixelswap_bare_b4_1_pred_app}
     \end{subfigure}
     \\ \vspace{-0.3cm} 
     \begin{subfigure}{0.24\textwidth}
         \centering
         \includegraphics[width=\textwidth]{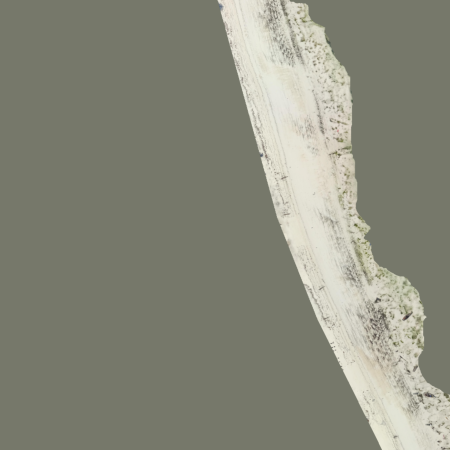}
         \label{fig:zanzibar_pixelswap_bare_zeron_b4_0}
     \end{subfigure}
     \begin{subfigure}{0.24\textwidth}
         \centering
         \includegraphics[width=\textwidth]{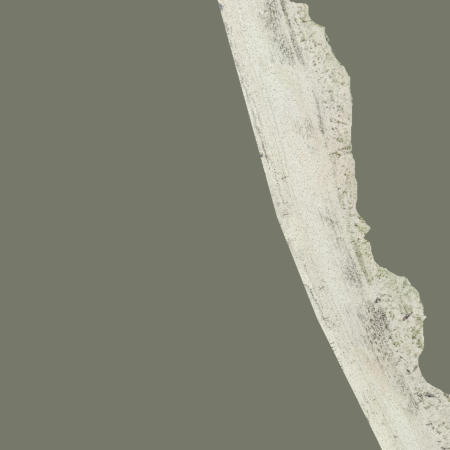}
         \label{fig:zanzibar_pixelswap_bare_zeron_b4_0.33}
     \end{subfigure}
     \begin{subfigure}{0.24\textwidth}
         \centering
         \includegraphics[width=\textwidth]{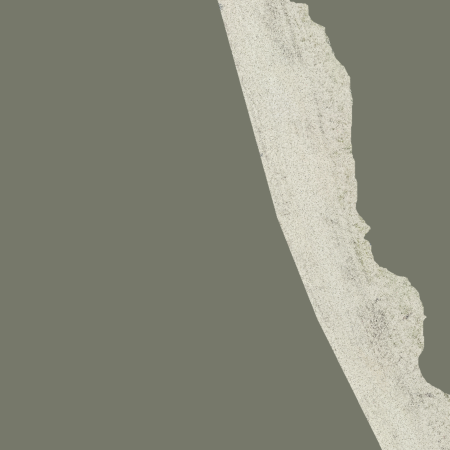}
         \label{fig:zanzibar_pixelswap_bare_zeron_b4_0.66}
     \end{subfigure}
     \begin{subfigure}{0.24\textwidth}
         \centering
         \includegraphics[width=\textwidth]{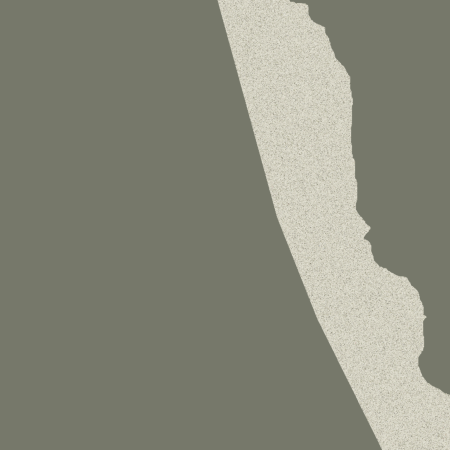}
         \label{fig:zanzibar_pixelswap_bare_zeron_b4_1}
     \end{subfigure}
     \\ \vspace{-0.3cm} \hspace{0.1mm}
     \begin{subfigure}{0.24\textwidth}
         \centering
         \includegraphics[width=\textwidth]{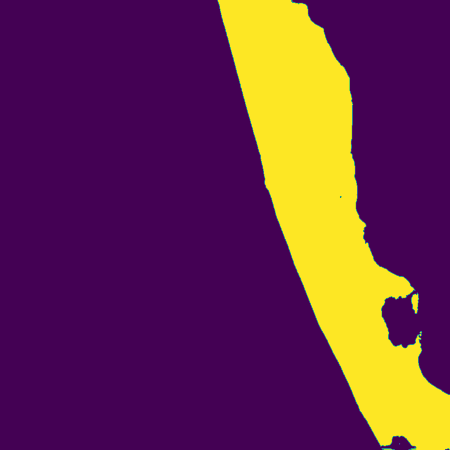}
         \label{fig:zanzibar_pixelswap_bare_zeron_b4_0_pred}
     \end{subfigure}
     \begin{subfigure}{0.24\textwidth}
         \centering
         \includegraphics[width=\textwidth]{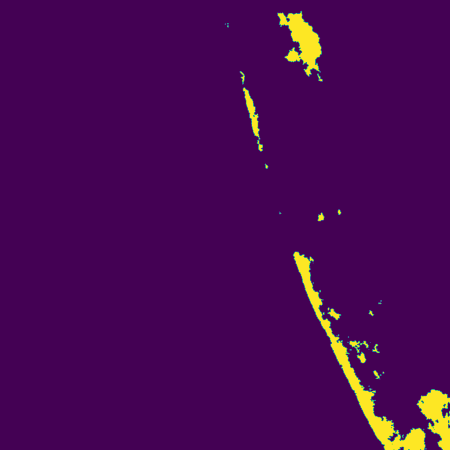}
         \label{fig:zanzibar_pixelswap_bare_zeron_b4_0.33_pred}
     \end{subfigure}
     \begin{subfigure}{0.24\textwidth}
         \centering
         \includegraphics[width=\textwidth]{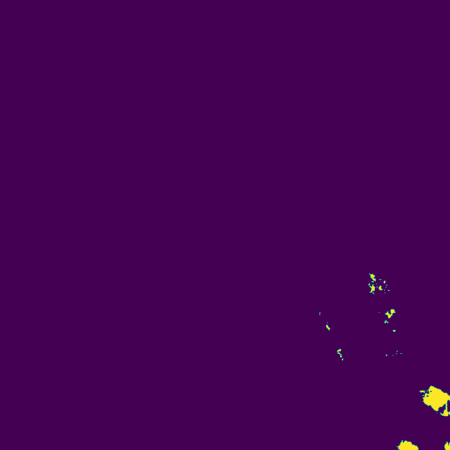}
         \label{fig:zanzibar_pixelswap_bare_zeron_b4_0.66_pred}
     \end{subfigure}
     \begin{subfigure}{0.24\textwidth}
         \centering
         \includegraphics[width=\textwidth]{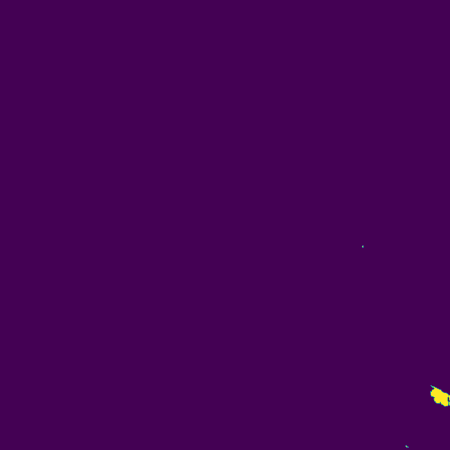}
         \label{fig:zanzibar_pixelswap_bare_zeron_b4_1_pred}
     \end{subfigure}
     \vspace{-10pt}
     \caption{Zanzibar region, \textbf{pixel-swap} transformation on the \textit{bare} class (expanded version of \Figure{fig:borderbias} (main paper)). Top two rows: Input images and model predictions. Bottom two rows: Input images, where pixels \emph{not} of the \textit{bare} class have been replaced by the per-channel mean of the training set, and model predictions below. Pixel-swap is applied to the input images with proportion $p$ swapped, where $p \in \{0, 0.33, 0.66, 1\}$ (from left to right). Note in the bottom row how the prediction accuracy deteriorates much faster as $p$ increases, compared to the case when the surrounding context is kept intact (second row). Also note how the border is no longer correctly classified for the input images when the surrounding context is removed. Predictions made using the U-Net-Efficientnet-B4 model.}
     \label{fig:borderbias_zeron}
\end{figure*}

\begin{figure*}[t]
     \centering
     \begin{subfigure}{0.24\textwidth}
         \centering
         \includegraphics[width=\textwidth]{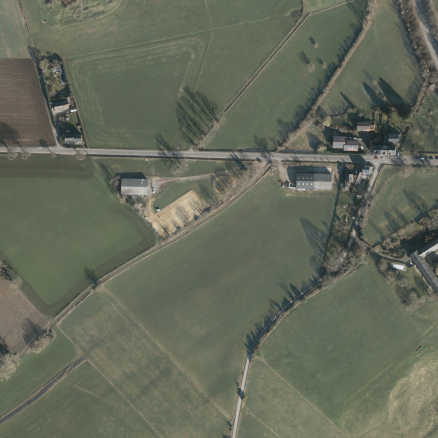}
     \end{subfigure}
     \begin{subfigure}{0.24\textwidth}
         \centering
         \includegraphics[width=\textwidth]{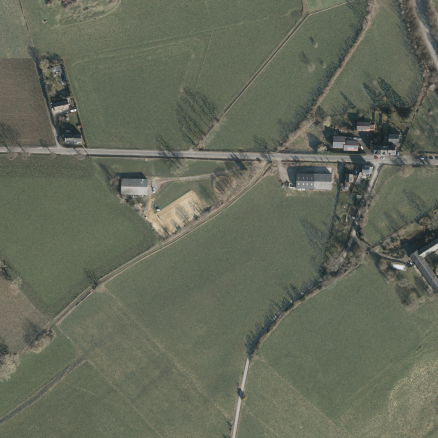}
     \end{subfigure}
     \begin{subfigure}{0.24\textwidth}
         \centering
         \includegraphics[width=\textwidth]{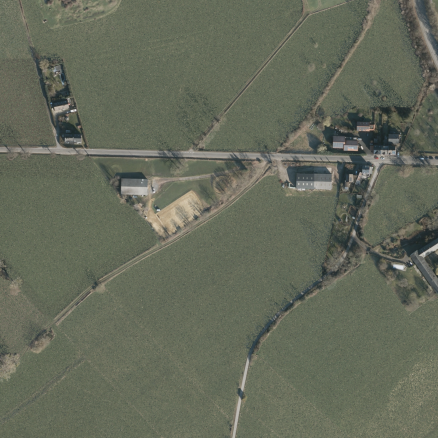}
     \end{subfigure}
     \begin{subfigure}{0.24\textwidth}
         \centering
         \includegraphics[width=\textwidth]{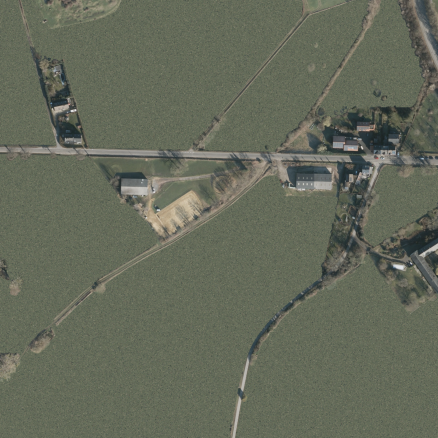}
     \end{subfigure}
     \\ \vspace{-0cm} 
     \begin{subfigure}{0.24\textwidth}
         \centering
         \includegraphics[width=\textwidth]{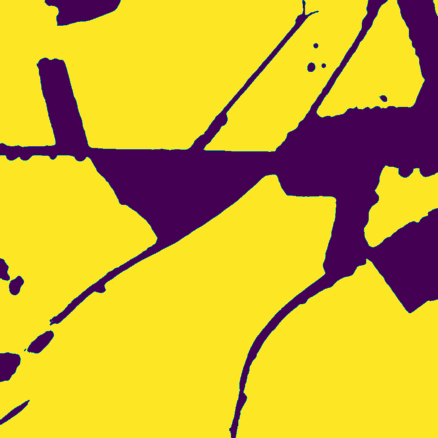}
     \end{subfigure}
     \begin{subfigure}{0.24\textwidth}
         \centering
         \includegraphics[width=\textwidth]{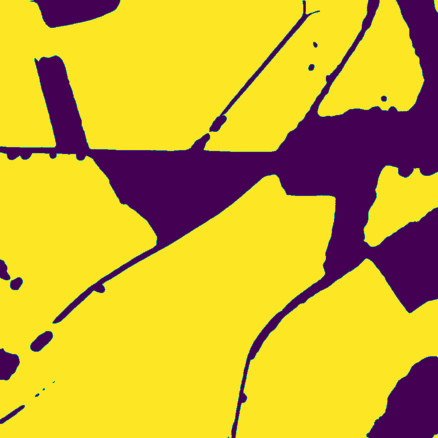}
     \end{subfigure}
     \begin{subfigure}{0.24\textwidth}
         \centering
         \includegraphics[width=\textwidth]{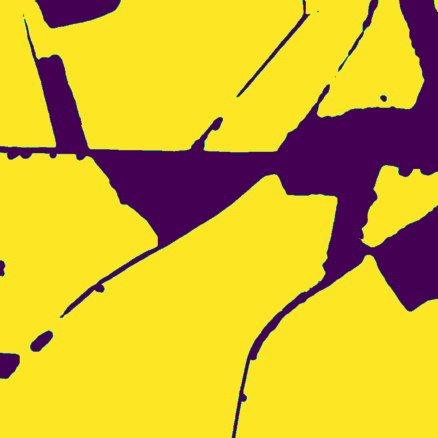}
     \end{subfigure}
     \begin{subfigure}{0.24\textwidth}
         \centering
         \includegraphics[width=\textwidth]{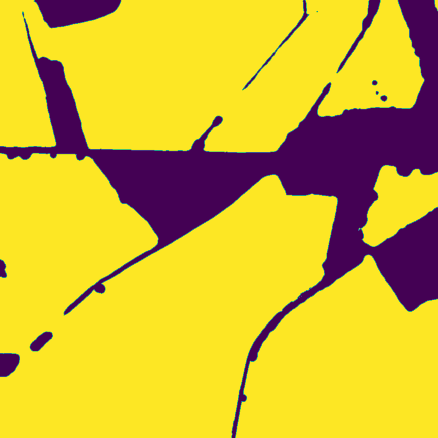}
     \end{subfigure}
     \\ \vspace{-0cm} 
     \begin{subfigure}{0.24\textwidth}
         \centering
         \includegraphics[width=\textwidth]{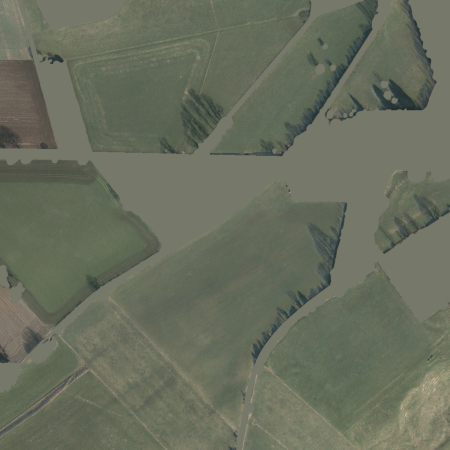}
     \end{subfigure}
     \begin{subfigure}{0.24\textwidth}
         \centering
         \includegraphics[width=\textwidth]{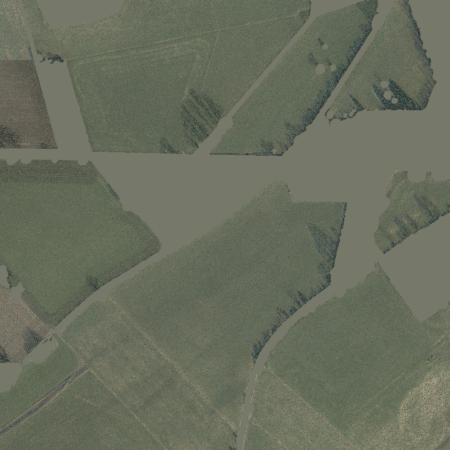}
     \end{subfigure}
     \begin{subfigure}{0.24\textwidth}
         \centering
         \includegraphics[width=\textwidth]{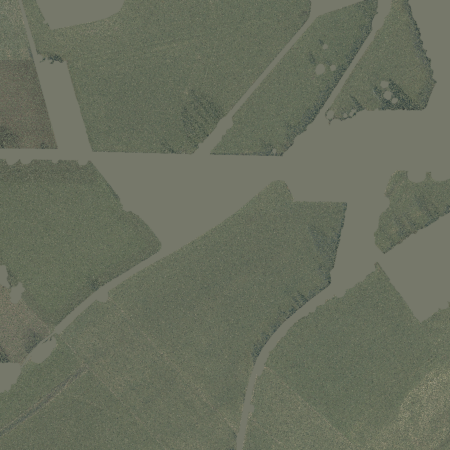}
     \end{subfigure}
     \begin{subfigure}{0.24\textwidth}
         \centering
         \includegraphics[width=\textwidth]{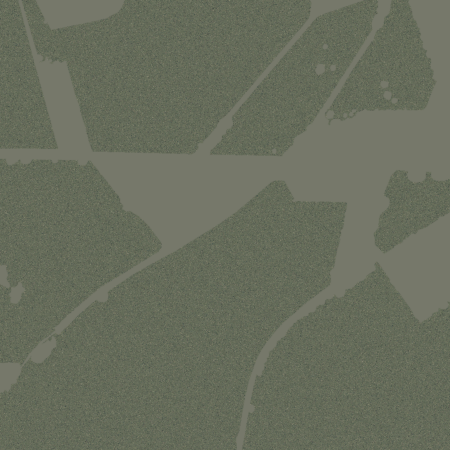}
     \end{subfigure}
     \\ \vspace{-0cm} 
     \begin{subfigure}{0.24\textwidth}
         \centering
         \includegraphics[width=\textwidth]{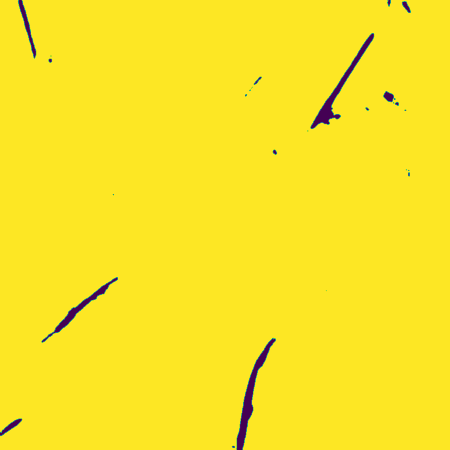}
     \end{subfigure}
     \begin{subfigure}{0.24\textwidth}
         \centering
         \includegraphics[width=\textwidth]{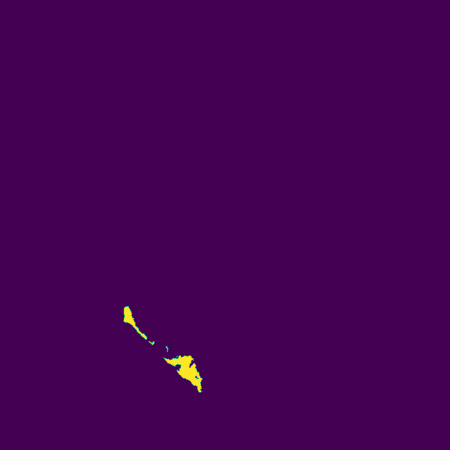}
     \end{subfigure}
     \begin{subfigure}{0.24\textwidth}
         \centering
         \includegraphics[width=\textwidth]{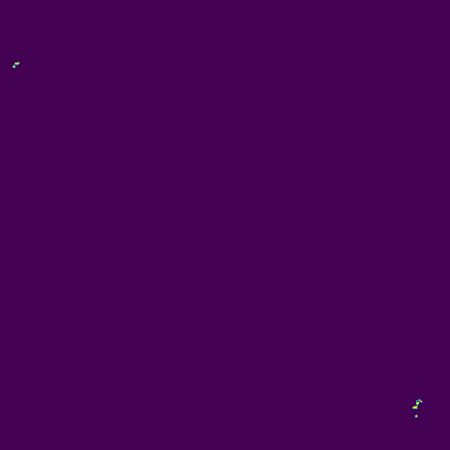}
     \end{subfigure}
     \begin{subfigure}{0.24\textwidth}
         \centering
         \includegraphics[width=\textwidth]{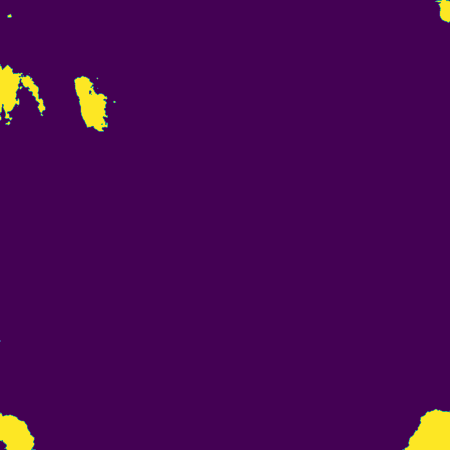}
     \end{subfigure}
     \caption{Aachen region, \textbf{pixel-swap} transformation on the \textit{agriculture} class.  Top two rows: Input images and model predictions. Bottom two rows: Input images, where pixels \emph{not} of the \textit{agriculture} class have been replaced by the per-channel mean of the training set, and model predictions below. Pixel-swap is applied to the input images with proportion $p$ swapped, where $p \in \{0, 0.33, 0.66, 1\}$ (from left to right). Removing the surrounding context yields significantly worse predictions for each proportion $p$ (bottom row), whereas predictions remain accurate independently of $p$ when context is kept (second row). Predictions made using the U-Net-Efficientnet-B4 model.} 
     \label{appfig:pixelswap_zeron_bigdemo}
\end{figure*}

\textbf{Additional color distortion experiments.} Here we include experimental results for additional forms of color distortion, which we commonly refer to as \emph{color-duplication}, as described next.

\textit{Color-duplication:} As in the main paper, let $\mathcal{I}_{c} = \{(i_k, j_k)\}_{k=1}^{K}$ denote the set of pixel positions corresponding to class $c$ in a given RGB aerial or satellite image $\bs{I}$. The color-duplication transformation is somewhat similar to the gray-scale transformation defined in the main paper. However, instead of a convex combination between the class $c$-related pixels of the original image $\bs{I}$ and a gray-scale counterpart $\bs{G}$, we first copy one color channel $\bs{I}_h$ (with $h \in \{R, G, B\}$ and where $R$, $G$ and $B$ respectively denote the red, blue and green color channels), and then construct the color-duplicated image $\bs{I}^D = [\bs{I}_h; \bs{I}_h; \bs{I}_h]$. Given $\bs{I}^D$, the class $c$-related color duplication at mixing proportion $\lambda$ is given by $\bs{I}(i,j) = (1-\lambda) \bs{I}(i,j) + \lambda \bs{I}^D(i,j) = (1-\lambda) [\bs{I}_R; \bs{I}_G; \bs{I}_B](i,j) + \lambda [\bs{I}_h; \bs{I}_h; \bs{I}_h](i,j)$, where $(i,j) \in \mathcal{I}_c$. For example, in the red-duplication case with $h=R$, the transformation is given by
$(1-\lambda) [\bs{I}_R; \bs{I}_G; \bs{I}_B](i,j) + \lambda [\bs{I}_R; \bs{I}_R; \bs{I}_R](i,j) = [\bs{I}_R; (1-\lambda)\bs{I}_G+\lambda \bs{I}_R; (1-\lambda)\bs{I}_B+\lambda \bs{I}_R](i,j)$. As usual, $\bs{I}$ is left unchanged for pixels of other classes. Finally, note that we still normalize the model input with the per-channel means of the training set, even if the channels are changed.

The results for the color-duplication transformations (red-, green- and blue-duplications) are shown in \Figure{fig:colorpicker}. As can be seen, despite the extreme nature of this type of color distortion, most classes are only marginally affected by it. Hence, taken into account also the gray-scale experiments from the main paper, these results suggest that deep learning models are not very sensitive to color transformations of EO imagery.

\begin{figure*}[t]
     \centering
     \begin{subfigure}{0.24\textwidth}
         \centering
         \includegraphics[width=\textwidth]{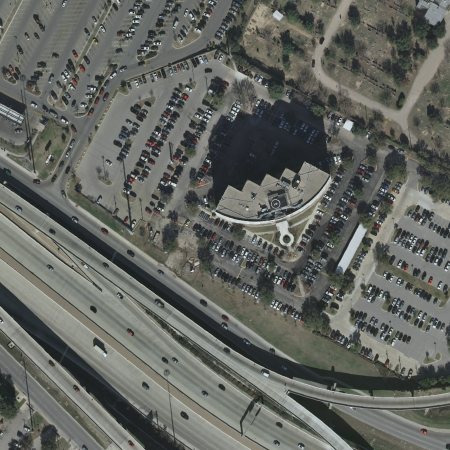}
     \end{subfigure}
     \begin{subfigure}{0.24\textwidth}
         \centering
         \includegraphics[width=\textwidth]{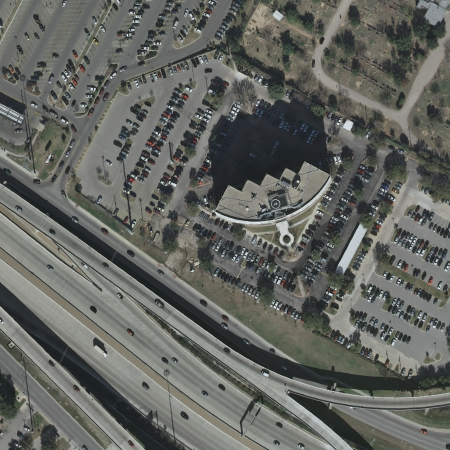}
     \end{subfigure}
     \begin{subfigure}{0.24\textwidth}
         \centering
         \includegraphics[width=\textwidth]{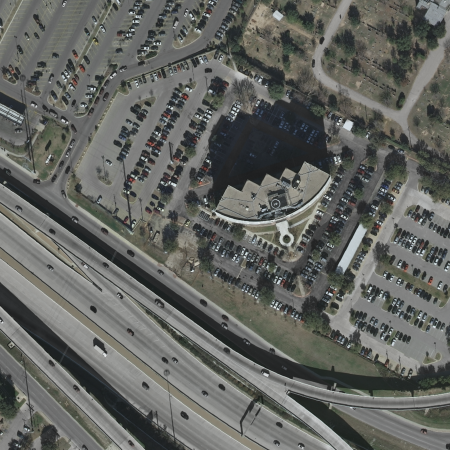}
     \end{subfigure}
     \begin{subfigure}{0.24\textwidth}
         \centering
         \includegraphics[width=\textwidth]{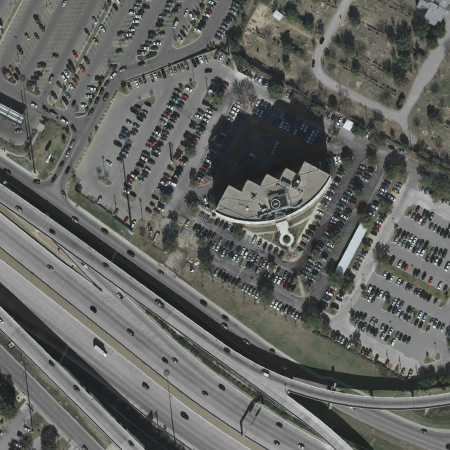}
     \end{subfigure}
     \\ \vspace{-0cm} 
     \begin{subfigure}{0.24\textwidth}
         \centering
         \includegraphics[width=\textwidth]{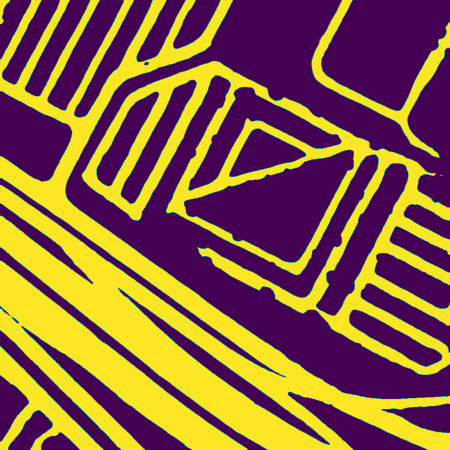}
     \end{subfigure}
     \begin{subfigure}{0.24\textwidth}
         \centering
         \includegraphics[width=\textwidth]{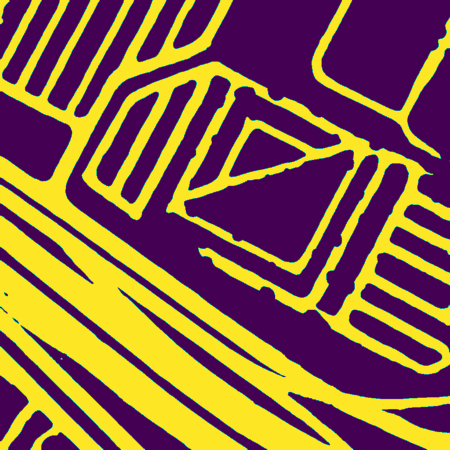}
     \end{subfigure}
     \begin{subfigure}{0.24\textwidth}
         \centering
         \includegraphics[width=\textwidth]{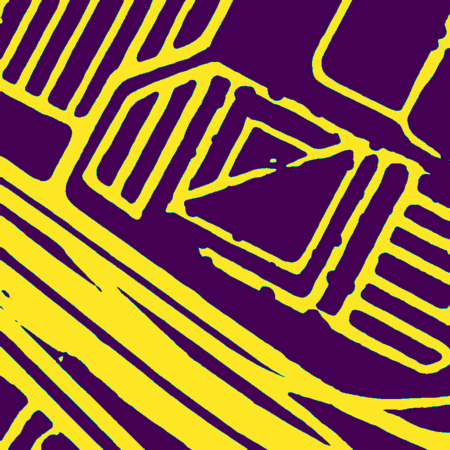}
     \end{subfigure}
     \begin{subfigure}{0.24\textwidth}
         \centering
         \includegraphics[width=\textwidth]{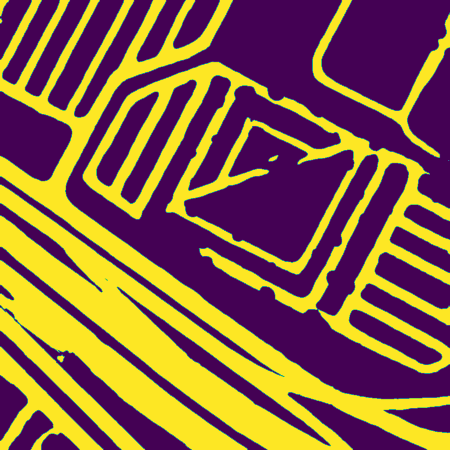}
     \end{subfigure}
     \\ \vspace{-0cm} 
     \begin{subfigure}{0.24\textwidth}
         \centering
         \includegraphics[width=\textwidth]{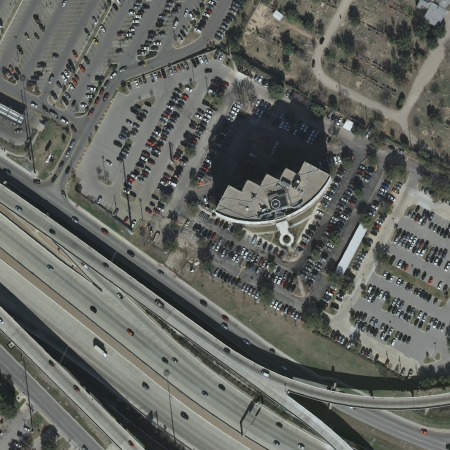}
     \end{subfigure}
     \begin{subfigure}{0.24\textwidth}
         \centering
         \includegraphics[width=\textwidth]{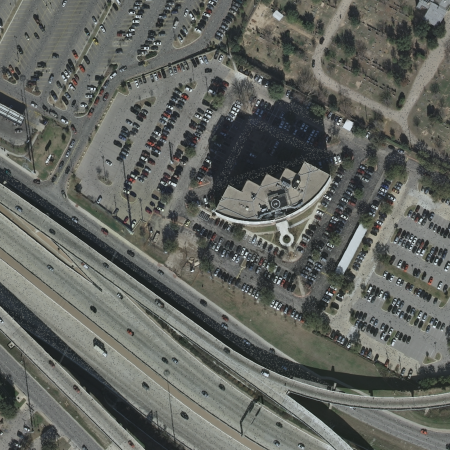}
     \end{subfigure}
     \begin{subfigure}{0.24\textwidth}
         \centering
         \includegraphics[width=\textwidth]{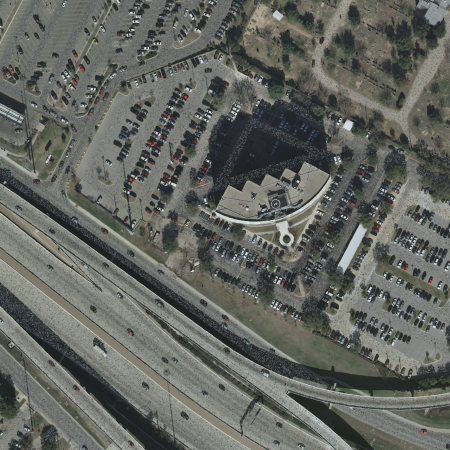}
     \end{subfigure}
     \begin{subfigure}{0.24\textwidth}
         \centering
         \includegraphics[width=\textwidth]{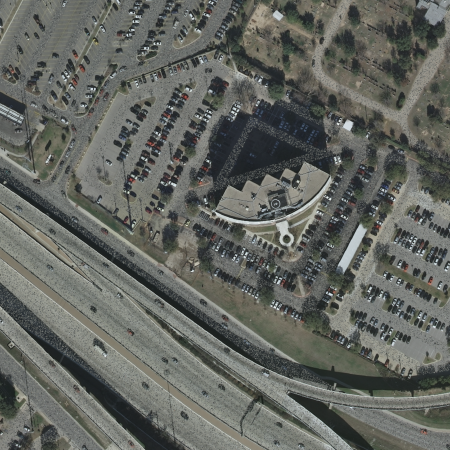}
     \end{subfigure}
     \\ \vspace{-0cm} 
     \begin{subfigure}{0.24\textwidth}
         \centering
         \includegraphics[width=\textwidth]{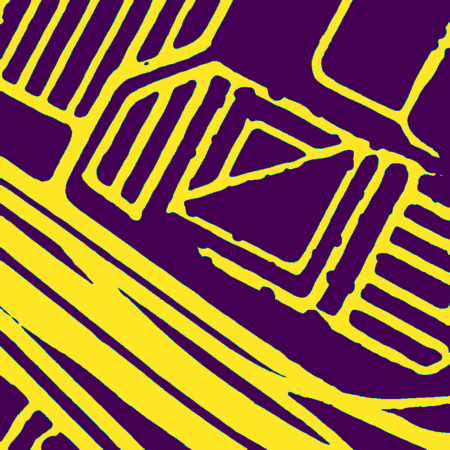}
     \end{subfigure}
     \begin{subfigure}{0.24\textwidth}
         \centering
         \includegraphics[width=\textwidth]{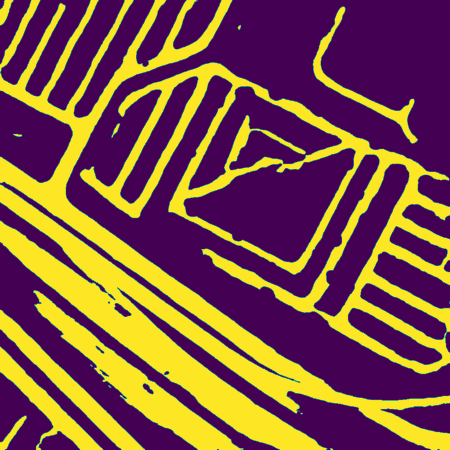}
     \end{subfigure}
     \begin{subfigure}{0.24\textwidth}
         \centering
         \includegraphics[width=\textwidth]{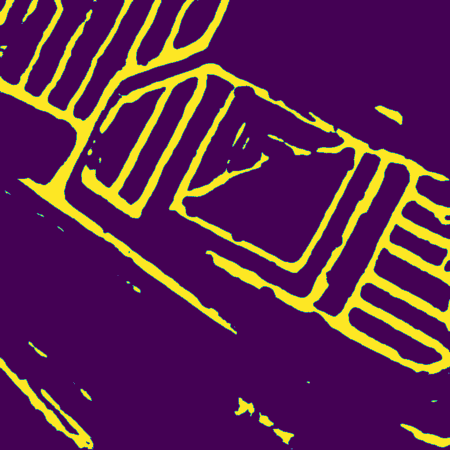}
     \end{subfigure}
     \begin{subfigure}{0.24\textwidth}
         \centering
         \includegraphics[width=\textwidth]{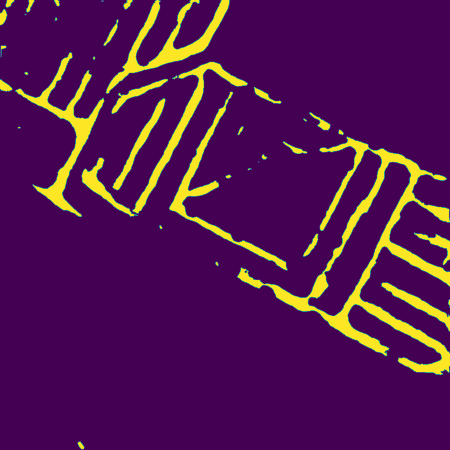}
     \end{subfigure}
     \caption{Austin region, transformations on the \textit{road} class. The top two rows show the \textbf{gray-scale} transformed images with gray-scale proportion $\lambda \in \{0, 0.33, 0.66, 1\}$ (from left to right) and corresponding model predictions below. The bottom two rows show \textbf{pixel-swap} transformed images with proportion $p$ swapped, where $p \in \{0, 0.1, 0.2, 0.3\}$ (from left to right) and corresponding model predictions below. We see that the predictions are very robust with respect to color distortion (top), and very sensitive to texture distortion (bottom).} 
     \label{appfig:austin_road}
\end{figure*}

\begin{figure*}[t]
\centering
\begin{subfigure}{0.24\textwidth}
\centering
\includegraphics[width=\textwidth]{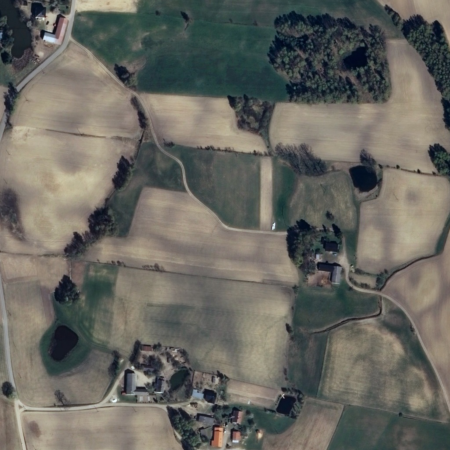}
\end{subfigure}
\begin{subfigure}{0.24\textwidth}
\centering
\includegraphics[width=\textwidth]{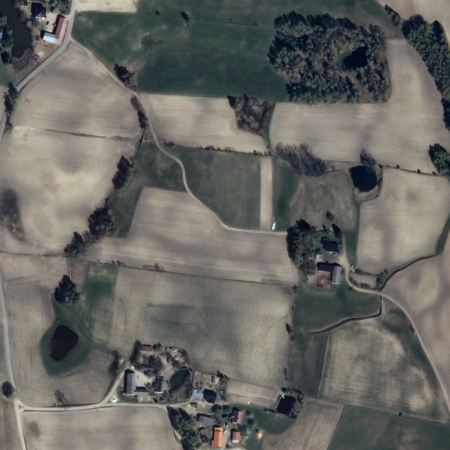}
\end{subfigure}
\begin{subfigure}{0.24\textwidth}
\centering
\includegraphics[width=\textwidth]{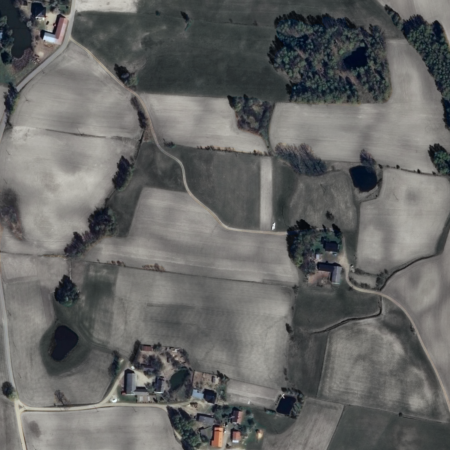}
\end{subfigure}
\begin{subfigure}{0.24\textwidth}
\centering
\includegraphics[width=\textwidth]{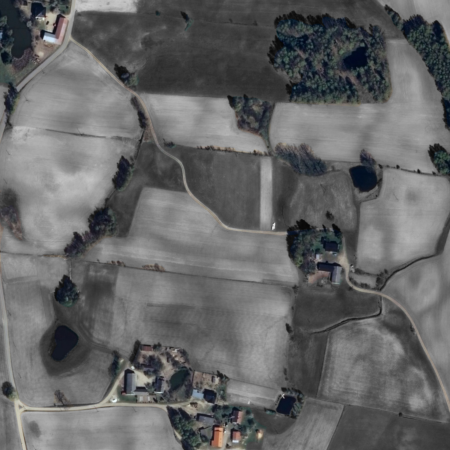}
\end{subfigure}
\\ \vspace{-0cm} 
\begin{subfigure}{0.24\textwidth}
\centering
\includegraphics[width=\textwidth]{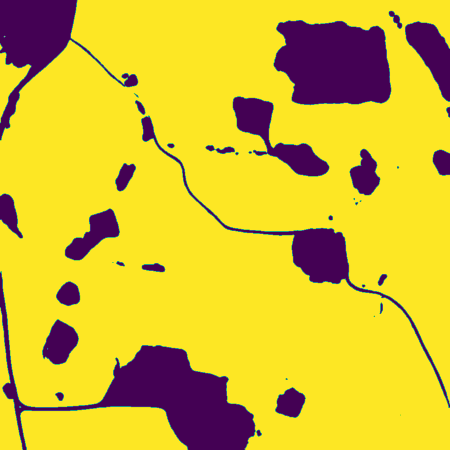}
\end{subfigure}
\begin{subfigure}{0.24\textwidth}
\centering
\includegraphics[width=\textwidth]{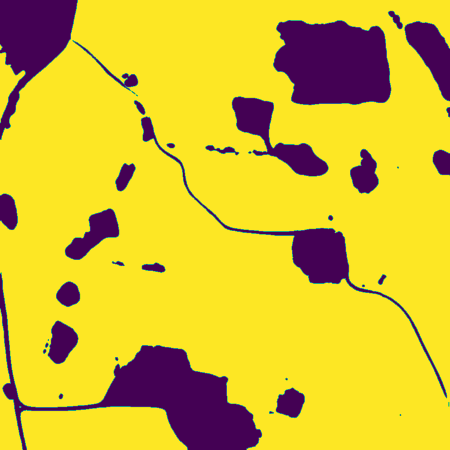}
\end{subfigure}
\begin{subfigure}{0.24\textwidth}
\centering
\includegraphics[width=\textwidth]{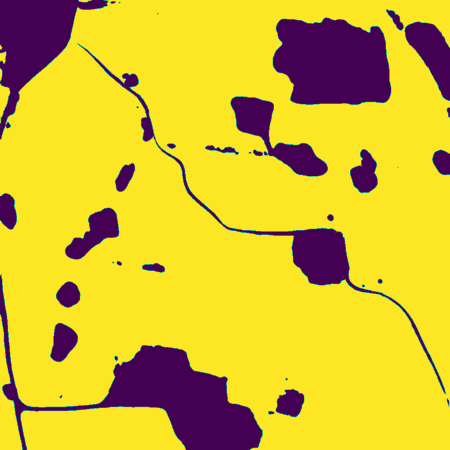}
\end{subfigure}
\begin{subfigure}{0.24\textwidth}
\centering
\includegraphics[width=\textwidth]{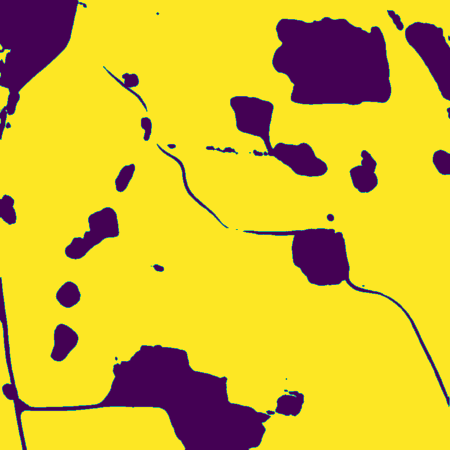}
\end{subfigure}
\ \vspace{-0cm} 
\begin{subfigure}{0.24\textwidth}
\centering
\includegraphics[width=\textwidth]{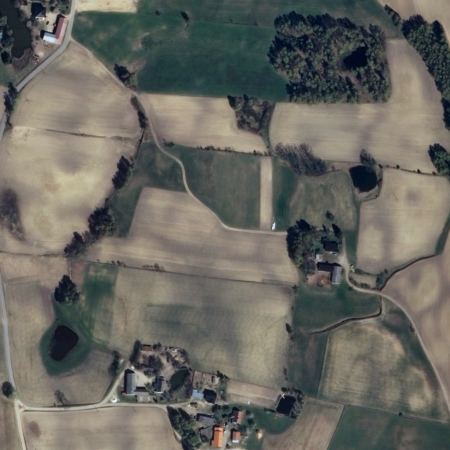}
\end{subfigure}
\begin{subfigure}{0.24\textwidth}
\centering
\includegraphics[width=\textwidth]{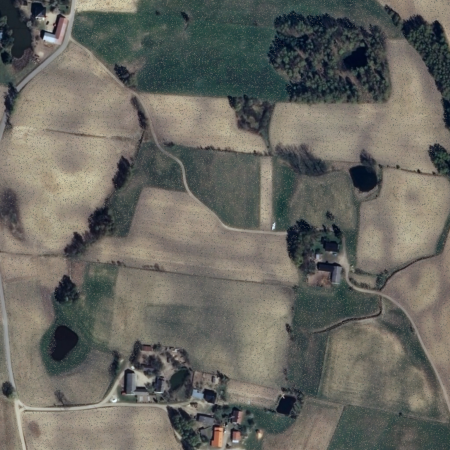}
\end{subfigure}
\begin{subfigure}{0.24\textwidth}
\centering
\includegraphics[width=\textwidth]{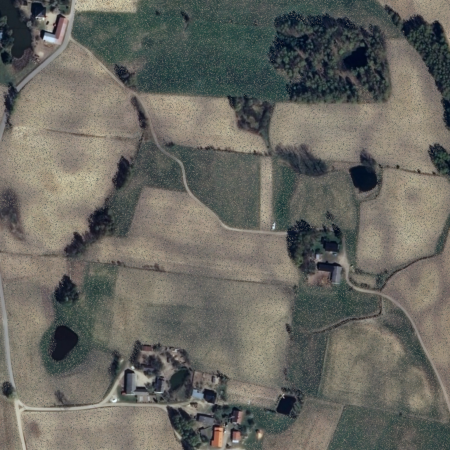}
\end{subfigure}
\begin{subfigure}{0.24\textwidth}
\centering
\includegraphics[width=\textwidth]{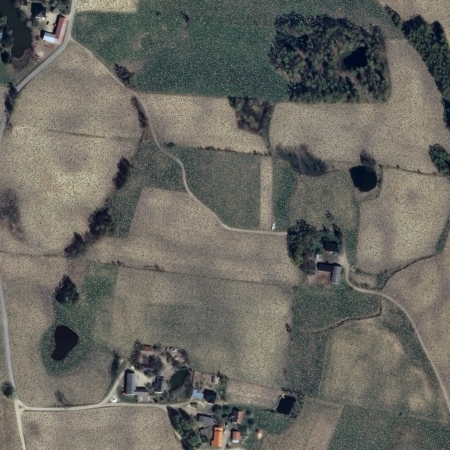}
\end{subfigure}
\ \vspace{-0cm} 
\begin{subfigure}{0.24\textwidth}
\centering
\includegraphics[width=\textwidth]{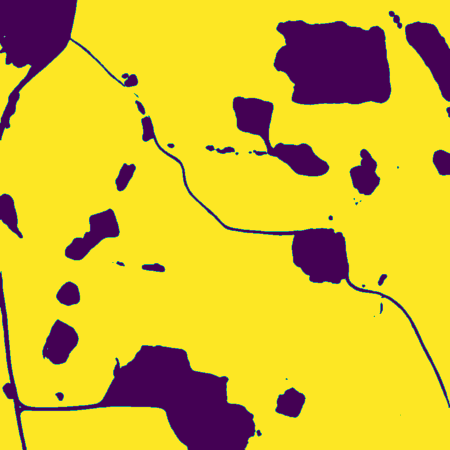}
\end{subfigure}
\begin{subfigure}{0.24\textwidth}
\centering
\includegraphics[width=\textwidth]{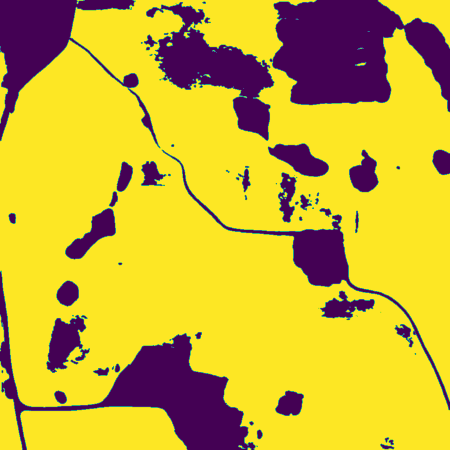}
\end{subfigure}
\begin{subfigure}{0.24\textwidth}
\centering
\includegraphics[width=\textwidth]{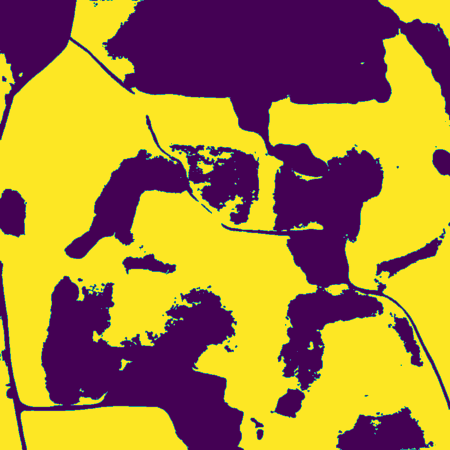}
\end{subfigure}
\begin{subfigure}{0.24\textwidth}
\centering
\includegraphics[width=\textwidth]{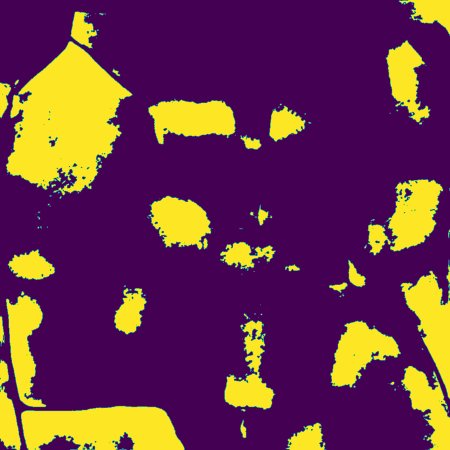}
\end{subfigure}
\caption{Pomorskie region, transformations on the \textit{agriculture} class. The top two rows show \textbf{gray-scale} transformed images with gray-scale proportion $\lambda \in \{0, 0.33, 0.66, 1\}$ (from left to right) and corresponding model predictions. The bottom two rows show \textbf{pixel-swap} transformed images with proportion $p$ swapped, $p \in \{0, 0.1, 0.2, 0.3\}$  (from left to right) and corresponding model predictions below. We see that the predictions are very robust with respect to color distortion (top),
and very sensitive to texture distortion (bottom)}
\label{appfig:pomorskie_agriculture}
\end{figure*}

\begin{figure*}[t]
     \centering
     \begin{subfigure}{0.24\textwidth}
         \centering
         \includegraphics[width=\textwidth]{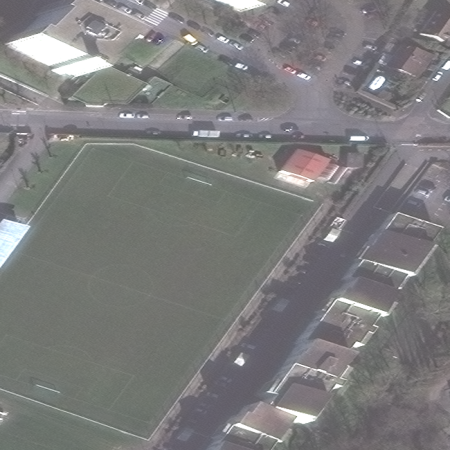}
     \end{subfigure}
     \begin{subfigure}{0.24\textwidth}
         \centering
         \includegraphics[width=\textwidth]{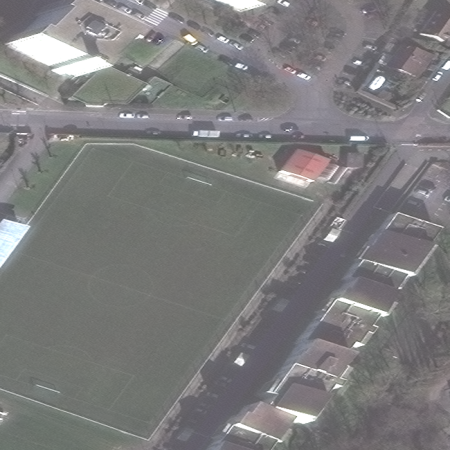}
     \end{subfigure}
     \begin{subfigure}{0.24\textwidth}
         \centering
         \includegraphics[width=\textwidth]{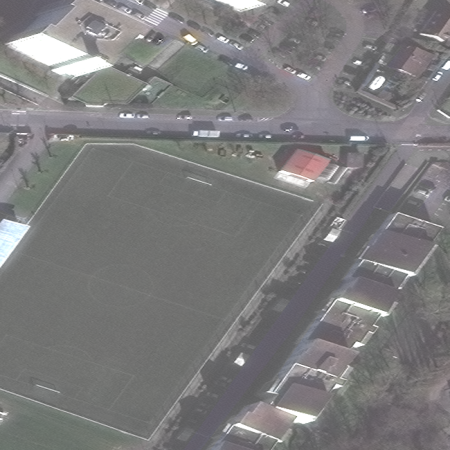}
     \end{subfigure}
     \begin{subfigure}{0.24\textwidth}
         \centering
         \includegraphics[width=\textwidth]{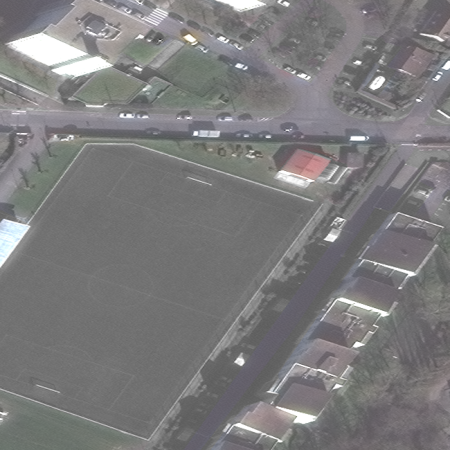}
     \end{subfigure}
     \\ \vspace{-0cm} 
     \begin{subfigure}{0.24\textwidth}
         \centering
         \includegraphics[width=\textwidth]{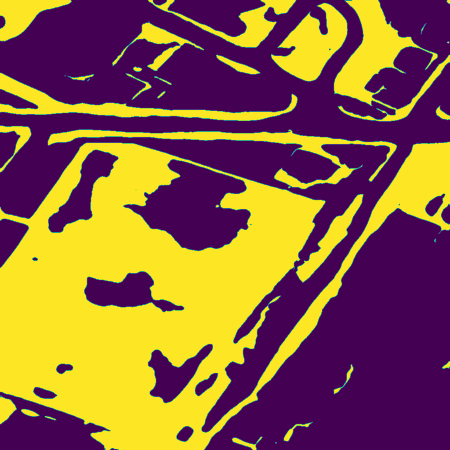}
     \end{subfigure}
     \begin{subfigure}{0.24\textwidth}
         \centering
         \includegraphics[width=\textwidth]{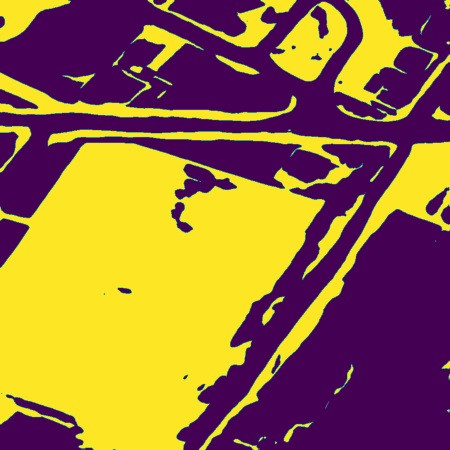}
     \end{subfigure}
     \begin{subfigure}{0.24\textwidth}
         \centering
         \includegraphics[width=\textwidth]{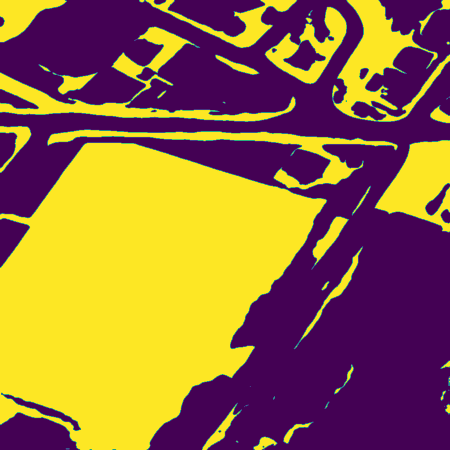}
     \end{subfigure}
     \begin{subfigure}{0.24\textwidth}
         \centering
         \includegraphics[width=\textwidth]{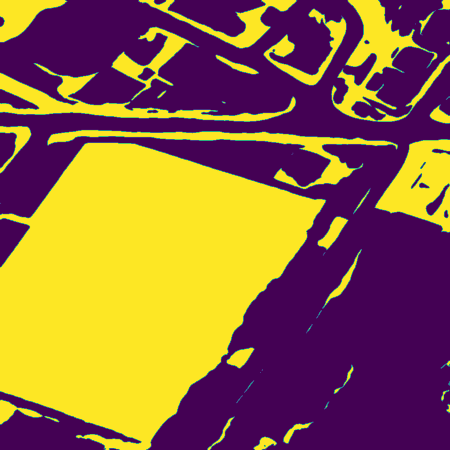}
     \end{subfigure}
     \\ \vspace{-0cm} 
     \begin{subfigure}{0.24\textwidth}
         \centering
         \includegraphics[width=\textwidth]{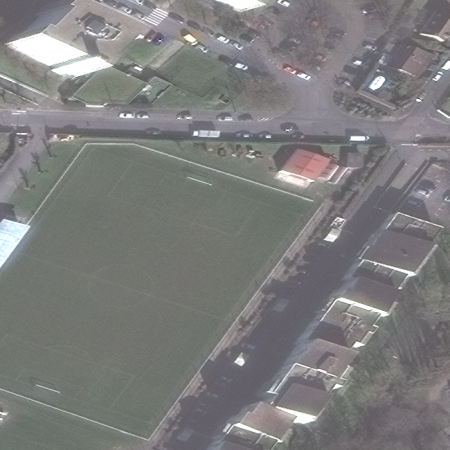}
     \end{subfigure}
     \begin{subfigure}{0.24\textwidth}
         \centering
         \includegraphics[width=\textwidth]{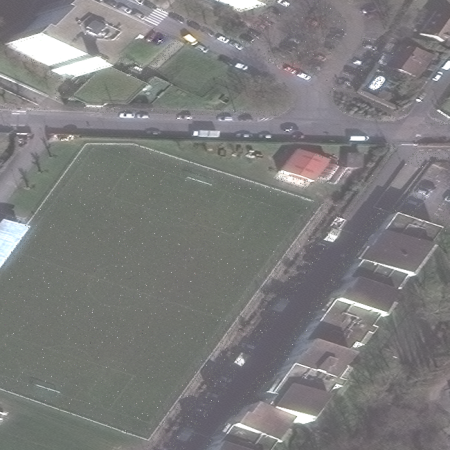}
     \end{subfigure}
     \begin{subfigure}{0.24\textwidth}
         \centering
         \includegraphics[width=\textwidth]{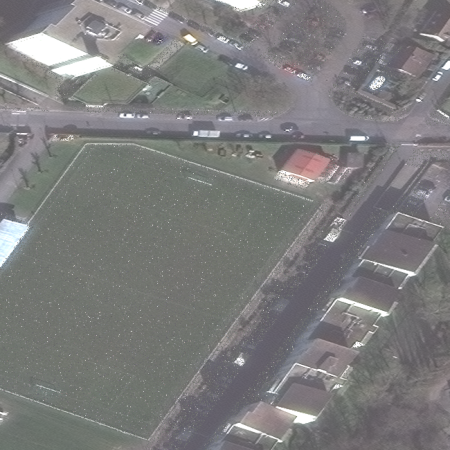}
     \end{subfigure}
     \begin{subfigure}{0.24\textwidth}
         \centering
         \includegraphics[width=\textwidth]{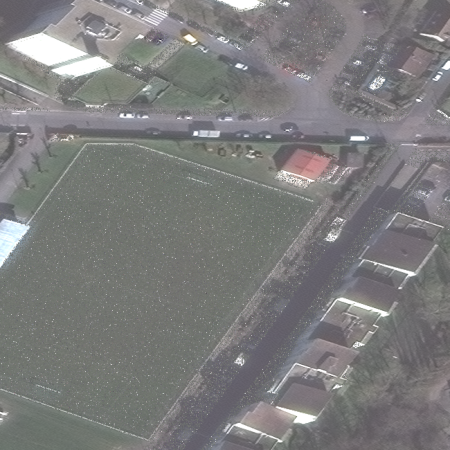}
     \end{subfigure}
     \\ \vspace{-0cm} 
     \begin{subfigure}{0.24\textwidth}
         \centering
         \includegraphics[width=\textwidth]{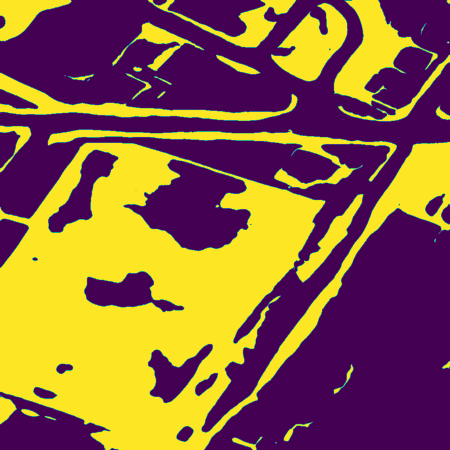}
     \end{subfigure}
     \begin{subfigure}{0.24\textwidth}
         \centering
         \includegraphics[width=\textwidth]{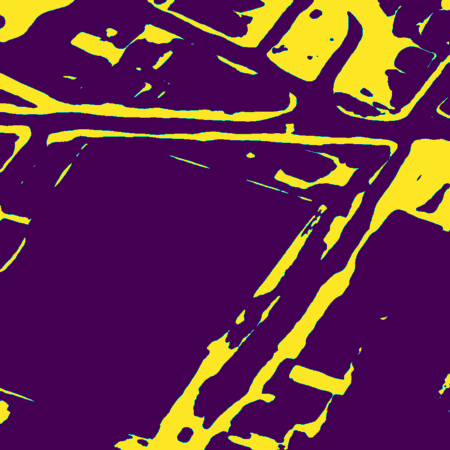}
     \end{subfigure}
     \begin{subfigure}{0.24\textwidth}
         \centering
         \includegraphics[width=\textwidth]{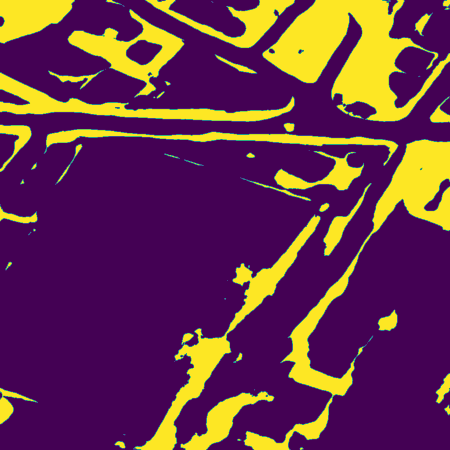}
     \end{subfigure}
     \begin{subfigure}{0.24\textwidth}
         \centering
         \includegraphics[width=\textwidth]{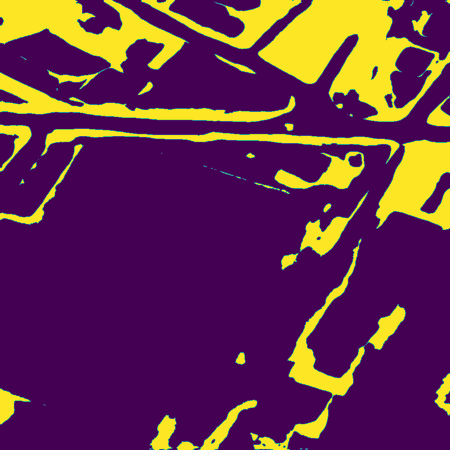}
     \end{subfigure}
     \caption{Paris region, transformations on the \textit{developed} class. The top two rows show \textbf{gray-scale} transformed images with gray-scale proportion $\lambda \in \{0, 0.33, 0.66, 1\}$ (from left to right) and corresponding model predictions. The bottom two rows show \textbf{pixel-swap} transformed images with proportion $p$ swapped, $p \in \{0, 0.1, 0.2, 0.3\}$  (from left to right) and corresponding model predictions below. We see that the predictions are very robust with respect to color distortion (top),
and very sensitive to texture distortion (bottom).} 
     \label{appfig:paris_developed}
\end{figure*}

\begin{figure*}[t]
     \centering
     \begin{subfigure}{0.24\textwidth}
         \centering
         \includegraphics[width=\textwidth]{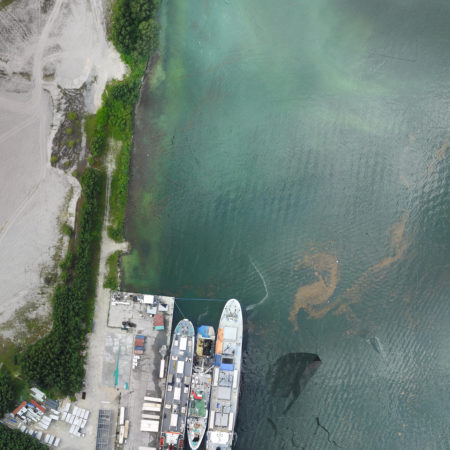}
     \end{subfigure}
     \begin{subfigure}{0.24\textwidth}
         \centering
         \includegraphics[width=\textwidth]{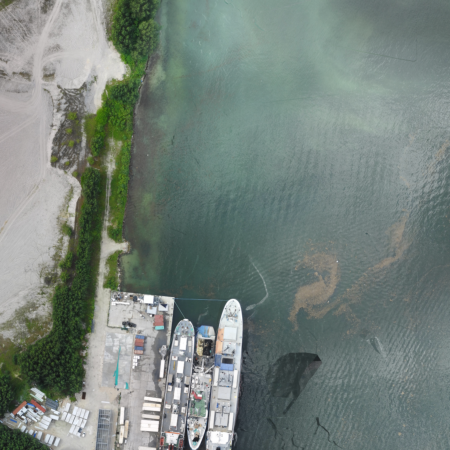}
     \end{subfigure}
     \begin{subfigure}{0.24\textwidth}
         \centering
         \includegraphics[width=\textwidth]{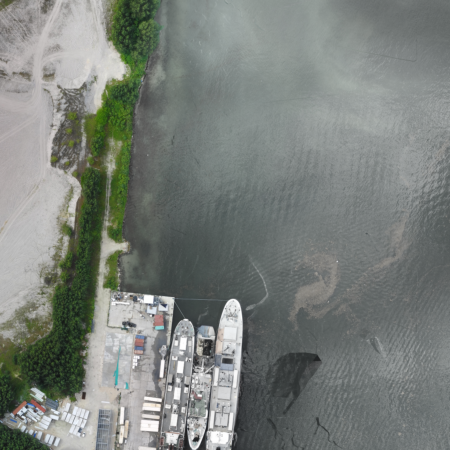}
     \end{subfigure}
     \begin{subfigure}{0.24\textwidth}
         \centering
         \includegraphics[width=\textwidth]{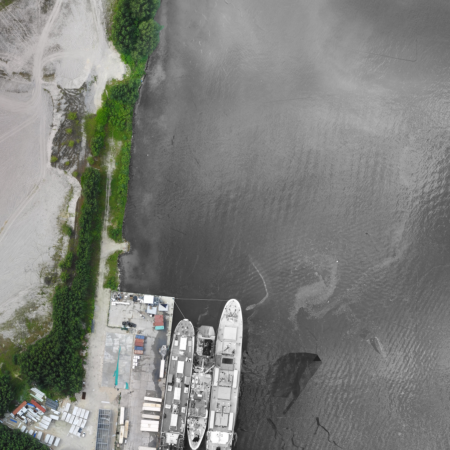}
     \end{subfigure}
     \\ \vspace{-0cm} 
     \begin{subfigure}{0.24\textwidth}
         \centering
         \includegraphics[width=\textwidth]{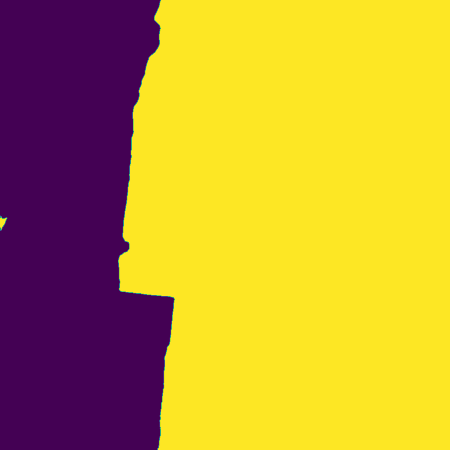}
     \end{subfigure}
     \begin{subfigure}{0.24\textwidth}
         \centering
         \includegraphics[width=\textwidth]{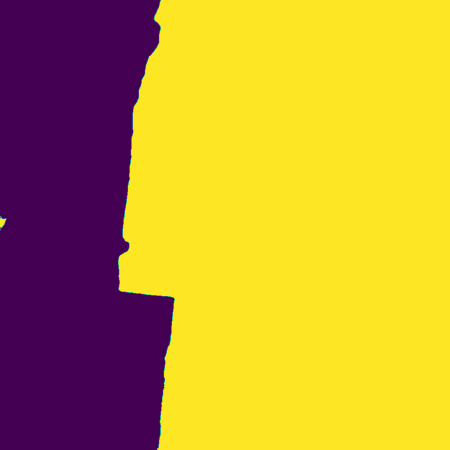}
     \end{subfigure}
     \begin{subfigure}{0.24\textwidth}
         \centering
         \includegraphics[width=\textwidth]{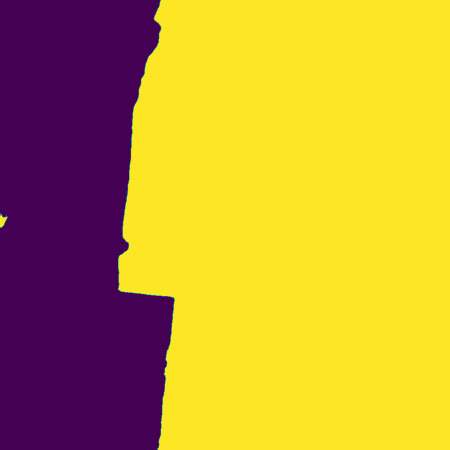}
     \end{subfigure}
     \begin{subfigure}{0.24\textwidth}
         \centering
         \includegraphics[width=\textwidth]{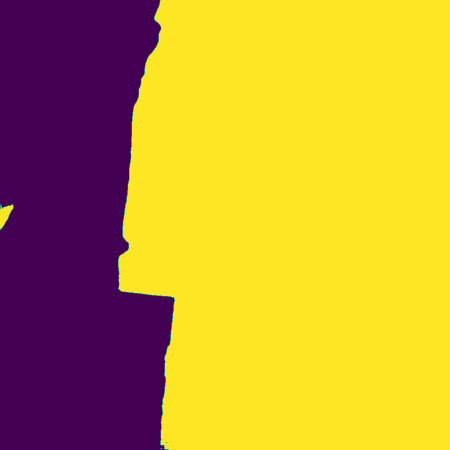}
     \end{subfigure}
     \\ \vspace{-0cm} 
     \begin{subfigure}{0.24\textwidth}
         \centering
         \includegraphics[width=\textwidth]{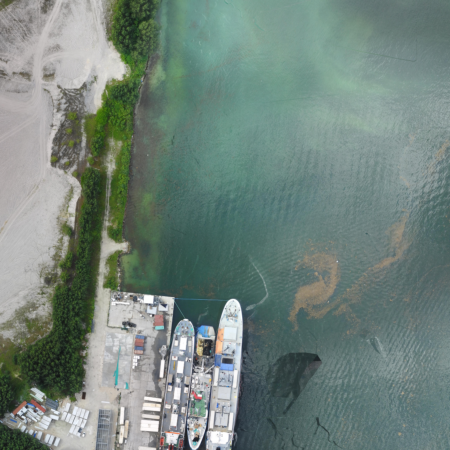}
     \end{subfigure}
     \begin{subfigure}{0.24\textwidth}
         \centering
         \includegraphics[width=\textwidth]{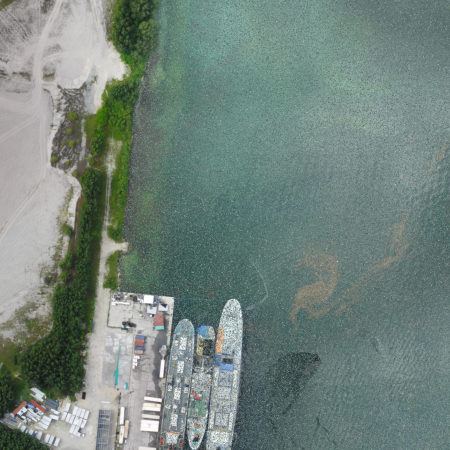}
     \end{subfigure}
     \begin{subfigure}{0.24\textwidth}
         \centering
         \includegraphics[width=\textwidth]{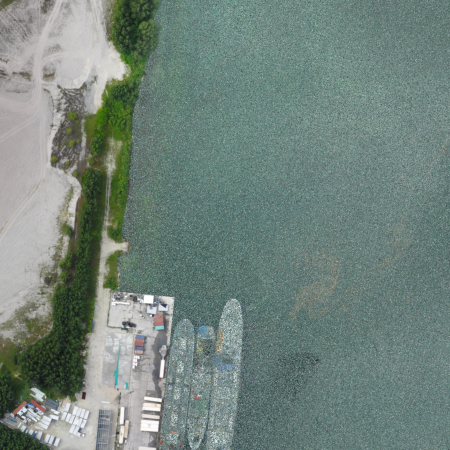}
     \end{subfigure}
     \begin{subfigure}{0.24\textwidth}
         \centering
         \includegraphics[width=\textwidth]{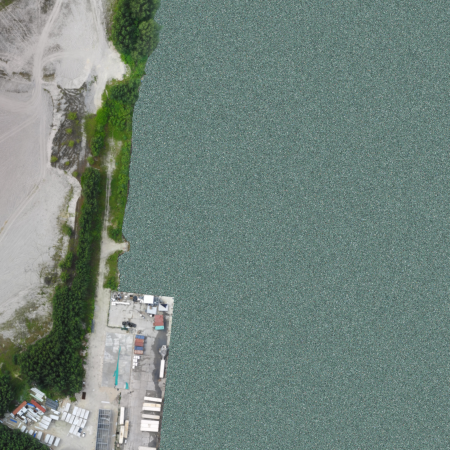}
     \end{subfigure}
     \\ \vspace{-0cm} 
     \begin{subfigure}{0.24\textwidth}
         \centering
         \includegraphics[width=\textwidth]{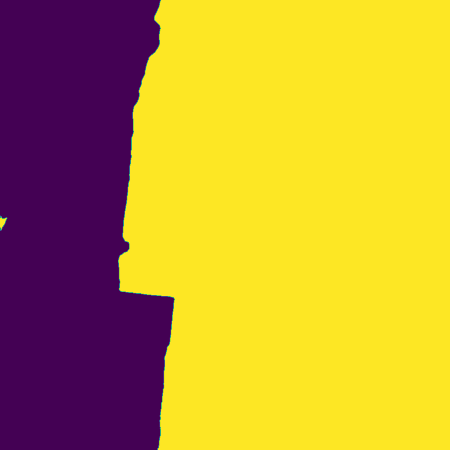}
     \end{subfigure}
     \begin{subfigure}{0.24\textwidth}
         \centering
         \includegraphics[width=\textwidth]{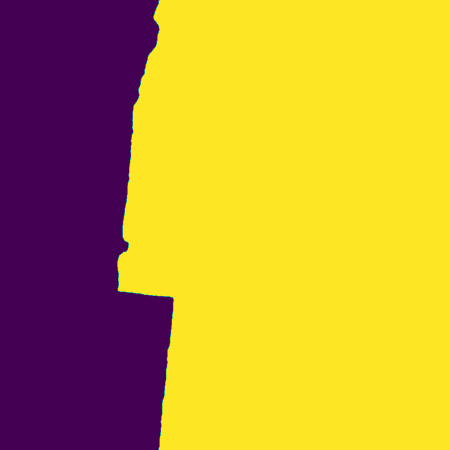}
     \end{subfigure}
     \begin{subfigure}{0.24\textwidth}
         \centering
         \includegraphics[width=\textwidth]{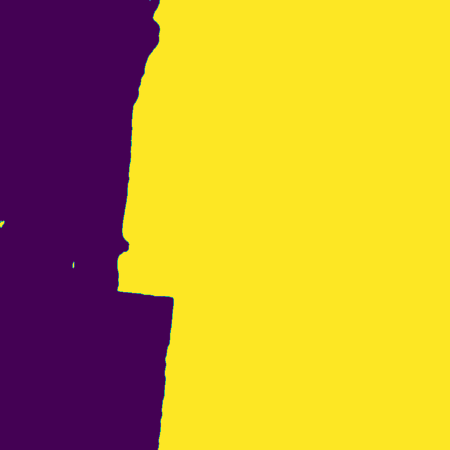}
     \end{subfigure}
     \begin{subfigure}{0.24\textwidth}
         \centering
         \includegraphics[width=\textwidth]{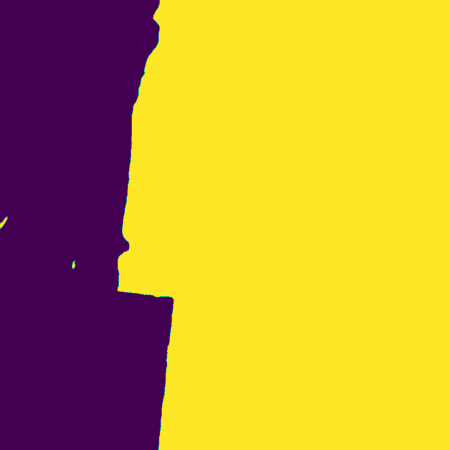}
     \end{subfigure}
     \caption{Mahe region, transformations on the \textit{water} class. The top two rows show \textbf{gray-scale} transformed images with gray-scale proportion $\lambda \in \{0, 0.33, 0.66, 1\}$ (from left to right) and corresponding model predictions. The bottom two rows show \textbf{pixel-swap} transformed images with proportion $p$ swapped, $p \in \{0, 0.33, 0.66, 1\}$  (from left to right) and corresponding model predictions below. We see that the predictions, in contrast to most other classes, are very robust with respect to color distortion (top) and texture distortion (bottom).} 
     \label{appfig:mahe_water}
\end{figure*}

\begin{figure*}[t]
     \centering
     \begin{subfigure}{0.24\textwidth}
         \centering
         \includegraphics[width=\textwidth]{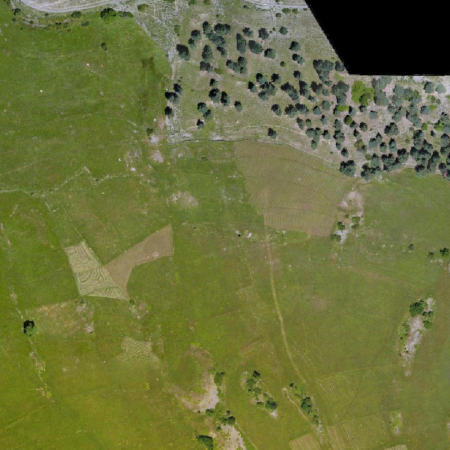}
     \end{subfigure}
     \begin{subfigure}{0.24\textwidth}
         \centering
         \includegraphics[width=\textwidth]{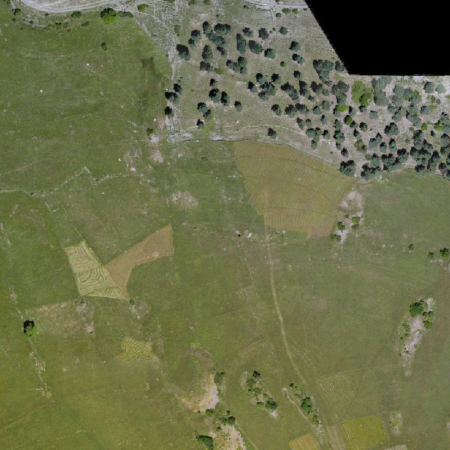}
     \end{subfigure}
     \begin{subfigure}{0.24\textwidth}
         \centering
         \includegraphics[width=\textwidth]{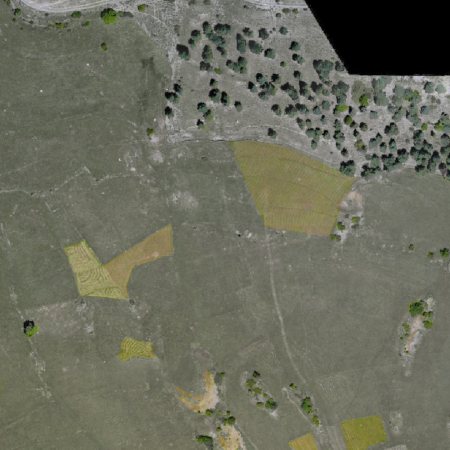}
     \end{subfigure}
     \begin{subfigure}{0.24\textwidth}
         \centering
         \includegraphics[width=\textwidth]{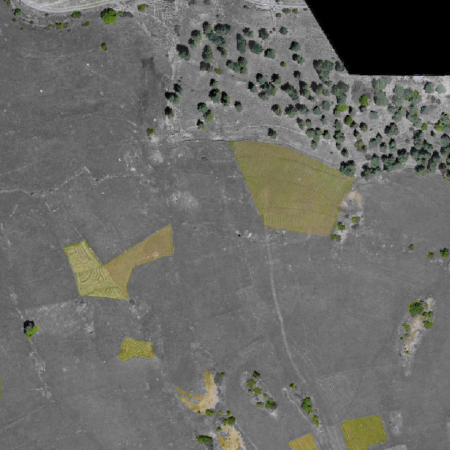}
     \end{subfigure}
     \\ \vspace{-0cm} 
     \begin{subfigure}{0.24\textwidth}
         \centering
         \includegraphics[width=\textwidth]{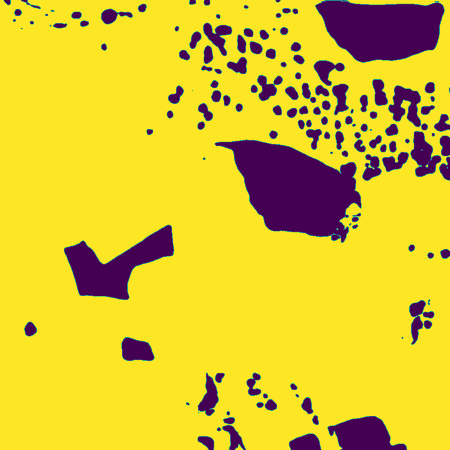}
     \end{subfigure}
     \begin{subfigure}{0.24\textwidth}
         \centering
         \includegraphics[width=\textwidth]{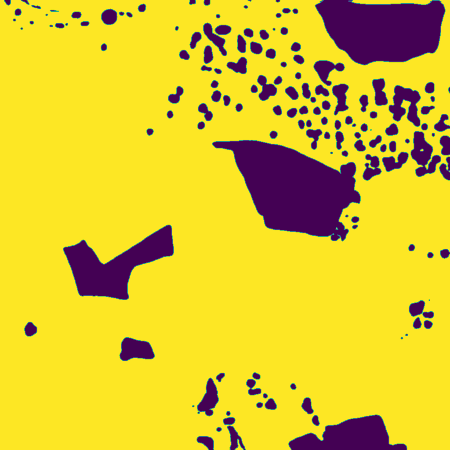}
     \end{subfigure}
     \begin{subfigure}{0.24\textwidth}
         \centering
         \includegraphics[width=\textwidth]{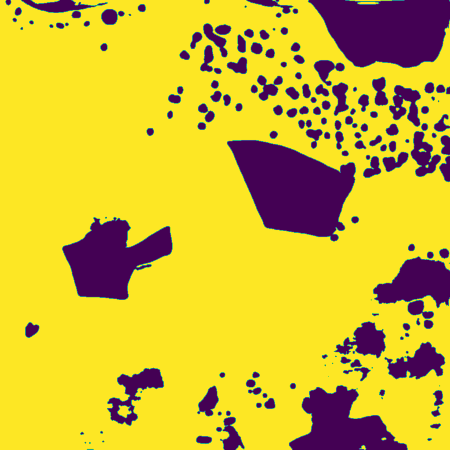}
     \end{subfigure}
     \begin{subfigure}{0.24\textwidth}
         \centering
         \includegraphics[width=\textwidth]{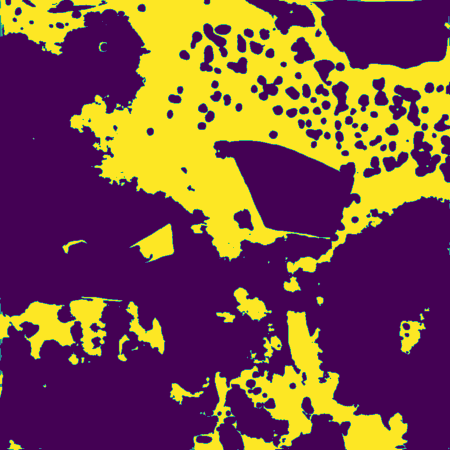}
     \end{subfigure}
     \\ \vspace{-0cm} 
     \begin{subfigure}{0.24\textwidth}
         \centering
         \includegraphics[width=\textwidth]{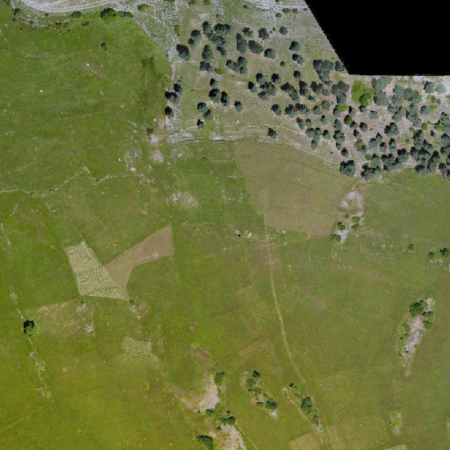}
     \end{subfigure}
     \begin{subfigure}{0.24\textwidth}
         \centering
         \includegraphics[width=\textwidth]{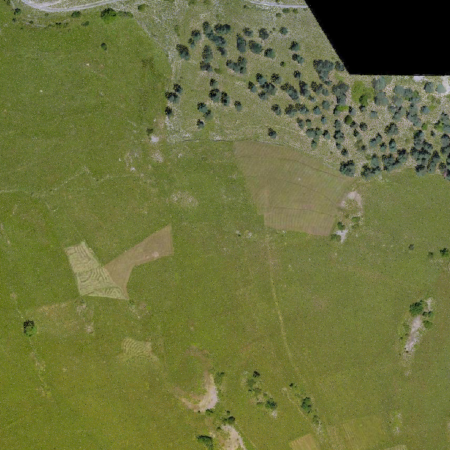}
     \end{subfigure}
     \begin{subfigure}{0.24\textwidth}
         \centering
         \includegraphics[width=\textwidth]{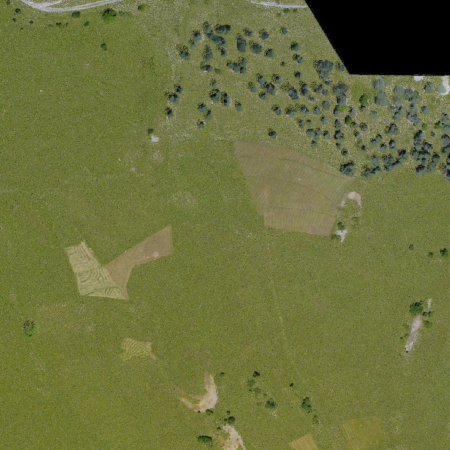}
     \end{subfigure}
     \begin{subfigure}{0.24\textwidth}
         \centering
         \includegraphics[width=\textwidth]{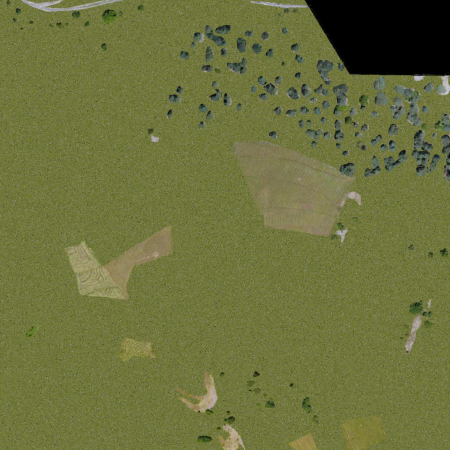}
     \end{subfigure}
     \\ \vspace{-0cm} 
     \begin{subfigure}{0.24\textwidth}
         \centering
         \includegraphics[width=\textwidth]{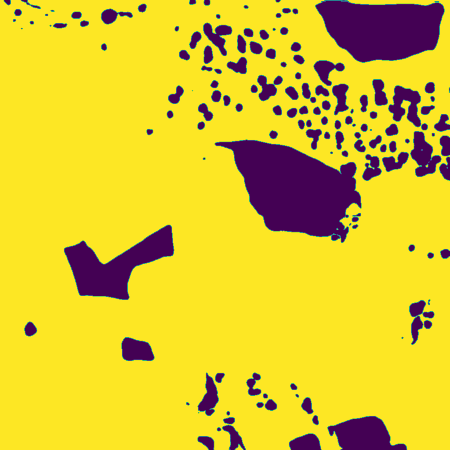}
     \end{subfigure}
     \begin{subfigure}{0.24\textwidth}
         \centering
         \includegraphics[width=\textwidth]{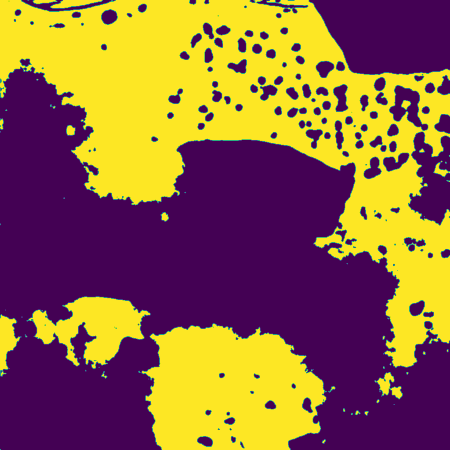}
     \end{subfigure}
     \begin{subfigure}{0.24\textwidth}
         \centering
         \includegraphics[width=\textwidth]{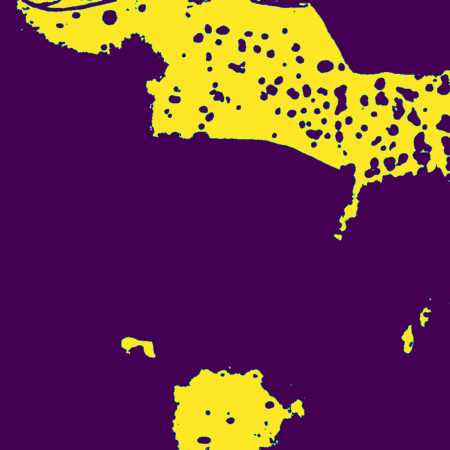}
     \end{subfigure}
     \begin{subfigure}{0.24\textwidth}
         \centering
         \includegraphics[width=\textwidth]{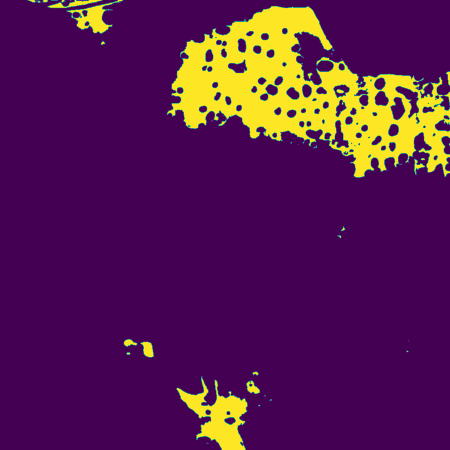}
     \end{subfigure}
     \caption{Svaneti region, transformations on the \textit{range} class. The top two rows show \textbf{gray-scale} transformed images with gray-scale proportion $\lambda \in \{0, 0.33, 0.66, 1\}$ (from left to right) and corresponding model predictions. The bottom two rows show \textbf{pixel-swap} transformed images with proportion $p$ swapped, $p \in \{0, 0.33, 0.66, 1\}$  (from left to right) and corresponding model predictions below. We see sensitivities with respect to color distortion (top), especially with gray-scale proportion $\lambda = 1$. The model predictions are however more sensitive to texture distortion (bottom).} 
     \label{appfig:svaneti_range}
\end{figure*}

\begin{figure*}[t]
     \centering
     \begin{subfigure}{0.24\textwidth}
         \centering
         \includegraphics[width=\textwidth]{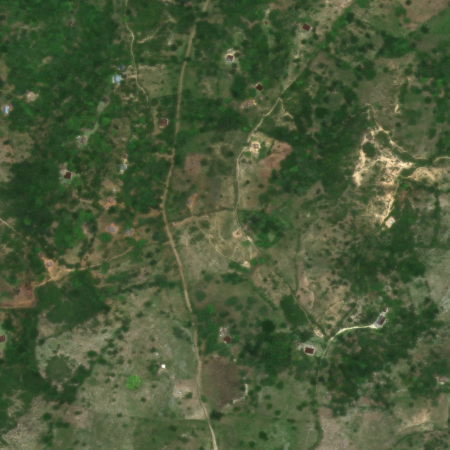}
     \end{subfigure}
     \begin{subfigure}{0.24\textwidth}
         \centering
         \includegraphics[width=\textwidth]{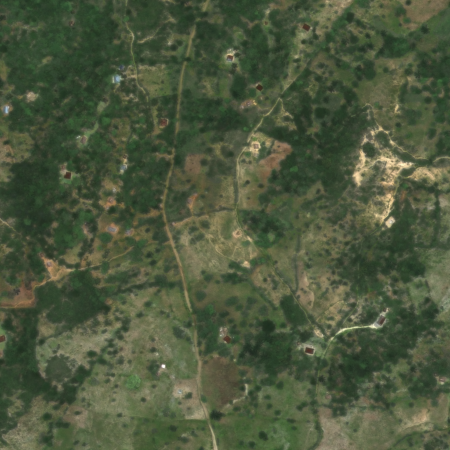}
     \end{subfigure}
     \begin{subfigure}{0.24\textwidth}
         \centering
         \includegraphics[width=\textwidth]{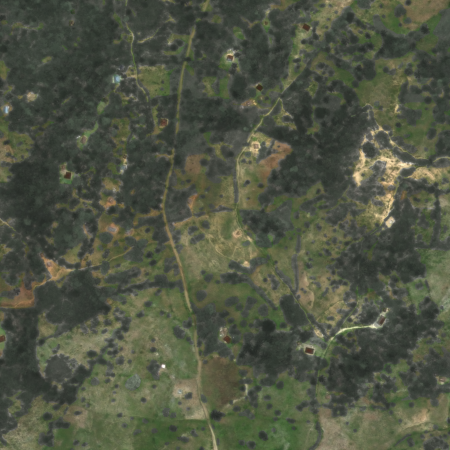}
     \end{subfigure}
     \begin{subfigure}{0.24\textwidth}
         \centering
         \includegraphics[width=\textwidth]{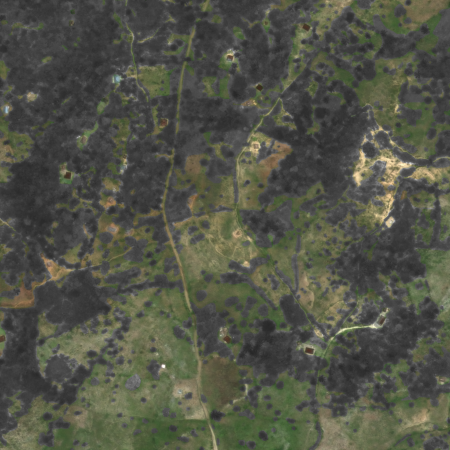}
     \end{subfigure}
     \\ \vspace{-0cm} 
     \begin{subfigure}{0.24\textwidth}
         \centering
         \includegraphics[width=\textwidth]{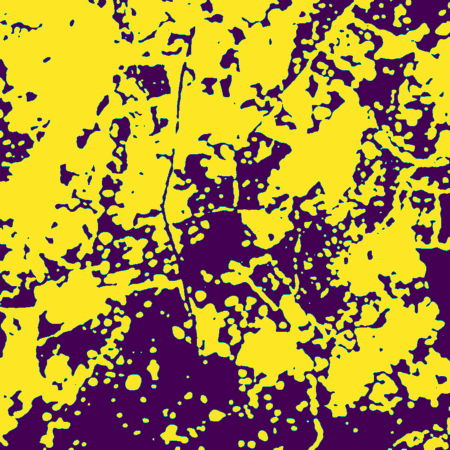}
     \end{subfigure}
     \begin{subfigure}{0.24\textwidth}
         \centering
         \includegraphics[width=\textwidth]{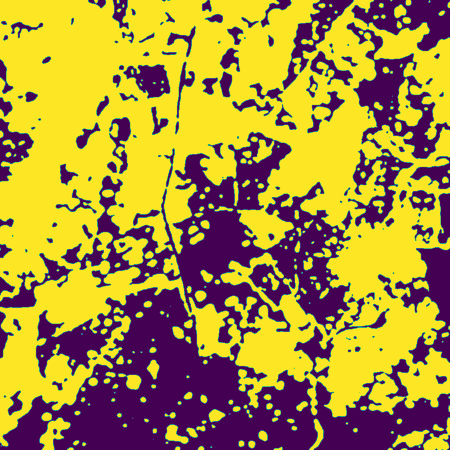}
     \end{subfigure}
     \begin{subfigure}{0.24\textwidth}
         \centering
         \includegraphics[width=\textwidth]{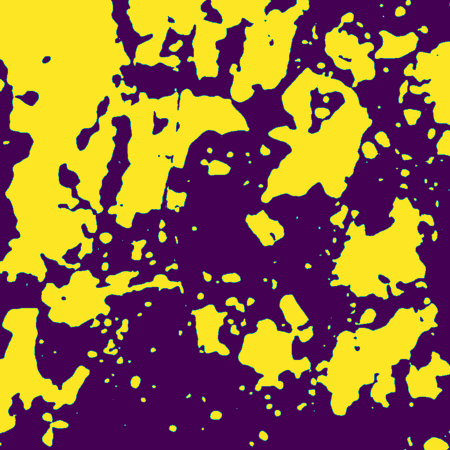}
     \end{subfigure}
     \begin{subfigure}{0.24\textwidth}
         \centering
         \includegraphics[width=\textwidth]{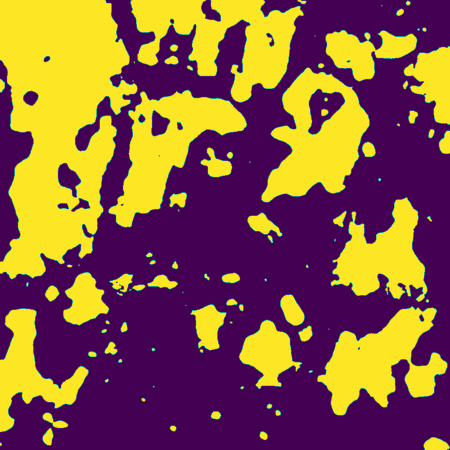}
     \end{subfigure}
     \\ \vspace{-0cm} 
     \begin{subfigure}{0.24\textwidth}
         \centering
         \includegraphics[width=\textwidth]{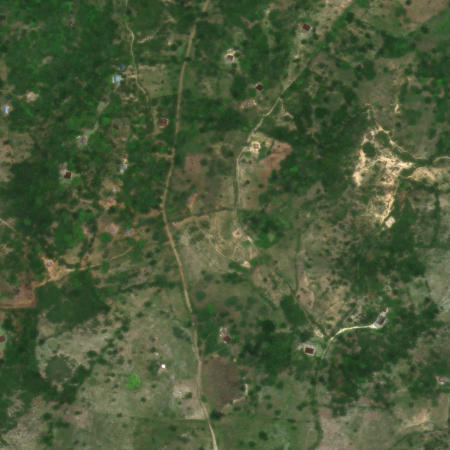}
     \end{subfigure}
     \begin{subfigure}{0.24\textwidth}
         \centering
         \includegraphics[width=\textwidth]{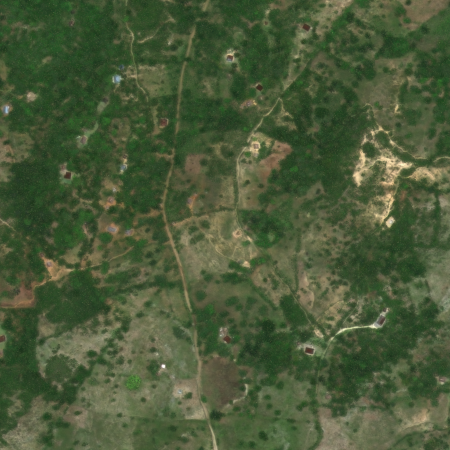}
     \end{subfigure}
     \begin{subfigure}{0.24\textwidth}
         \centering
         \includegraphics[width=\textwidth]{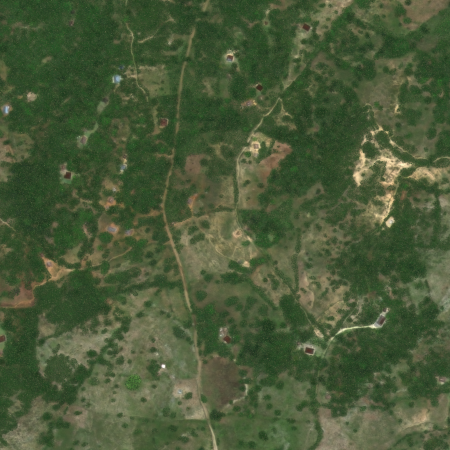}
     \end{subfigure}
     \begin{subfigure}{0.24\textwidth}
         \centering
         \includegraphics[width=\textwidth]{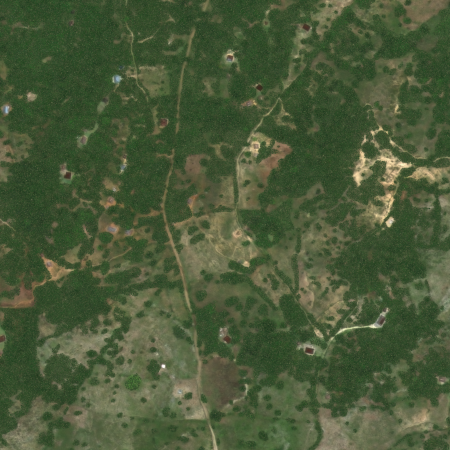}
     \end{subfigure}
     \\ \vspace{-0cm} 
     \begin{subfigure}{0.24\textwidth}
         \centering
         \includegraphics[width=\textwidth]{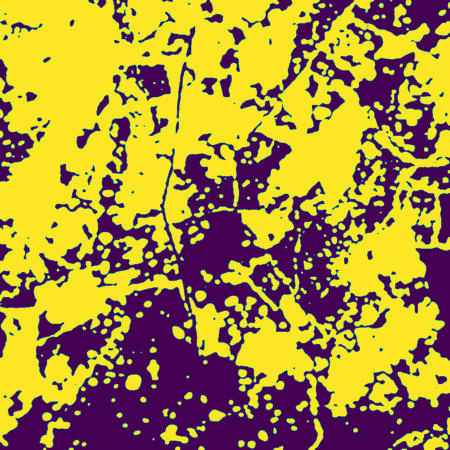}
     \end{subfigure}
     \begin{subfigure}{0.24\textwidth}
         \centering
         \includegraphics[width=\textwidth]{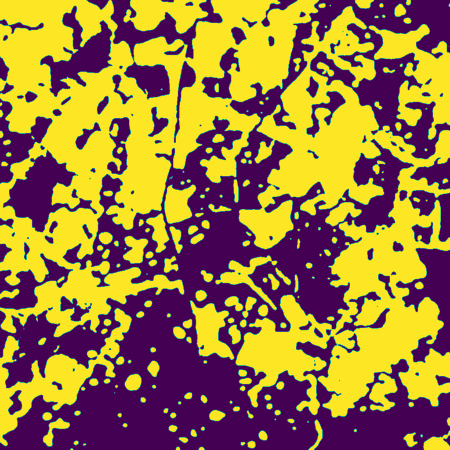}
     \end{subfigure}
     \begin{subfigure}{0.24\textwidth}
         \centering
         \includegraphics[width=\textwidth]{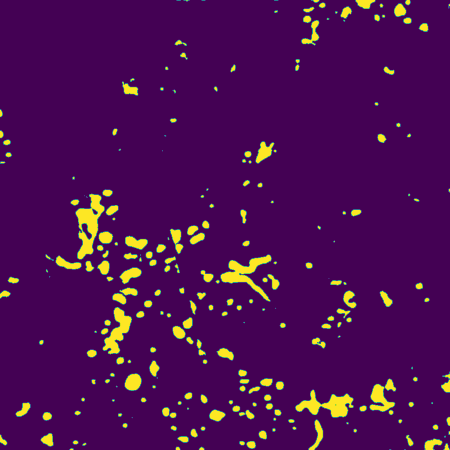}
     \end{subfigure}
     \begin{subfigure}{0.24\textwidth}
         \centering
         \includegraphics[width=\textwidth]{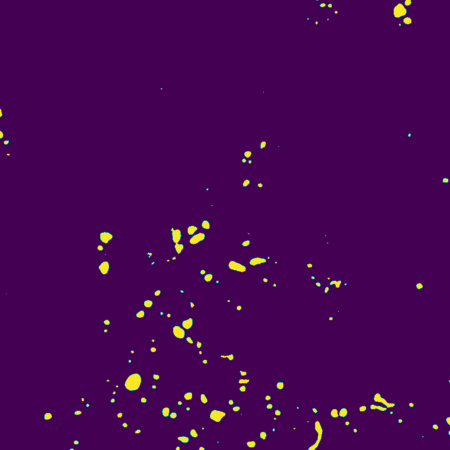}
     \end{subfigure}
     \caption{Jeremie region, transformations on the \textit{tree} class. The top two rows show \textbf{gray-scale} transformed images with gray-scale proportion $\lambda \in \{0, 0.33, 0.66, 1\}$ (from left to right) and corresponding model predictions. The bottom two rows show \textbf{pixel-swap} transformed images with proportion $p$ swapped, $p \in \{0, 0.2, 0.4, 0.6\}$  (from left to right) and corresponding model predictions below. We see that the predictions are robust with respect to color distortion with exception for gray-scale proportion $\lambda = 1$ (top),
and sensitive to texture distortion (bottom).} 
     \label{appfig:jeremie_tree}
\end{figure*}

\begin{figure*}[t]
     \centering
     \begin{subfigure}{0.24\textwidth}
         \centering
         \includegraphics[width=\textwidth]{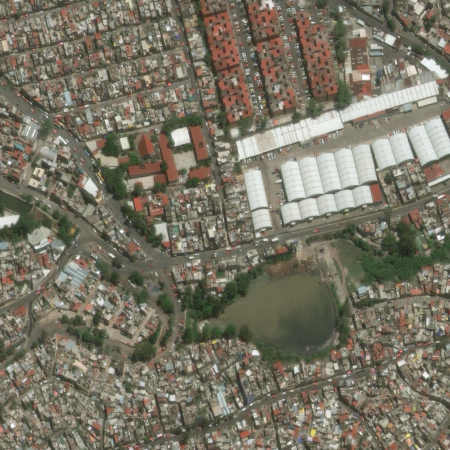}
     \end{subfigure}
     \begin{subfigure}{0.24\textwidth}
         \centering
         \includegraphics[width=\textwidth]{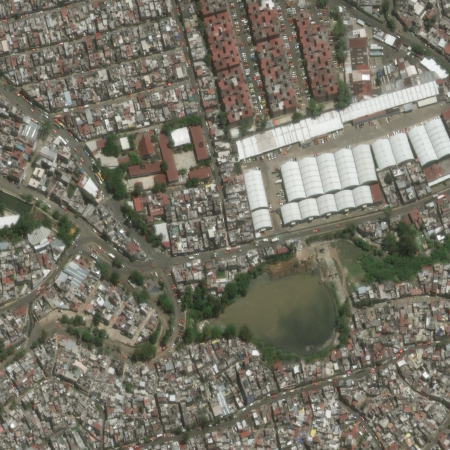}
     \end{subfigure}
     \begin{subfigure}{0.24\textwidth}
         \centering
         \includegraphics[width=\textwidth]{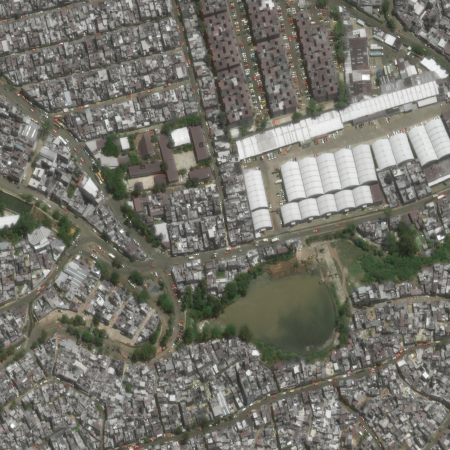}
     \end{subfigure}
     \begin{subfigure}{0.24\textwidth}
         \centering
         \includegraphics[width=\textwidth]{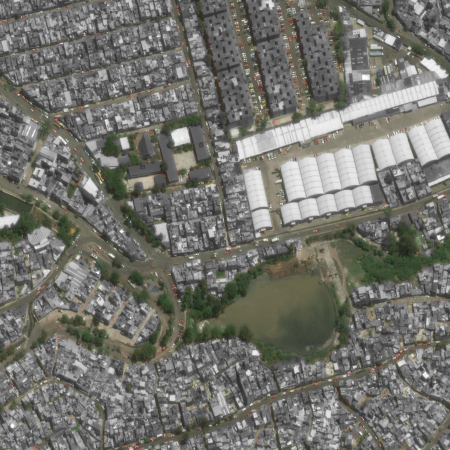}
     \end{subfigure}
     \\ \vspace{-0cm} 
     \begin{subfigure}{0.24\textwidth}
         \centering
         \includegraphics[width=\textwidth]{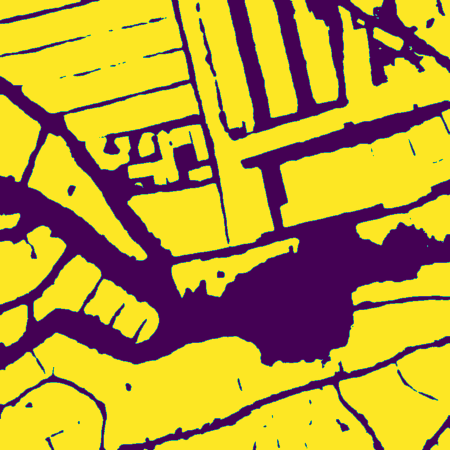}
     \end{subfigure}
     \begin{subfigure}{0.24\textwidth}
         \centering
         \includegraphics[width=\textwidth]{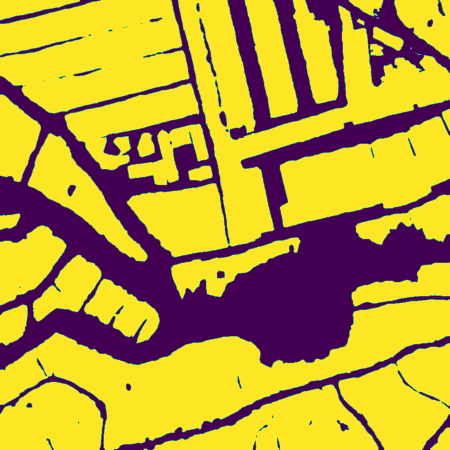}
     \end{subfigure}
     \begin{subfigure}{0.24\textwidth}
         \centering
         \includegraphics[width=\textwidth]{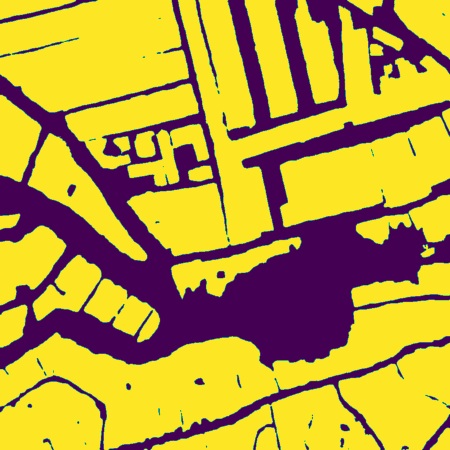}
     \end{subfigure}
     \begin{subfigure}{0.24\textwidth}
         \centering
         \includegraphics[width=\textwidth]{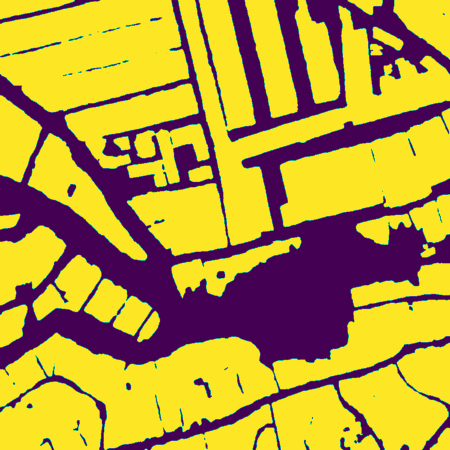}
     \end{subfigure}
     \\ \vspace{-0cm} 
     \begin{subfigure}{0.24\textwidth}
         \centering
         \includegraphics[width=\textwidth]{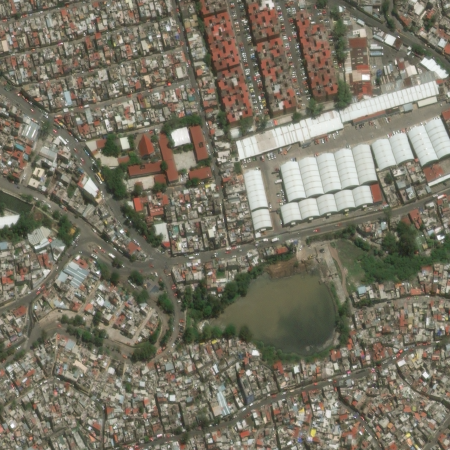}
     \end{subfigure}
     \begin{subfigure}{0.24\textwidth}
         \centering
         \includegraphics[width=\textwidth]{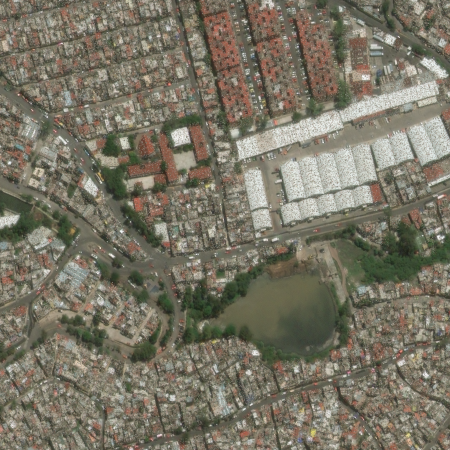}
     \end{subfigure}
     \begin{subfigure}{0.24\textwidth}
         \centering
         \includegraphics[width=\textwidth]{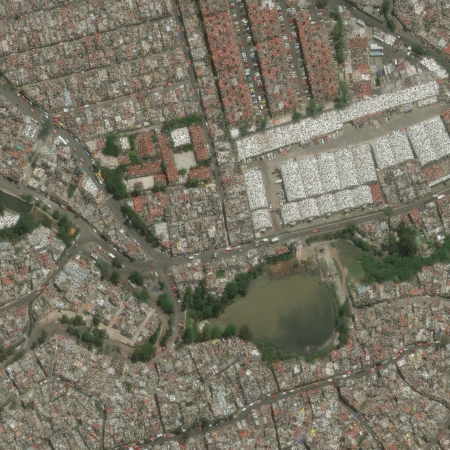}
     \end{subfigure}
     \begin{subfigure}{0.24\textwidth}
         \centering
         \includegraphics[width=\textwidth]{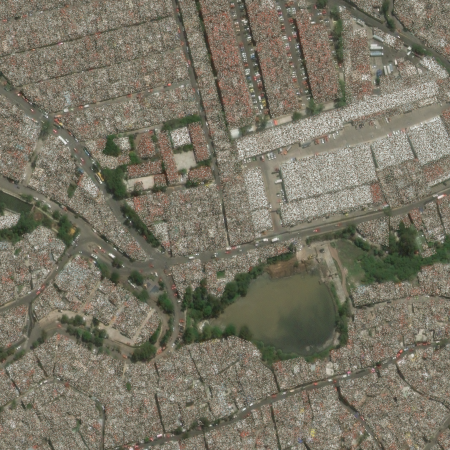}
     \end{subfigure}
     \\ \vspace{-0cm} 
     \begin{subfigure}{0.24\textwidth}
         \centering
         \includegraphics[width=\textwidth]{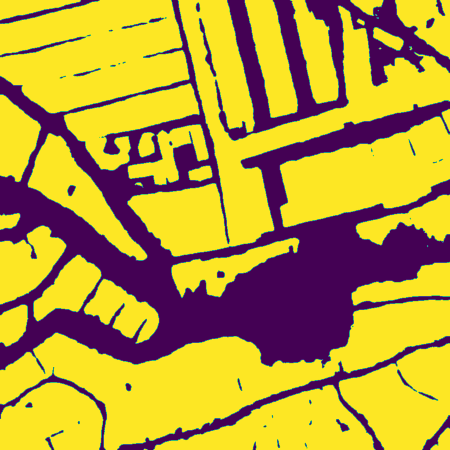}
     \end{subfigure}
     \begin{subfigure}{0.24\textwidth}
         \centering
         \includegraphics[width=\textwidth]{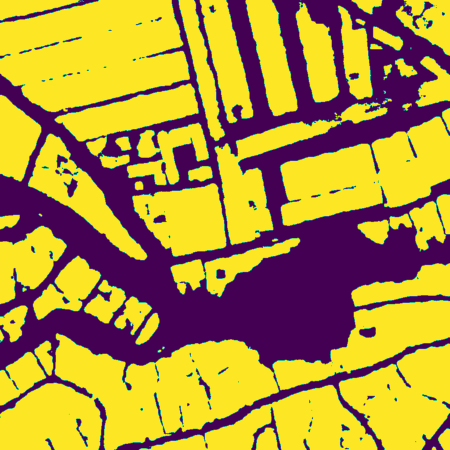}
     \end{subfigure}
     \begin{subfigure}{0.24\textwidth}
         \centering
         \includegraphics[width=\textwidth]{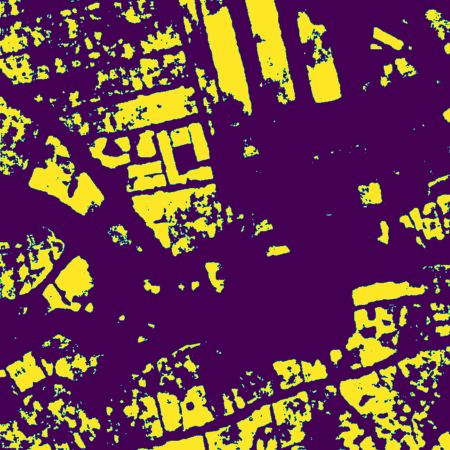}
     \end{subfigure}
     \begin{subfigure}{0.24\textwidth}
         \centering
         \includegraphics[width=\textwidth]{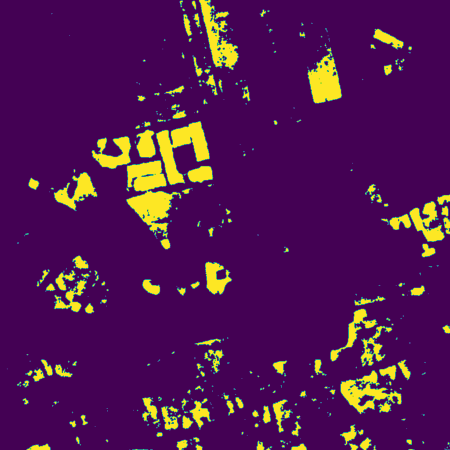}
     \end{subfigure}
     \caption{Mexico City region, transformations on the \textit{building} class. The top two rows show \textbf{gray-scale} transformed images with gray-scale proportion $\lambda \in \{0, 0.33, 0.66, 1\}$ (from left to right) and corresponding model predictions. The bottom two rows show \textbf{pixel-swap} transformed images with proportion $p$ swapped, $p \in \{0, 0.2, 0.4, 0.6\}$  (from left to right) and corresponding model predictions below. We see that the predictions are very robust with respect to color distortion (top),
and sensitive to texture distortion (bottom).} 
     \label{appfig:mexico_city_building}
\end{figure*}

\begin{figure*}[t]
     \centering
     \begin{subfigure}{0.24\textwidth}
         \centering
         \includegraphics[width=\textwidth]{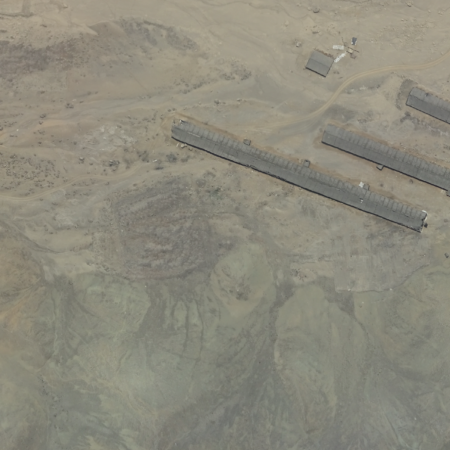}
     \end{subfigure}
     \begin{subfigure}{0.24\textwidth}
         \centering
         \includegraphics[width=\textwidth]{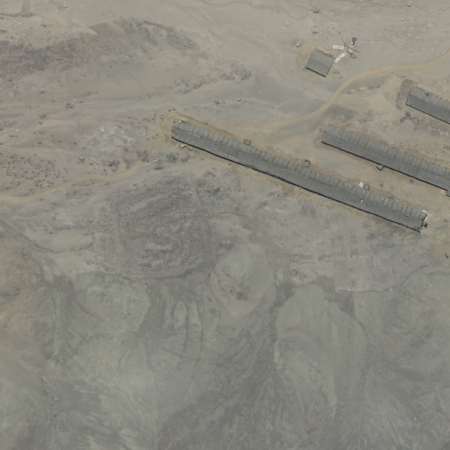}
     \end{subfigure}
     \begin{subfigure}{0.24\textwidth}
         \centering
         \includegraphics[width=\textwidth]{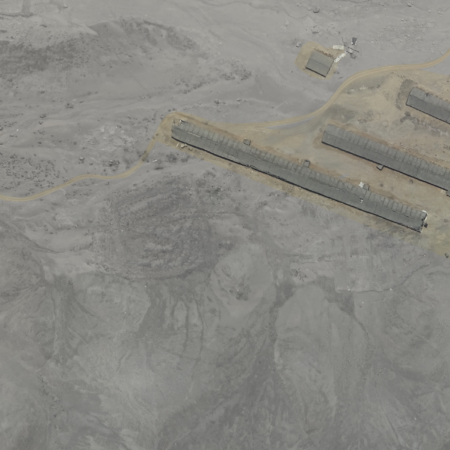}
     \end{subfigure}
     \begin{subfigure}{0.24\textwidth}
         \centering
         \includegraphics[width=\textwidth]{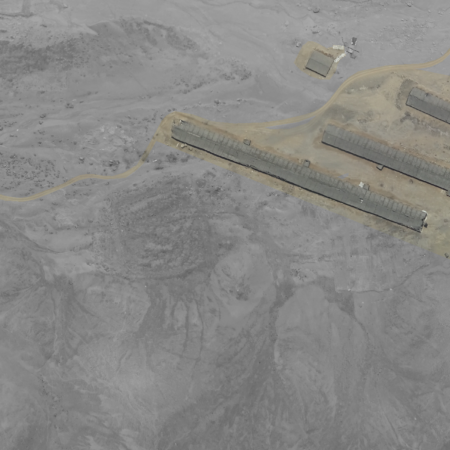}
     \end{subfigure}
     \\ \vspace{-0cm} 
     \begin{subfigure}{0.24\textwidth}
         \centering
         \includegraphics[width=\textwidth]{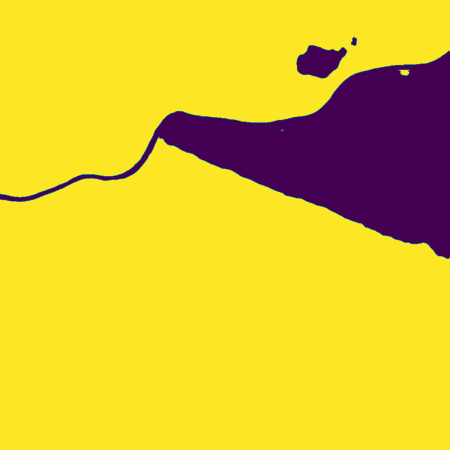}
     \end{subfigure}
     \begin{subfigure}{0.24\textwidth}
         \centering
         \includegraphics[width=\textwidth]{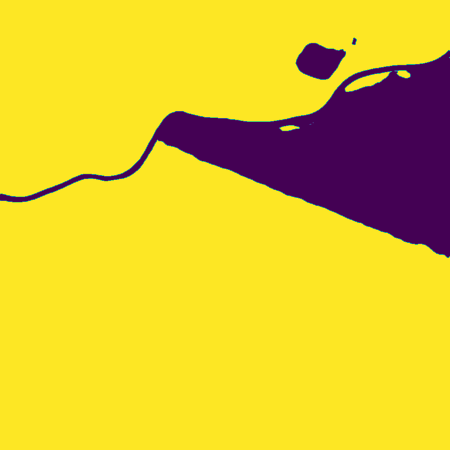}
     \end{subfigure}
     \begin{subfigure}{0.24\textwidth}
         \centering
         \includegraphics[width=\textwidth]{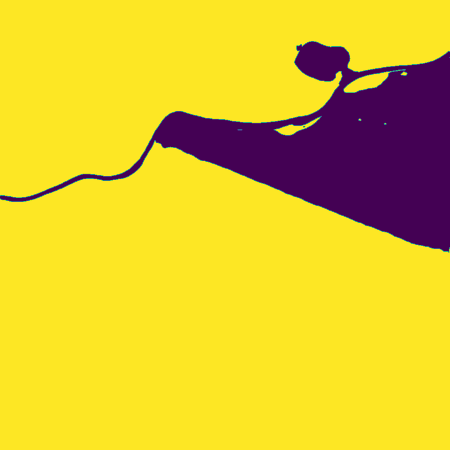}
     \end{subfigure}
     \begin{subfigure}{0.24\textwidth}
         \centering
         \includegraphics[width=\textwidth]{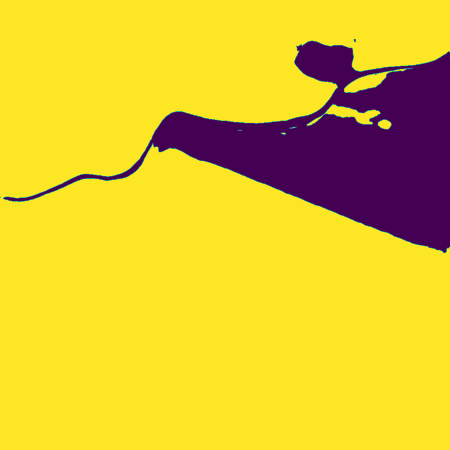}
     \end{subfigure}
     \\ \vspace{-0cm} 
     \begin{subfigure}{0.24\textwidth}
         \centering
         \includegraphics[width=\textwidth]{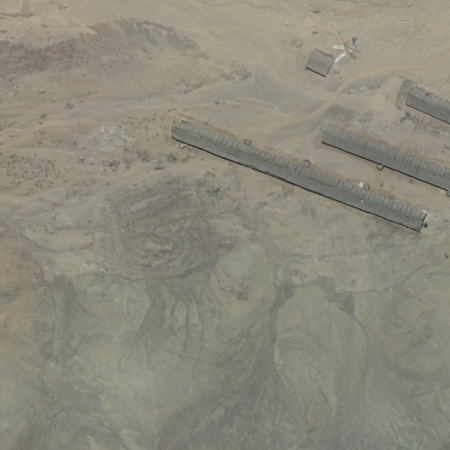}
     \end{subfigure}
     \begin{subfigure}{0.24\textwidth}
         \centering
         \includegraphics[width=\textwidth]{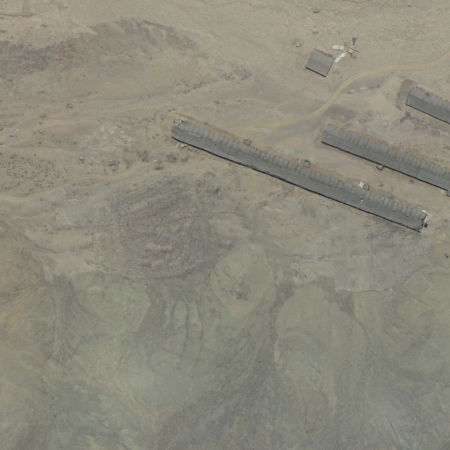}
     \end{subfigure}
     \begin{subfigure}{0.24\textwidth}
         \centering
         \includegraphics[width=\textwidth]{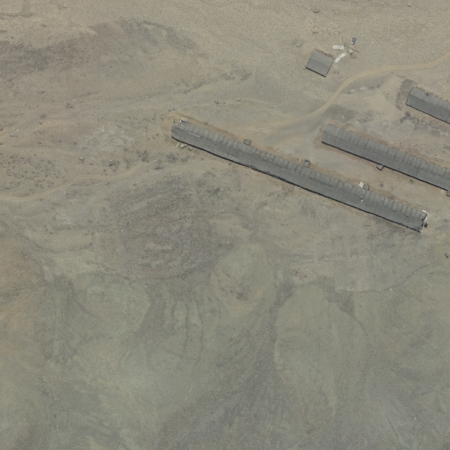}
     \end{subfigure}
     \begin{subfigure}{0.24\textwidth}
         \centering
         \includegraphics[width=\textwidth]{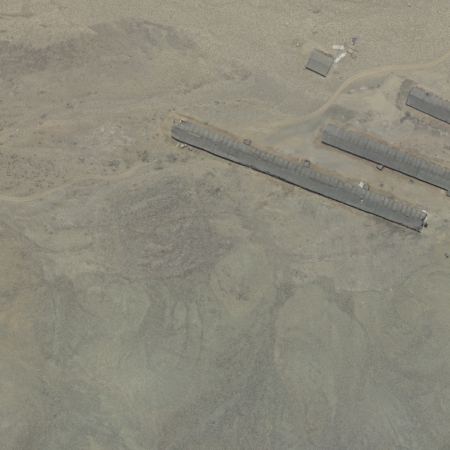}
     \end{subfigure}
     \\ \vspace{-0cm} 
     \begin{subfigure}{0.24\textwidth}
         \centering
         \includegraphics[width=\textwidth]{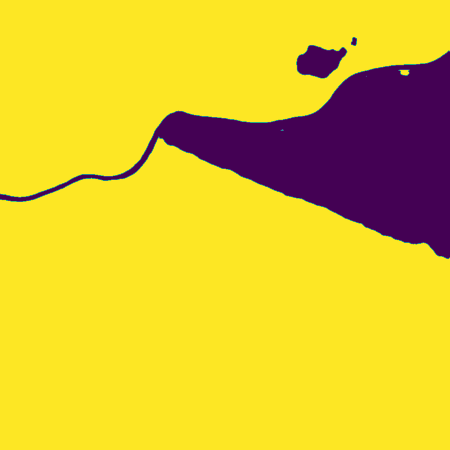}
     \end{subfigure}
     \begin{subfigure}{0.24\textwidth}
         \centering
         \includegraphics[width=\textwidth]{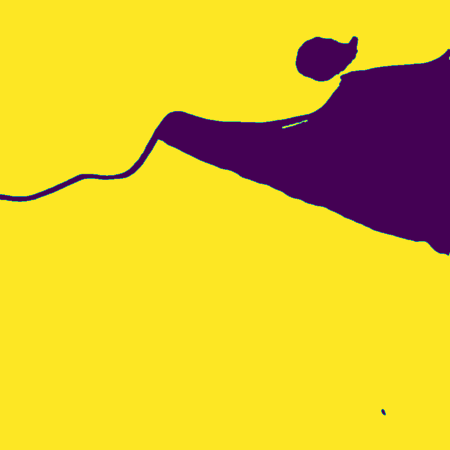}
     \end{subfigure}
     \begin{subfigure}{0.24\textwidth}
         \centering
         \includegraphics[width=\textwidth]{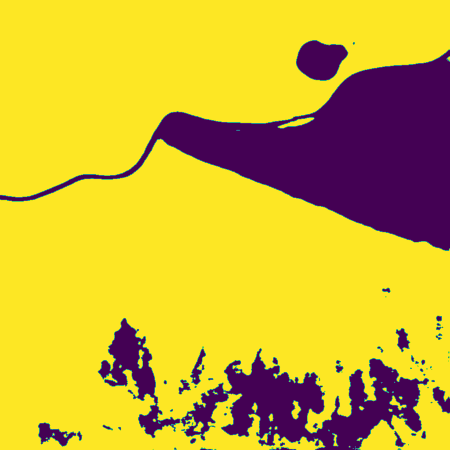}
     \end{subfigure}
     \begin{subfigure}{0.24\textwidth}
         \centering
         \includegraphics[width=\textwidth]{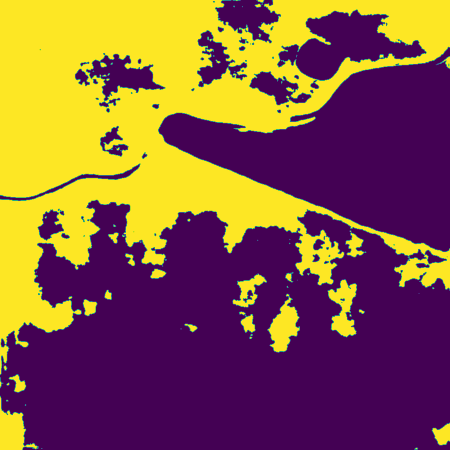}
     \end{subfigure}
     \caption{Lima region, transformations on the \textit{bare} class. The top two rows show \textbf{gray-scale} transformed images with gray-scale proportion $\lambda \in \{0, 0.33, 0.66, 1\}$ (from left to right) and corresponding model predictions. The bottom two rows show \textbf{pixel-swap} transformed images with proportion $p$ swapped, $p \in \{0, 0.1, 0.2, 0.3\}$  (from left to right) and corresponding model predictions below. We see that the predictions are very robust with respect to color distortion (top),
and very sensitive to texture distortion (bottom).} 
     \label{appfig:lima_bare}
\end{figure*}

\textbf{More results for the image distortion experiments in the main paper.} Here we show more visual results for the gray-scale and pixel-swap distortion experiments (i.e.~the distortions investigated in the main paper); see \Figure{appfig:austin_road} - \ref{appfig:lima_bare}. In particular, we show qualitative results for each of the eight classes in the dataset, as we change the intensity of the respective distortions. Note that the illustrated distortion intensity ranges vary between the examples; these range choices are to a large extent guided by the class-specific sensitivities seen in \Figure{fig:main_result_plot} (cf.~main paper), and are set in such a way that interesting accuracy degradations are highlighted where possible. All model predictions in these qualitative examples are from the U-Net-Efficientnet-B4 model. Finally, for completeness we also show a plot similar to \Figure{fig:main_result_plot} (main paper), but for the training set -- see \Figure{fig:main_results_plot_train}.

\end{document}